\documentclass[3p]{elsarticle}

\usepackage{lineno,hyperref, booktabs, textcomp, multirow, tabularx}
\usepackage{color, colortbl}

\modulolinenumbers[5]

\usepackage{ragged2e}
\usepackage{array}
\usepackage{tcolorbox}
\usepackage{subcaption}
\usepackage{soul}

\newenvironment{blocktemplate}
    {\begin{center}
    \begin{tcolorbox}[width=0.6\linewidth]
    \fontfamily{cmtt}\selectfont\small
    }
    { 
    \par
    \end{tcolorbox} 
    \end{center}
    }

\newcommand{\symtilde}{\raise.3ex\hbox{$\scriptstyle\sim$}}

\usepackage[normalem]{ulem}

\begin{document}

\begin{frontmatter}

\title{Acquiring and Modeling Abstract Commonsense Knowledge via Conceptualization}

\author{Mutian He}
\ead{mhear@connect.ust.hk}

\author{Tianqing Fang}
\ead{tfangaa@cse.ust.hk}

\author{Weiqi Wang}
\ead{wwangbw@cse.ust.hk}

\author{Yangqiu Song}
\address{The Hong Kong University of Science and Technology}
\ead{yqsong@cse.ust.hk}

\begin{abstract}
Conceptualization, or viewing entities and situations as instances of abstract concepts in mind and making inferences based on that, is a vital component in human intelligence for commonsense reasoning. 
Despite recent progress in artificial intelligence to acquire and model commonsense attributed to neural language models and commonsense knowledge graphs (CKGs), conceptualization is yet to be introduced thoroughly, making current approaches ineffective to cover knowledge about countless diverse entities and situations in the real world.
To address the problem, we thoroughly study the role of conceptualization in commonsense reasoning, and formulate a framework to replicate human conceptual induction by acquiring abstract knowledge about events regarding abstract concepts, as well as higher-level triples or inferences upon them.
We then apply the framework to ATOMIC, a large-scale human-annotated CKG, aided by the taxonomy Probase. We annotate a dataset on the validity of contextualized conceptualizations from ATOMIC on both event and triple levels, develop a series of heuristic rules based on linguistic features, and train a set of neural models to generate and verify abstract knowledge. 
Based on these components, a pipeline to acquire abstract knowledge is built. 
A large abstract CKG upon ATOMIC is then induced, ready to be instantiated to infer about unseen entities or situations. 
Finally, we empirically show the benefits of augmenting CKGs with abstract knowledge in downstream tasks like commonsense inference and zero-shot commonsense QA.

\end{abstract}


\end{frontmatter}


\section{Introduction}
\label{Sec:intro}

Commonsense knowledge like ``people want to eat when feeling hungry,'' as a critical component for artificial intelligence, has regained focus among researchers and obtained substantial progress in recent few years. 
This is particularly signified by commonsense knowledge graphs (CKGs), the predominant form to represent commonsense. 
Recent large-scale text-based CKGs collect and solidify the implicit knowledge into triples $\langle h,r,t \rangle$, with the textual head node \emph{h} and tail node \emph{t} connected by a relation (i.e., edge) \emph{r}. 
ATOMIC \cite{DBLP:conf/aaai/SapBABLRRSC19} is a typical example, which contains amounts of human-annotated triples for head events and their causes or consequences in loosely-formed texts, such as $\langle$\emph{h}: \textit{PersonX is hungry}, \emph{r}: xWant (then PersonX wants), \emph{t}: \textit{to have lunch}$\rangle$.
With a large amount of commonsense knowledge consolidated into CKGs as in Table~\ref{Tab:CKG_size}, ATOMIC-like CKGs~\cite{DBLP:conf/aaai/SapBABLRRSC19, DBLP:conf/emnlp/ForbesHSSC20socialchemistry, DBLP:conf/emnlp/MostafazadehKMB20, zhang2022aser, DBLP:conf/www/FangZWSH21} have been widely utilized in a broad range of downstream scenarios, including dialogue systems~\cite{DBLP:conf/aaai/YoungCCZBH18, DBLP:journals/corr/abs-2212-10465,DBLP:conf/aaai/SabourZH22}, narrative understanding~\cite{DBLP:conf/aaai/GabrielBSBFC21, DBLP:conf/aaai/ChenCY19}, script learning~\cite{DBLP:conf/acl/ZhouGSPZJ21, DBLP:conf/coling/LvZH20}, and question answering (QA) \cite{DBLP:conf/emnlp/LinCCR19, DBLP:conf/aaai/MaIFBNO21, DBLP:conf/emnlp/ShwartzWBBC20,DBLP:conf/naacl/KimKKAHY22}.

\begin{table}[h]
\centering
\begin{tabular}{lcccc}
\toprule
& ATOMIC \cite{DBLP:conf/aaai/SapBABLRRSC19} & ATOMIC-2020 \cite{DBLP:conf/aaai/HwangBBDSBC21} & DISCOS \cite{DBLP:conf/www/FangZWSH21}\\
\midrule
\#Head events & 24.3K & 44.0K & 1,103.0K \\
\#Triple & 793.3K & 1,246.6K & 3,235.9K \\
Average Degree  & 32.6 & 28.4 & 2.9 \\ \midrule
Appeared Probase Concept & 0.34\% & 0.76\% & 8.00\%\\
Average Distinct Concept & 0.093 & 0.114 & 0.048 \\
\bottomrule
\end{tabular}
\caption{\label{Tab:CKG_size} Statistics of recent ATOMIC-like textual event-centric large-scale CKGs. Considering Probase concepts of at least ten occurrences, Appeared Probase Concept is calculated by the proportion of concepts directly mentioned in heads by word matching, and Average Distinct Concept is the average number of distinct concepts per head. 
}
\end{table}

\begin{figure}[t]
\centering
\includegraphics[width=\textwidth]{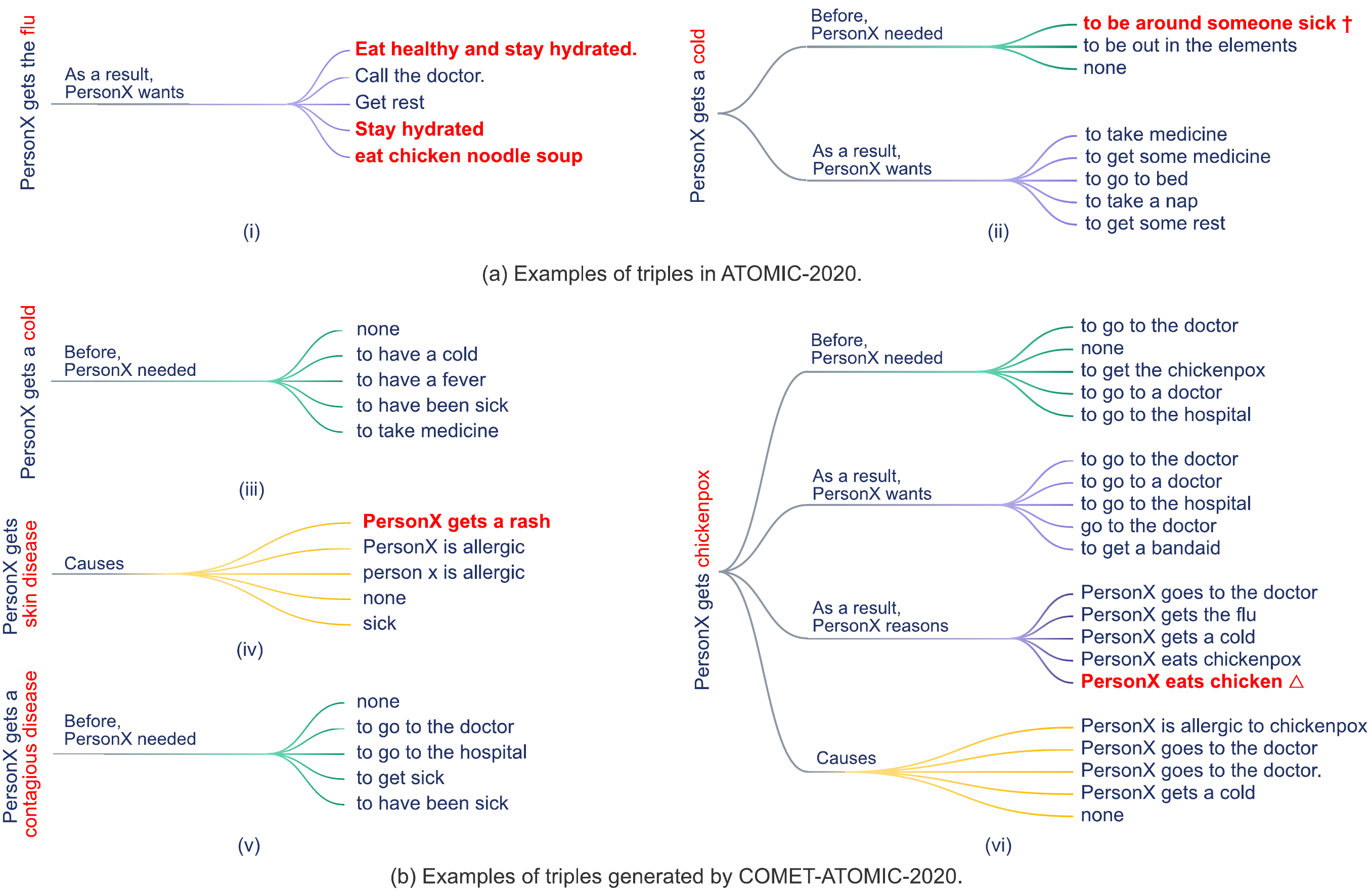}
\caption{Examples of triples from ATOMIC-2020 and COMET-ATOMIC-2020 from their online demo (available at \url{https://mosaickg.apps.allenai.org/home}), indicating the insufficient coverage and lack of generalization: Reasonable consequences in (i) applicable to many diseases are not covered in (ii). A root cause of (ii) is covered ($\dag$), but COMET fails to predict that for other \textit{contagious diseases} 
 in (iii), (vi), and (v) when we directly query about that abstract concept. It also gives a common consequence of \textit{skin diseases} (iv), but not in its instance \textit{chickenpox} (vi), and reflects superficial textual links instead ($\bigtriangleup$). }
\label{Fig:kg_samples}
\end{figure}

Despite the success of current CKGs, conceptualization, a key ingredient in human intelligence and commonsense remains \textit{terra incognita}. There is an infinite number of entities and situations in the real world, including vast objects, persons, events, states, and actions. 
Even with the extremely large scale as in Table~\ref{Tab:CKG_size}, finite CKGs may never cover all the events involving inexhaustible entities and situations,
let alone the relations or triples between them.
For example, as shown in Figure~\ref{Fig:kg_samples} (i), $\langle$\textit{PersonX gets the flu}, xWant, \textit{get hydrated}$\rangle$ is covered by ATOMIC-2020~\cite{DBLP:conf/aaai/HwangBBDSBC21}, an extension of ATOMIC combined with ConceptNet~\cite{DBLP:conf/aaai/SpeerCH17}. But this tail is not found for the head \textit{PersonX gets a cold} (ii), although the concepts of \textit{flu} and \textit{cold} are closely related as similar respiratory infections, making the tail valid for both of them. Moreover, in this finite CKG, we won't be able to find all such heads like \textit{PersonX gets pneumonia}, \textit{PersonX gets chickenpox}, \textit{PersonX gets a bad flu}, etc. Using Probase, a large-scale taxonomy of concepts for entities and situations \cite{wu2012probase}, we inspect the proportion appearing in ATOMIC as well DISCOS, which is automatically generated and hence much larger. Despite the large scale, most concepts covered by Probase are never mentioned in the head events as indicated by \textit{Appeared Probase Concept}, not to say potential triples or inferences about them.
It is similar in ATOMIC-2020 even though it explicitly incorporates ConceptNet for more knowledge about physical entities. Furthermore, as shown by the low \textit{Average Distinct Concepts} per node, directly expanding the scale of CKGs would not be cost-effective to cover various entities and situations.

Alternatively, commonsense models like COMET \cite{DBLP:conf/acl/BosselutRSMCC19} and COMET-ATOMIC-2020 \cite{DBLP:conf/aaai/HwangBBDSBC21} claim to bypass this challenge by employing the text-form nodes.
Fine-tuned on ATOMIC from recent pretrained language models (PTLMs) like GPT2 \cite{radford2019gpt2} or BART \cite{DBLP:conf/acl/LewisLGGMLSZ20}, they are able to draw inferences by generating possible tails given arbitrary ATOMIC-like textual events. 
However, neural language models (LM) follow the principle of \textbf{induction on word co-occurrence} like the co-appearance of pox and their causes in texts. It is doubtful if COMET can truly go beyond superficial textual clues, as exemplified in Figure~\ref{Fig:kg_samples} (vi) $\dag$ linking the cause of chickenpox to chicken. In fact, there is evidence that such LMs suffer from limited generalization to broader scenarios \cite{DBLP:conf/acl/WangICR21,DBLP:conf/emnlp/ZhouKLLHPR21}. 
Specifically, as a \textit{contagious disease}, a critical precondition to \textit{get a cold} is to \textit{be around someone sick}. But even this triple is covered by ATOMIC-2020 as in (vi) $\bigtriangleup$, COMET could only generate some over-general or even meaningless causes of \textit{get a cold} as in (iii), not to mention the causes of relevant events like \textit{gets a contagious disease} (v) or \textit{chickenpox} (vi). Admittedly, COMET shows some generalization to unseen entities and situations, such as associate \textit{chickenpox} with \textit{bandaid}. 
But in (vi) many critical inferences like \textit{need to be around patients}, \textit{get a rash}, etc. are missing. Furthermore, it is impossible for COMET to infer about knowledge beyond training, such as \textit{PersonX gets COVID-19}, not to mention the prohibitive computation costs and data requirements of such large models.

Instead, humans rely on \textbf{conceptualization}.
\textit{Concepts are the glue that holds our mental world together} \cite{murphy2004concept1}.
We capture the commonsense in this world through conceptualization, e.g., through summarizing what we learn from each particular past experience of seeing each movie, and linking them to the abstract concept of ``seeing a movie'' in our mind, so that we can realize what to do in its future new instances. 
Without concepts in mind, one may not know that a new apple is edible, no matter how many apples the person has eaten. 
Without concepts in mind, one may not know how to buy tickets whenever seeing a new movie, even if the person has done it before. 
Intelligence without concepts, either human or artificial, would be lost in a mess of isolated facts on disconnected instances, unable to understand the world as a whole.
For people not knowing chickenpox, when told that it is an instance of \textit{contagious disease} or \textit{skin disease}, they may easily draw the critical inferences mentioned above. This is from knowledge about the abstract concepts of contagious disease and skin disease, i.e. \textbf{abstract knowledge} \cite{tenenbaum2011grow}. Such abstract knowledge cannot be directly experienced, but indirectly induced from experiences on their instances like flu, measles, etc. This \textbf{conceptual induction} \cite{murphy2004concept8} process leverages inherent taxonomic relations between concepts instead of mere word co-occurrence. This is not limited to objects: Knowing that getting chickenpox is a painful experience, people can easily infer that others will feel sorry for that just like other painful experiences. People understand and reason about events, actions, and states in a similar conceptual way.

Such conceptual induction, though desirable for AI, is never easy to replicate on machines.
First, due to the inherent flexibility of language, even for simple texts used in ATOMIC, it can be difficult to identify the actual entity or situation referred to, e.g. to decide that \textit{hot dog} is not a dog, \textit{a cup of tea} refers to \textit{tea} not \textit{cup}, and the action \textit{give a speech} is more than \textit{giving} something.
Second, an instance is possibly related to multiple concepts, depending on the context and what we intend to induce about:
A tomato could be vegetable in the kitchen but a fruit in a biology class; A dog can run since it is an animal, but it is furry since it is a mammal.
Third, we need to draw inferences about abstract concepts. Chickenpox causes rashes, and flu causes coughs. In both cases, a patient sees a doctor since both are \textit{diseases}, but an unspecified \textit{disease} doesn't typically cause rashes or coughs.
More, it suffers from reporting bias in that we talk about instances much more often than concepts. LMs trained on textual data may struggle with knowledge of more abstract concepts even if it's obvious for humans, as shown by the ``contagious disease'' example in Figure~\ref{Fig:kg_samples} (v).

Although there are sporadic works of conceptualization for acquiring and modeling commonsense, they are based on early symbolic approaches \cite{ramachandran2005first} or focused on specific and restricted scenarios such as conceptualization of arguments of each event \cite{zhang2022aser,DBLP:conf/naacl/PoradaSTC21} without considering the higher-level conceptualization of events and their relations. 
In contrast, we intend to thoroughly chart this area by formulating and implementing the whole process of conceptual induction in modern commonsense reasoning: We acquire \textbf{abstract knowledge} in the form of \textbf{abstract events} and \textbf{abstract triples}, which are summarized from existing \textbf{situational CKGs} through \textbf{conceptualization}, using a combination of taxonomy, linguistic rules, and neural models. The abstract knowledge can then be utilized to handle unseen entities and situations in the future. 
Table~\ref{Tab:glossary} summarises the terms we use to discuss the process, to be further elaborated later. 

\begin{table}[t]
\centering
{\def\arraystretch{1.5}
\begin{tabularx}{\textwidth}{p{1.7cm}>{\RaggedRight\arraybackslash}p{14.1cm}} \toprule
Term & Explanation \\ \midrule
Entity &
  A person, organization, object, phenomenon, etc., mentioned in language, usually as a noun or nominal phrase, such as ``cold" in ``PersonX gets a cold." \\
Eventuality &
  A state, action, or event mentioned in language \footnotemark , usually as a sentence or clause, such as ``PersonX gets a cold," and ``he enjoys himself" in ``PersonX says he enjoys himself." \\
Concept &
  An abstract concept in our mind, in this paper represented by nodes in a taxonomy like Probase. We consider every entity or eventuality as an instance of a concept. \\
Abstract Event &
  A textual event containing a bracket-enclosed concept like ``PersonX gets [disease]," representing a \textit{class} of its \textit{instance events}, like ``PersonX gets flu," ``PersonX gets a cold," etc. \\
Abstract Triple &
  A triple with an abstract head event, representing an inference on the head, typically shared in a class of events, such as  $\langle$\emph{h}: PersonX gets [disease], \emph{r}: xWant, \emph{t}: see a doctor$\rangle$. \\
Situational Knowledge &
  Knowledge represented in current event-centric CKGs (situational CKGs) on concrete and specific events and their inferences. \\
Abstract Knowledge &
  Knowledge summarized from situational knowledge regarding concepts, which is generalizable to the instances, represented in abstract events and triples. \\
Event/Triple Conceptualization &
  The process or action to build an abstract event/triple based on the a textual event/triple from the situational knowledge. This is by conceptualizing an entity/eventuality mentioned in the text into a concept and then replacing it with that concept. \\
\bottomrule
\end{tabularx}
}
\caption{\label{Tab:glossary} Explanation of several important terms we use in this paper, based on the terminology of ASER \cite{zhang2022aser} and Probase \cite{wu2012probase}.}
\end{table}
\footnotetext[2]{~\textit{Eventuality} \cite{bach1986algebra}, used interchangeably with \textit{situation} \cite{mourelatos1978events}, is a linguistic term denoting various types of predicates abovementioned in the general sense. We keep the ATOMIC terminology to call heads as \textit{events}, though it is linguistically imprecise.}

We leverage PTLMs for their strong language understanding capability, which is critical to capturing the contextualized meaning of words in a sentence to decide the appropriate conceptualization.
They are fine-tuned on a large dataset of valid and invalid conceptualizations we annotate on 92K entities and situations (eventualities) in head events, as well as 81K abstract triples upon them. 
In this way, the models are not merely induced from language but also aware of conceptual relations.
Through this pipeline, we induce a conceptualized version of ATOMIC, \textbf{Abstract ATOMIC}, covering 70K abstract events and 2.9M abstract triples. 
We further show that directly incorporating Abstract ATOMIC is already helpful in downstream tasks such as COMET-like inference and zero-shot commonsense QA. 
To summarize, our contributions are:
\begin{enumerate}
    \item We formulate the task and framework of machine conceptual induction and abstract knowledge acquisition.
    \item We annotate a large-scale dataset for valid and invalid conceptualization on both event and triple level.
    \item We develop a series of neural models and sophisticated heuristic rules to carry out the steps in the pipeline of abstract knowledge acquisition.
    \item We induce a sizable set of abstract knowledge from ATOMIC using our pipeline.
    \item We investigate the value of abstract knowledge in commonsense modeling and reasoning tasks.
\end{enumerate}

\begin{figure}[t]
\centering
\includegraphics[width=0.85\textwidth]{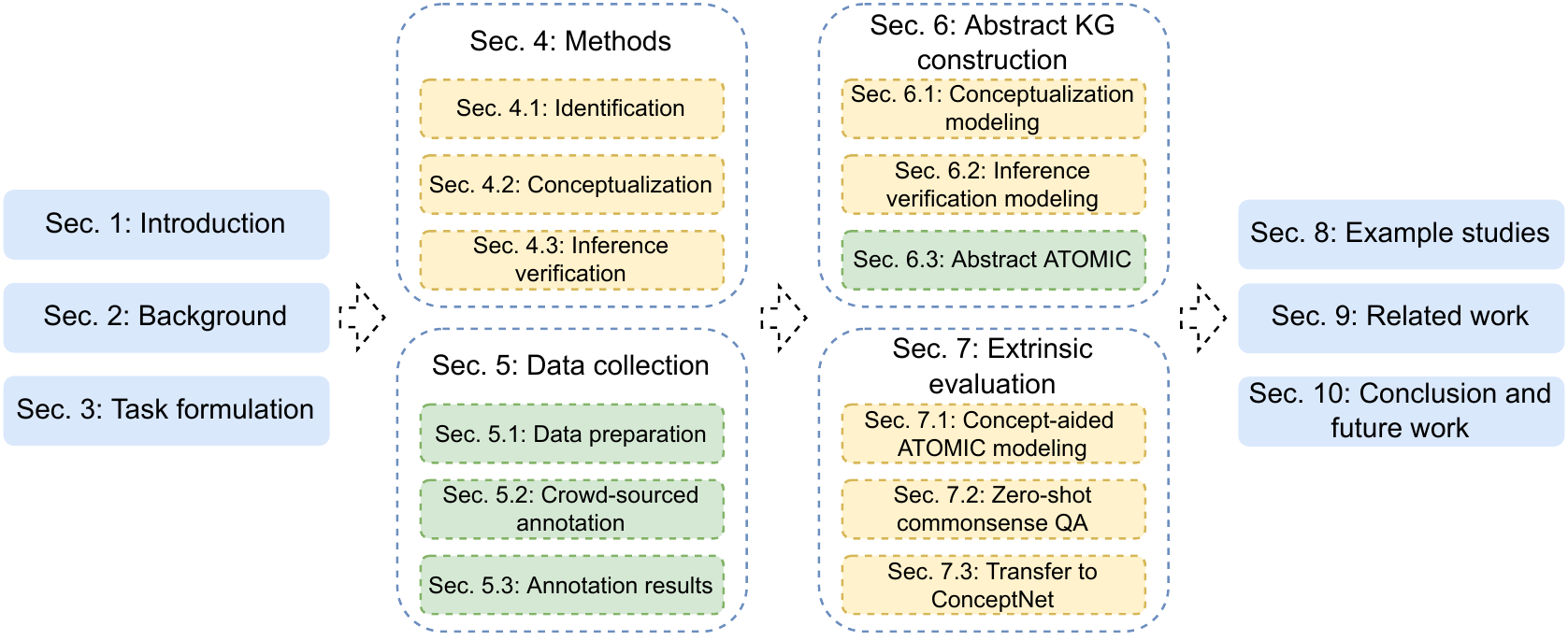}
\caption{Paper roadmap. Parts focused on empirical methods are marked in yellow, and those on data processing in green.}
\label{Fig:nav}
\end{figure}

As shown in Figure~\ref{Fig:nav}, the paper is organized as follows: Section 2 discusses the background of commonsense knowledge, its representation and modeling, and the cognitive process of conceptualization. Section 3 formulates the three-stage abstract knowledge acquisition task for conceptual induction. Section 4 elaborates on the exact methods to implement each stage. Section 5 explains the data collection process. Section 6 builds the neural models in each stage using the data, which are then applied to harvest an abstract CKG from ATOMIC. Section 7 evaluates the abstract CKG in downstream tasks. Section 8 studies examples from the data. Section 9 reviews related previous works. Section 10 concludes the paper. All data and codes are publicly released at \url{https://github.com/HKUST-KnowComp/atomic-conceptualization}.

\section{Background}

In this section, we introduce the important background of this study, including the characteristics of commonsense knowledge, relevant CKGs and taxonomies, the role of PTLMs for commonsense, previous research on human conceptualization, as well as their implications for our design choices.

\subsection{Commonsense knowledge}

Commonsense knowledge, like {\em a person will pay the bill after buying groceries}  and {\em a person is larger than a dog}, is a set of ``millions of basic facts and understanding possessed by most people'' \cite{liu2004conceptnet}. 
Even though such commonsense can be understood by most sixth-grade pupils, its AI replication proves elusive. Particularly, compared to general or factual knowledge like {\em Paris is in France}, there are three unique challenges:
\begin{enumerate}
    \item Scale and Variety: Commonsense knowledge covers not only knowledge about enormous entities in this world themselves, but also diverse eventualities
    involving them in our lives, with an exceptionally long tail \cite{DBLP:journals/jair/Davis17} that modern CKGs struggle to cover, as mentioned in Section~\ref{Sec:intro}. 
    \item Defeasibility: Unlike definite facts from factual knowledge that is always true, each piece of commonsense knowledge is known as a \textit{factoid} \cite{DBLP:conf/aaai/GordonDS10} that is only \textbf{plausible} or \textbf{typically true} according to human intuition before logical reasoning \cite{choi2022curious}. They can be false under specific circumstances or even conflict with each other. For example, it is possible to find a dog that is larger than a person, and when a person gets an invitation, either accepting it or rejecting it is totally plausible. 
    \item Implicity: 
    People tend not to explain the knowledge that is held by all parties in a conversation \cite{grice1975logic}. For example, people may say, ``There is a fire, so someone calls firefighters,'' without explaining the firefighters' job. Therefore, usual, trivial, or certain knowledge is rarely stated in human language. Due to this phenomenon of \textbf{reporting bias} \cite{DBLP:conf/cikm/GordonD13}, induction only from human language, even with strong neural models, will fail to fully capture commonsense in humans' minds \cite{DBLP:conf/coling/ShwartzC20}.
\end{enumerate}

\subsection{Commonsense knowledge graphs}
\label{Sec:ckg}

Below we introduce ATOMIC, the CKG we choose to perform conceptual induction upon, as well as the source of taxonomic knowledge we use, both holding a number of advantages, making them for our goal.

\begin{figure}[t]
\centering

\includegraphics[width=\textwidth]{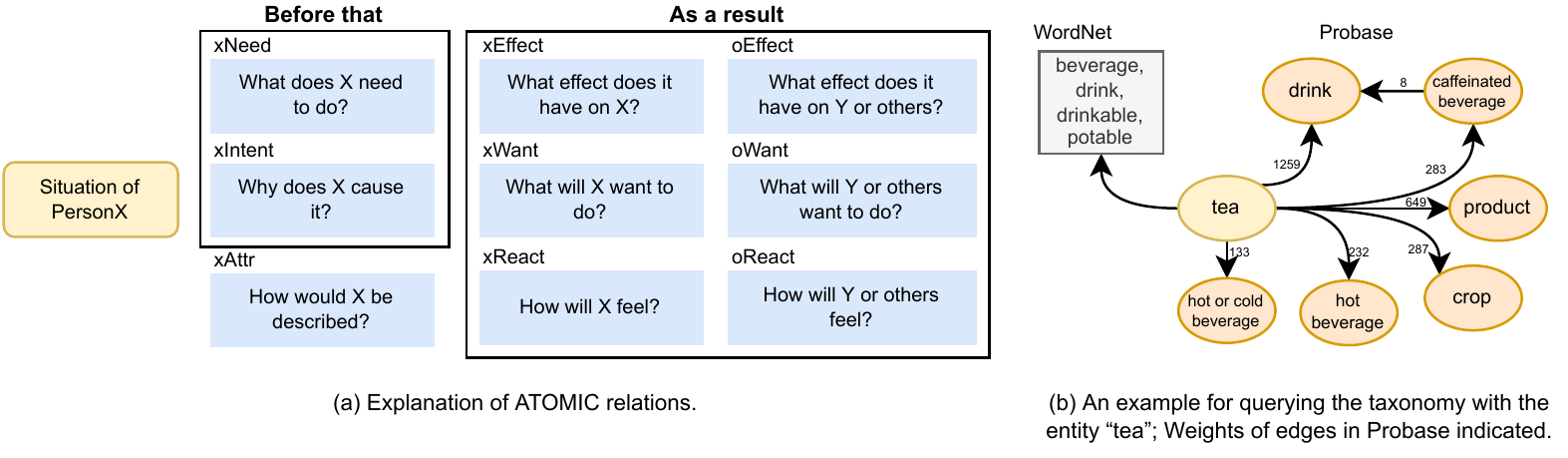}
\caption{Explanations and examples for knowledge graphs used in our methods.}
\label{Fig:atomic}
\end{figure}

\paragraph{Current CKG: the ATOMIC approach}
\label{Sec:ATOMIC_approach}
Knowledge graph (KG) is the primary form of machine knowledge representation. 
A KG, denoted as $\mathcal{G}=\{\mathcal{V}, \mathcal{E}, \mathcal{R}\}$, is a graph consisting of a set of nodes $\mathcal{V}$, and a set of edges or triples $\mathcal{E}=\{(h, r, t) \mid h, t \in \mathcal{V}, r\in \mathcal{R}\}$. Typically, nodes in a KG represent things of interest, while edges, typed with a relatively small and pre-specified set $\mathcal{R}$, denote the relation between the head $h$ and tail $t$.
KGs for factual knowledge like DBPedia \cite{DBLP:conf/semweb/AuerBKLCI07dbpedia} typically contain facts about real-world entities as heads and tails, like $\langle$\textit{h}: Paris, \textit{r}: AtLocation, \textit{t}: France$\rangle$, while earlier commonsense knowledge bases (KBs) like Cyc \cite{DBLP:journals/cacm/Lenat95CYC} and  \cite{gordon2017formal} consolidate commonsense into LISP-style logical assertions for formal logic reasoning. 

However, as mentioned above, human commonsense is after all not logical. 
In the recent interest in commonsense within the NLP community, event-centric or situational CKGs gain more popularity, which typically represent nodes in rather free-form texts. Nodes in the earlier ConceptNet are mostly entities, with only a small set of verbal nodes in phrases like ``eat food'' and ``go to a gift shop'' 
\cite{DBLP:conf/aaai/SpeerCH17}, while ATOMIC is the very first large-scale event-centric textual CKG \cite{DBLP:conf/aaai/SapBABLRRSC19}.
As shown in Figure~\ref{Fig:atomic} (a), ATOMIC covers head events regarding everyday situations of an unspecified PersonX, which are then linked to tails with a rich set of nine predefined relations of various subjective and objective causes and consequences of the event. 
The head events are harvested from various corpora 
by selecting common verbs and their arguments. Subjects of the event are normalized to PersonX, and other mentioned persons as PersonY and PersonZ. Events with diverse and infrequent arguments are conflated using wildcards like “PersonX eats \_”. Then crowd-sourced workers are asked to compose possible tails of each relation, so that reporting bias is alleviated. 

The ATOMIC paradigm relies on humans to annotate commonsense that is \textit{plausible}, which is subjective and intuitive. Different from formal logic, it better aligns with the unique characteristic of commonsense knowledge and addresses the challenges above, as thoroughly discussed in \cite{choi2022curious}. Moreover, textual representations can be easily incorporated into neural models with strong semantic understanding capabilities, to be applied to various downstream scenarios. All these factors support our choice to focus on ATOMIC. 

\paragraph{Taxonomy} As a specific type of KG, a taxonomy represents 
\textbf{taxonomic knowledge} with nodes being a set of concepts $\mathcal{C}$, connected by the \textit{is-a} or hypernym-hyponym edges, which is essential to our study. 
A widely used taxonomy is WordNet \cite{miller1998wordnet}, which organizes senses of English words into 117K \textit{synsets} designated by linguistic experts, and builds a tree-like hierarchy of \textit{is-a} relations between the synsets. Although it enjoys the finest quality, a hierarchy is too strict in reality (to be discussed in Section~\ref{Sec:conceptualization}), the taxonomy has limited coverage, and they are often too coarse-grained to differentiate specific concepts. 

Therefore, we mainly use Probase \cite{wu2012probase} as our source for taxonomic knowledge. Harvested from the Web corpora, Probase contains 2.7M nodes as \textit{concepts}, linked by 4.5M \textit{is-a} or concept-instance pairs. As shown in Figure~\ref{Fig:atomic} (b), concepts in Probase are not represented in manually disambiguated senses, but more variable and fine-grained multi-word terms in natural language, e.g. \textit{caffeinated beverage}. The \textit{is-a} edges are not black-and-white assertions but backed by the number of evidence from text corpora. Furthermore, Probase is not a strict tree-like hierarchy but a complex network, which is even not acyclic. A concept can be viewed as the instance of (i.e. conceptualized to) more abstract super-concepts in multiple ways due to the complexity of word meanings and usages, which is more realistic and better imitates human cognition. 
In this way, larger-scale, more fine-grained, and more realistic taxonomic knowledge is introduced.

\subsection{Language models for commonsense knowledge}
\label{Sec:commonsense_modeling}

Significant progress has been made in NLP thanks to recent PTLMs from BERT \cite{devlin2018bert} to GPT series \cite{radford2018improving,radford2019gpt2,brown2020language}, 
which contains rich commonsense knowledge learned from training corpora \cite{davison2019commonsense,shi2021transformers,DBLP:conf/cogsci/WeirPD20}.
Hence they have greatly pushed forward the state-of-the-art on multiple commonsense tasks like COPA and WSC after fine-tuning \cite{DBLP:conf/iclr/WangSMHLB19}. Also, as mentioned in Section~\ref{Sec:intro}, rich textual knowledge from pretraining has been successfully leveraged to make commonsense inferences from textual events following the COMET path of fine-tuning generative PTLMs on ATOMIC-style triples \cite{DBLP:conf/aaai/HwangBBDSBC21,DBLP:conf/acl/BosselutRSMCC19}. Given these successes, PTLMs shall play an essential role in our attempt of conceptual induction on text-form commonsense. 

Nevertheless, generations from such fine-tuned models lack diversity and novelty, often semantically close to each other or training samples, and not well generalized to unseen entities and eventualities \cite{DBLP:conf/acl/WangICR21,DBLP:conf/emnlp/ZhouKLLHPR21,DBLP:conf/emnlp/DuDLL19,DBLP:journals/corr/abs-2008-05925}. 
More, there is evidence that PTLMs miss certain important aspects of commonsense \cite{ shi2021transformers,DBLP:conf/cogsci/ForbesHC19,da2019cracking,DBLP:journals/corr/abs-2203-08452}, often become too vague and over-generalized \cite{CAR}, fail on negations \cite{DBLP:journals/tacl/Ettinger20,DBLP:conf/acl/KassnerS20}, and suffer from reporting bias \cite{DBLP:conf/coling/ShwartzC20,DBLP:conf/emnlp/PaikARK21}. Particularly relevant, PTLMs are found ineffective in capturing conceptual relations between entities as well as knowledge about them \cite{DBLP:conf/emnlp/PengWHJ0L0022,DBLP:conf/acl/WuJJXT23}. All the weaknesses highlight that PTLMs are not silver bullets, and we must meticulously design the fine-tuning tasks and dataset to make the best out of PTLMs for our goal.


\subsection{Human conceptualization}
\label{Sec:conceptualization}

As mentioned above, we infer knowledge of new and unseen entities and eventualities as instances of abstract concepts by \textbf{conceptual induction} using analogy with knowledge in our mind \cite{DBLP:journals/aim/ForbusH17}. From this point of view, unlike LMs chained in the cave of texts, the Platonic forms or ideas are in our minds that transcend from instances in the mundane world. This whole paper is inspired by this process. Particularly, we make our key decision choices based on the following theories and observations on our mind and cognition:

\paragraph{Flexibility, complexity, and context dependency} Among the theories of concepts, we are particularly inspired by the knowledge theory, that we learn about a concept in integration with related knowledge \cite{murphy2004concept3}, which, as for AI, could be factoids or triples involving the concept in CKGs. This contradicts the classical or definitional theory that concepts are characterized by strict definitions and form a hierarchy \cite{murphy2004concept2}, as reflected by the idea of WordNet.
Humans will disagree on borderline cases for conceptualization, and the task of deciding on proper conceptualization can be effortful \cite{murphy2004concept1}. 
Also, the conceptualization of an entity or eventuality can be diverse and context-dependent, sometimes due to a shift of perspectives. For example, to conceptualize a whale as a mammal is useful for inferring physiological features but not ecological ones \cite{murphy2004concept7}. Another aspect of variety comes from different levels of abstractness in the hierarchy. Humans recognize concepts in a non-strict hierarchy with \textit{is-a} relation in a mixed manner. The hierarchy might be partially pre-built in our minds \cite{minsky1980k} just like the taxonomies available, but further inferred on the fly according to the context \cite{murphy2004concept7}. 
Importantly, humans do not adopt strict conceptual induction either \cite{murphy2004concept7}: we recognize that birds can fly, even if penguins and ostriches can't.
It is also noteworthy that we follow a broad definition of concepts including not only entities but also eventualities, since human cognition may process eventualities in a similar hierarchical way, as formulated in the K-line theory \cite{minsky1980k}. 

\paragraph{Basic-level concepts} Concepts in a chain of \textit{is-a} relations are not equal. There are \textit{basic-level concepts} (BLCs) of intermediate abstractness which humans tend to use to refer to instances \cite{DBLP:journals/jis/Hajibayova13}. For example, when seeing a large black bulldog, people more often shout ``There is a dog'' not ``There is a large black bulldog'' or ``There is an animal." BLCs lead to an additional aspect of reporting bias, which we term \textbf{concept bias}, that even for human-annotated sentences as in ATOMIC, instances are more often represented by the words for BLCs. Nevertheless, finding BLCs is already a challenging task. More importantly, non-BLCs are also useful: Subordinate concepts carry more specific information about the particular target entity or eventuality of reasoning, while the more abstract superordinates have broader coverage \cite{murphy2004concept7}. 
For example, when inferring the health impact of drinking Dr. Pepper, it is reasonable to conceptualize it as the impact of drinking [sugary beverage], certainly a non-BLC, and infer that it will lead to weight gain, based on possible known impacts from drinking soda, Frappuccino, etc. 

\paragraph{Word meaning} 
We are particularly inspired by referential semantics \cite{sep-meaning}, which posits that the meaning of language is what it refers to in a (possibly) real world. When I narrate ``I get a cold,'' ``I'' refers to a specific person, and ``cold'' a specific phenomenon of disease happens to me at a certain time point. 
We are also inspired by event semantics \cite{davidson1967logical} that a clause refers to an event, an existence that takes place in the world at a certain time.
Hence the whole sentence refers to a specific event of ``I'' getting a ``cold'' in the real world.
In this way, nouns or clauses in language can be viewed as references to specific instances of entities or eventualities in the real world, yet to be conceptualized into the abstract existence in our mind.

\subsection{Implications}
\label{Sec:intuition}
The findings above reveal the complexity of human conceptualization, similar to human commonsense. There are earlier methods for formal logical conceptual reasoning on a strict hierarchy like First-orderized Cyc \cite{ramachandran2005first}, theories on abstraction like \cite{DBLP:journals/ai/GiunchigliaW92}, and default inheritance reasoning \cite[\S3.1]{davis2015commonsense}. While following the spirit of ATOMIC, we believe that conceptualization should be based on annotated text-form plausible situations, contextualized neural LMs, and flexible \textit{is-a graphs} like Probase, upon both entities and eventualities. Also, both BLCs and non-BLCs, as represented by the rich fine-grained concepts in Probase, should be taken into account. We also believe that we shall view the existing ATOMIC triples as textual descriptions of plausible situations, and process the more abstract knowledge after conceptualization in a distinct way.

\section{Task formulation}
\label{Sec:definition}

In this section, we introduce the conceptual induction framework by first specifying the goal of collecting a distinct set of abstract knowledge, and then elaborating on the complexity of event and triple conceptualization. In the end, we propose our pipeline for machine conceptual induction to meet our demands.

\subsection{Situational vs abstract knowledge}

\paragraph{Limitation of situational knowledge}
An early attempt to introduce conceptualization to CKGs \cite{DBLP:conf/eacl/DurmeMS09} pointed out that then CKGs extracted from text corpora are primarily restricted to those directly existing in the world. Following this perspective and the discussion in Section~\ref{Sec:intuition}, we consider current \textbf{situational CKGs} as an excerpt of the world that picks numerous textual descriptions of actual possible situations plus their causes and consequences, ATOMIC being an example with heads extracted from corpora. 
For instance, an ATOMIC triple about ``a person gets a cat, then the person will buy cat food'' 
can be viewed as a textual description of a real situation and the consequent action possibly happen to an actual person and actual cat. 
The validity of an event or triple is tacitly defined as the plausibility of the situation in real life according to the annotator. The more plausible is a situation in the real world, the more valid is the corresponding event or triple in the CKG. 
That is to say, $p(triple) \propto p(situation)$, $p$ representing the validity or plausibility. 

They do reflect plenty of commonsense knowledge generalizable to broader scenarios, since the events are often prototypical and in lack of more specific modifiers. ATOMIC goes one step further by normalizing persons and sometimes objects. 
Hence, humans can implicitly understand the inferences from the head event ``PersonX gets a cat'' as general knowledge about a class of various events like ``Alice gets a black cat.''  
However, it is elusive for the machine to determine to what extent will an ATOMIC event or triple generalize, say,  
whether the ``cat food'' inference applies to, e.g., ``My neighbor adopts a ragdoll'' or ``The vet gets a sick cat to treat''
, given the tails of that head event in a situational CKG like ATOMIC.
Without explicitly inducing a distinct set of abstract knowledge, generalization to different entities and eventualities can be reduced to be circuitous, superficial, or oversimplified, as exemplified in Section~\ref{Sec:intro}.

\paragraph{Abstract knowledge}
\label{Sec:abs_knowledge}
In contrast, we pursue \textbf{abstract knowledge} explicitly \textit{above} the realm of \textbf{situational knowledge}. We also represent it by textual events and triples, but characterized by the concept within. For example, an abstract event can be ``PersonX drinks [tea],'' with a concept ``[tea]'' indicated by the brackets. Instead of representing an event on its own, it represents a \textit{class} of instance events, such as ``PersonX drinks a cup of tea,'' ``PersonX drinks black tea with milk,'' ``PersonX drinks earl grey,'' or just ``PersonX drinks tea,'' which can be all \textit{possibly} found in ATOMIC. Similarly, an abstract triple given an abstract event represents a class of triples, or inferences on a class of instance events, e.g., $\langle$\textit{h}: PersonX drinks [tea], \textit{r}: xNeed, \textit{t}: boil water$\rangle$, which reflects the precondition on the class of events mentioned above.

More formally, an \textbf{abstract CKG} to represent abstract knowledge is defined as $\mathcal{G}_a=\{\mathcal{V}, \mathcal{T}, \mathcal{C}, \mathcal{V}_a, \mathcal{E}_a, \mathcal{R}\}$. In addition to the vertices $\mathcal{V}$ and relations $\mathcal{R}$ as in Section~\ref{Sec:ckg} in standard CKGs, there is a set of templates $\mathcal{T}$. Each template $t \in \mathcal{T}$ is a string with a slot in it, like ``PersonX drinks \_." 
We then leverage a predefined textual concept set $\mathcal{C}$, e.g. the Probase one. 
After that, we can create a set of abstract vertices $\mathcal{V}_a$ by pairing some $t \in \mathcal{T}$ with some $c \in \mathcal{C}$ like [tea] to form a $v_a=(t,c)$
\footnote{A template is similar to a logical sentence with a free variable, which could be assigned as a concept $c$. However, as free-form text, it is elusive to further analyze the meaning of $t$ and its interaction with $c$ in a formalized way beyond simply pairing them up as $(t,c)$ or possibly $t(c)$. Hence we choose to follow the ATOMIC paradigm and focus on a non-formal NLP perspective.
}: ``PersonX drinks [tea]."

Although both heads and tails in ATOMIC can be conceptualized, we discover that tails in ATOMIC are less complete as sentences and much more noisier to parse and analyze for conceptualization. 
To reach better quality for our abstract knowledge, 
we limit the abstract vertices to head events. That is to say, $\mathcal{V}_a$ is a set of abstract head events, while the set of tails is kept as-is. In this way, the abstract triples are formed as $\mathcal{E}_a=\{(h, r, t) \mid h \in \mathcal{V}_a, t \in \mathcal{V}, r\in \mathcal{R}\}$.

\paragraph{Validity of abstract knowledge}
Following this definition, the validity of an abstract event or triple is determined by the plausibility of the class of instances behind it. More formally, for an abstract event $v=(t, c)$, considering the instance $c$,  $p(v)$ depends on $p\bigl( (t, i) \mid isA(i, c) \bigr)$. Similarly, $p\bigl(\langle h, r, t\rangle\bigr), h=(t, c)$ depends on $p\bigl(\langle h', r, t\rangle\bigr)$ in which $h' = (t, i)$ and $isA(i, c)$. \ul{We consider an abstract event or triple as valid, if its instances are \textbf{typically} plausible}, following the spirit discussed in Section~\ref{Sec:intuition} to reflect human cognition. We don't need to (and are also unable to) examine all the instances of a mentioned concept to decide if a sentence makes sense.
After all, we accept that birds can fly, and that we need to boil water before drinking tea, even though we are clearly aware of the existence of penguins and bottled tea. 

Notably, under this criterion, an abstract event can be valid even if it is weird or counter-intuitive when viewed as a natural sentence or ordinary situational knowledge found in the original ATOMIC. The reason is that the abstract event represents a class of events, and the concept in the event can be rather fine-grained and distinct from BLCs we typically use in our language. Examples include ``PersonX drinks [natural beverage],'' ``[relaxing event]'' (if we conceptualize the whole event),  ``PersonX spends [time interval] reading,''  and ``PersonX is [technical job],'' etc., which are valid based on their instances, not themselves.

\subsection{Levels of conceptualization}

Now that the target is specified, the next issue is how to acquire the desired abstract knowledge.
Similar to humans, we induce possible abstract knowledge upon instances we encounter, which can be found in situational CKGs in the form of events/triples. For this induction, we carry out three levels of conceptualization, upon 1) each entity/eventuality (as an instance of a concept); 2) each event (as an instance of an abstract event); and 3) each triple (as an instance of an abstract triple), as elaborated below:

\paragraph{Entity/eventuality conceptualization}
\label{Sec:es_conceptualization}

Within each textual ATOMIC event, we may identify some entities and eventualities as \textbf{conceptualization candidates}. 
Specifically, each ATOMIC event as a whole is also an eventuality.
We consider all of them as instances of some concepts $c \in \mathcal{C}$.
For example, within the event ``PersonX drinks coca cola,'' the candidates we may find include the object ``coca cola'' and the whole event itself.\footnote{Since ATOMIC is already generalized across unspecified persons, we leave ``PersonX'' untouched.} 
Then, by 1) finding the concepts within $\mathcal{C}$ that match the candidate; and 2) utilizing known taxonomic relation between concepts, we can determine which concept the candidate is an instance of. For example, the object of the specific ``coca cola'' is an instance of (i.e. \ul{conceptualized to}) the concepts 
``[coca cola],'' ``[beverage],'' ``[sugary beverage],'' ``[phosphate containing beverage],'' etc., while the whole event can be conceptualized to 
``[drinking beverage],'' ``[relaxing event],'' ``[action],'' etc. In this case, the unspecified PersonX is implicitly considered as the semantic agent or initiator of the eventuality.

As discussed in Section~\ref{Sec:conceptualization}, we want to consider possible concepts with diverse granularity and perspectives to best reflect the flexibility of human conceptualization. To capture such great flexibility in a practical way, we choose to combine the nodes in WordNet and Probase as our concept set $\mathcal{C}$ and use their set of is-a relations, so that we can achieve the best inclusiveness. 

\begin{figure}[t!]
\centering
\includegraphics[width=\textwidth]{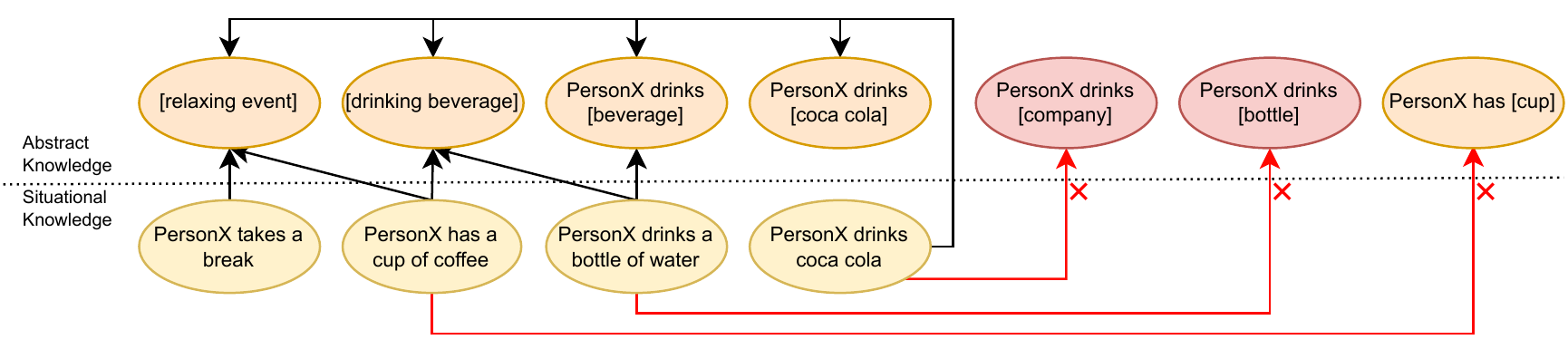}
\caption{Examples of events in situational and abstract knowledge, linked by the edges of \textit{event conceptualization}. Invalid abstract events of ``PersonX drinks [company]'' and ``PersonX drinks [bottle]'' are marked red; invalid edges like the one between ``PersonX has [cup]'' and ``PersonX has a cup of coffee'' are marked red with a cross. 
}
\label{Fig:concept_samples}
\end{figure}

\paragraph{Event conceptualization}
 By replacing each conceptualization candidate in the text with 
 a concept, abstract events can be formed. Figure~\ref{Fig:concept_samples} exemplifies this \textbf{event conceptualization} process, forming a bipartite graph between the abstract events and the original ATOMIC ones with many-to-many edges, as:
\begin{itemize}
    \item An entity or eventuality can be conceptualized with different levels of abstractness, such as ``PersonX drinks coca cola'' as ``[drinking coca cola],'' ``[drinking beverage],'' or ``[action]."
    \item An entity or eventuality can be conceptualized in different perspectives, such as ``coca cola'' as ``[sugary beverage],'' ``[phosphate containing beverage],'' or ``[iced drink],'' not in a strict hierarchical taxonomy.
    \item An event can be conceptualized into various abstract events of different template by conceptualizing different entities or eventualities appeared in it.
    \item Different events can be conceptualized into the same abstract event.
\end{itemize}

Notably, conceptualization of a candidate in an event doesn't automatically grant a valid event conceptualization, unless it falls under the class represented by the formed abstract event,
as shown in Figure~\ref{Fig:concept_samples}:
\begin{itemize}
    \item Coca Cola is a [company], but ``PersonX drinks [company]'' derived from ``PersonX drinks Coca Cola'' doesn't make sense.
    \item ``PersonX has [cup]'' is a valid abstract event representing a class of events like ``PersonX has a cup,'' ``PersonX has a beautiful tea cup,'' etc., but not really ``PersonX has a cup of coffee."
\end{itemize}

In both cases, the candidate entity doesn't really refer to an instance of the corresponding concept: ``PersonX \textit{has}'' here means ``PersonX \textit{drinks},'' 
and what PersonX actually drinks is the beverage, not the company or the cup. Therefore, we require that an abstract event can only represent the instance events that, the corresponding entity/eventuality is indeed an instance of the concept under the context, and hence the instance event being semantically covered by the abstract one. To the extreme, ``PersonX drinks [company]'' is considered invalid as it may not represent any actual event. This criterion mimics how we actually conceptualize in a contextualized way, but nevertheless adds to the complexity of the problem and demands semantic understanding of the event to determine if an event conceptualization is valid.

\begin{figure}[t]
\centering
\includegraphics[width=0.9\textwidth]{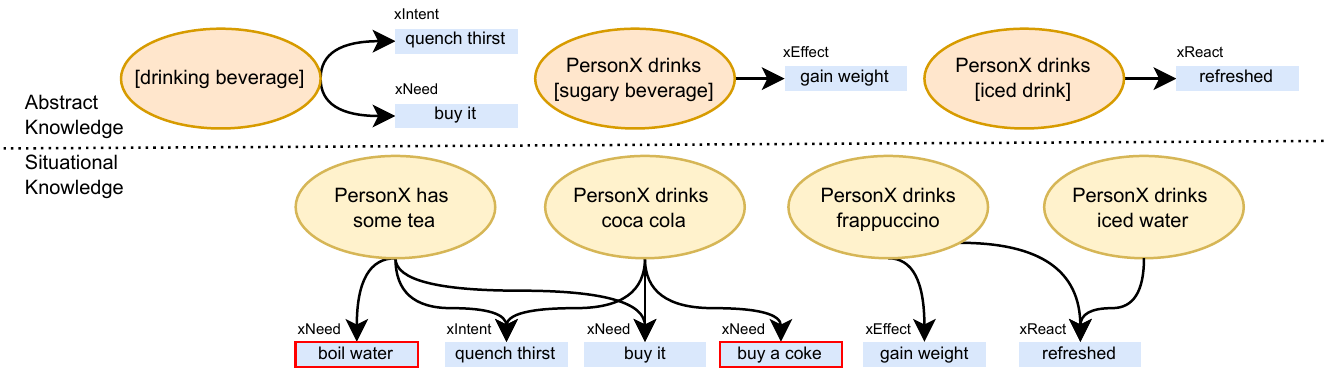}
\caption{Examples of triples in situational knowledge, as well as triples in abstract knowledge inferred from them. Inferences like ``xNeed: boil water'' that are not suitable for the corresponding abstract event are marked in red.
}
\label{Fig:concept_samples_triples}
\end{figure}

\paragraph{Triple conceptualization}
Using the triples of the instance events, we then find the inferences that are typically plausible for this class of events to form or induce abstract triples, as exemplified in Figure~\ref{Fig:concept_samples_triples}: the abstract triple $\langle$\textit{h}: [drinking beverage], \textit{r}: xNeed, \textit{t}: buy it$\rangle$ can be induced from the same inference upon instance events ``PersonX has some tea'' and ``PersonX drinks coca cola." Note that this is an incomplete induction: Although this inference is also valid for ``PersonX drinks Frappuccino,'' the original CKG may fail to cover that. On the other hand, some inferences on the instance event like ``xNeed: boil water,'' are not typically held for the abstract event and will not be included as abstract triples.

\begin{figure}[t!]
\centering
\includegraphics[width=0.9\textwidth]{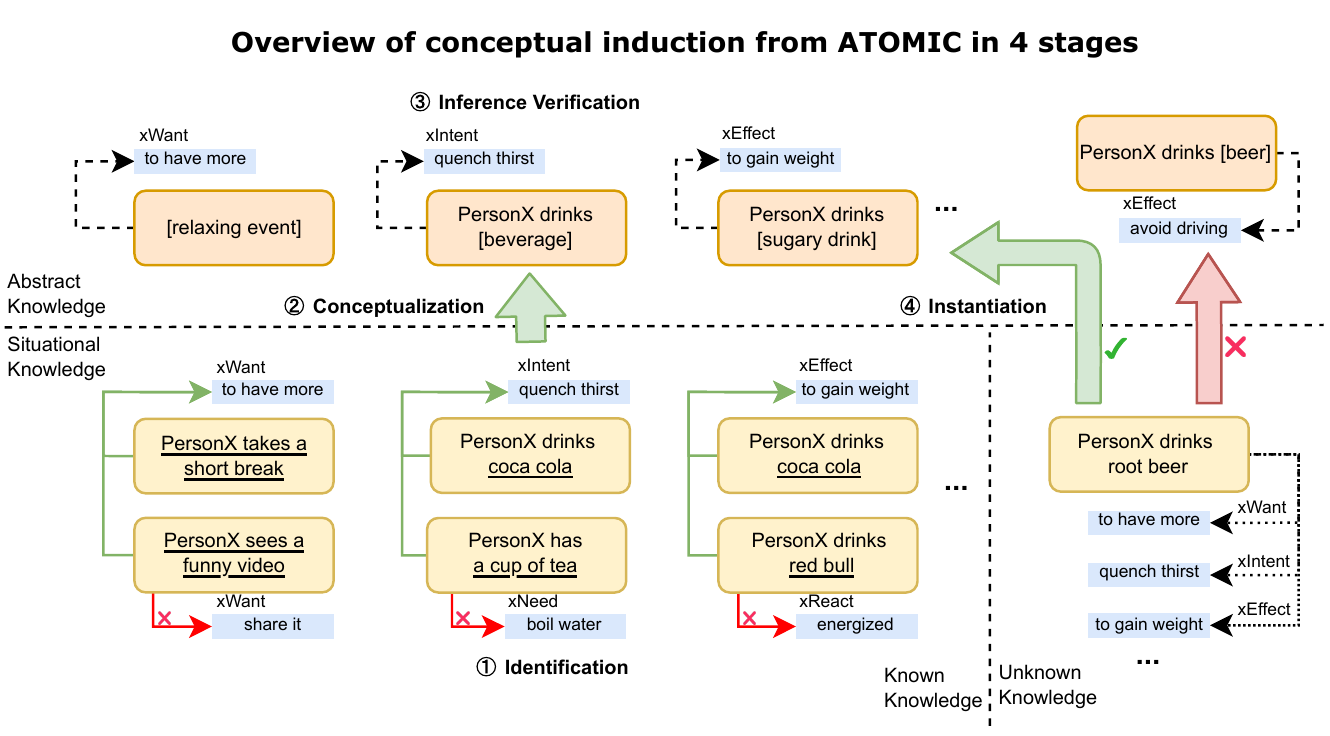}
\caption{An overview of our 4-stage formulation of conceptual induction using examples. 
By identifying candidates (underlined) in the events and conceptualizing them into concepts (e.g. ``coca cola'' and ``a cup of tea'' into ``[beverage]''), abstract events are built.
Next, inferences typically held on the instance events are collected as abstract triples, otherwise indicated in red lines with crosses.
The whole procedure mimics how we learn abstract knowledge from instances. 
When a new event like ``PersonX drinks root beer'' comes, we can conceptualize it into abstract events by reusing stage 1\&2, while incorrect conceptualizations like ``PersonX drinks [beer]'' are excluded. We can then make inferences of this event using corresponding abstract triples.}

\label{Fig:concept_idea}
\end{figure}


\subsection{Summary}
Given the discussions above, we imitate human conceptual induction in four stages as shown in Figure~\ref{Fig:concept_idea}:
\begin{enumerate}
    \item Identification: Given an original event, identify the entities or eventualities as candidates of conceptualization, as indicated by underlined words.
    \item Conceptualization: Given an identified candidate in an event, correctly conceptualize it into some concepts in $\mathcal{C}$ to form abstract events.
    \item Inference Verification: Considering all the inferences on instance events of each abstract event, determine which of them are typically held, so as to form the abstract triples correctly.
    \item Instantiation: When encountering any new event, conceptualize it into abstract events, so that we can leverage the corresponding abstract triples to make inferences about it. 
\end{enumerate}

For stage four, when queried about a new event presumed to be valid, finding its abstract events is identical to the first two stages. As the abstract triples or inferences are typically valid among the instance events including the queried one, we may assume that they are quite plausible to be held to the queried one, or at least valuable for further inspection. Therefore, the task to replicate conceptual induction highlights the first three stages
to acquire a comprehensive set of abstract knowledge in the form of abstract events and triples, ready to be instantiated whenever unknown events are encountered. 





\section{Methods}
\label{Sec:methods}
\begin{figure}[t]
\centering
\includegraphics[width=0.8\textwidth]{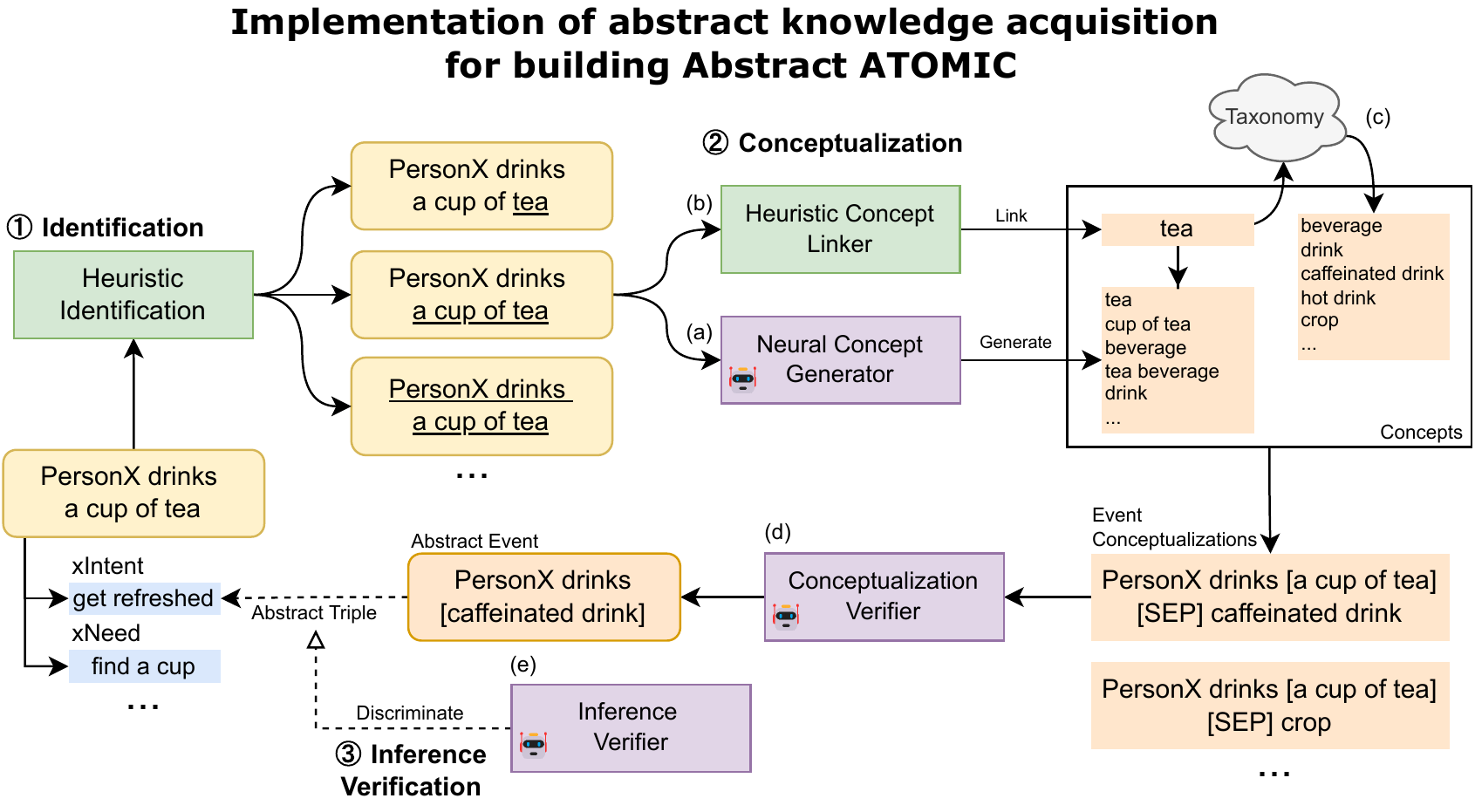}
\caption{Our implementation of acquiring abstract knowledge from ATOMIC, demonstrated by an example. We first identify entities/eventualities by heuristic rules. Then the identified candidates go through two pathways: (a) A transformer-based \textit{neural concept generator} to predict possible conceptualizations, and (b) heuristic rules to link to concepts in $\mathcal{C}$, to be further directly conceptualized using \textit{is-a} relations in the taxonomy (c). Next, we form the event conceptualizations and use a similar \textit{conceptualization verifier} (d) to determine their validity, forming abstract events. Finally, possible abstract triples are collected from instance triples and verified by an \textit{inference verifier} (e).
}
\label{Fig:method}
\end{figure}

Below we present how we implement each step in abstract knowledge acquisition, as outlined in Figure~\ref{Fig:method}.

\subsection{Identification}

First, we need to correctly
identify conceptualization candidates from words or phrases in a text-form event. 
However, simply matching the words leads to errors due to word inflection and ignorance of boundaries between syntactic constituents. For example, in ``She gives her pet food,'' ``pet food'' can be incorrectly identified as a separate entity. 
Instead, we first parse the event with spaCy \cite{honnibal2017spacy}, and then work on the dependency tree to inspect each subtree (i.e. \textbf{constituent}) to decide if it is a possible nominal candidate (entity) or predicative candidate (eventuality) according to linguistic features of the head (root) word, including the surface form, lemma, coarse or fine-grained part-of-speech (POS), and dependency tag (DEP). 

Unfortunately, both the original data and the predicted linguistic features contain errors, making it imprecise to rely on a single type of feature. 
Hence we curate a set of error-tolerant rules for identification. We iteratively test and improve them on the ATOMIC data, aimed at collecting most of the valid candidates while avoiding errors, to make the results useful for future conceptual induction with our best effort.

To be more specific, a constituent is considered an entity/nominal candidate if:
\begin{itemize}
    \item Its DEP indicates a nominal subject (\textit{nsubj}), object (\textit{iobj}, \textit{dobj}), or nominal attribute (\textit{attr}); or,
    \item It has nominal POS (like \textit{NOUN}, \textit{PROPN}, \textit{PRON}) or it is a gerund, and it has potentially nominal DEP (such as complement).
\end{itemize}

A constituent is considered an eventuality/predicative candidate if: 
\begin{itemize}
    \item Its DEP indicates a predicate or clause, like \textit{ROOT} (for a whole event), \textit{acl}, \textit{advcl}, \textit{relcl}, etc.; or,
    \item It has predicative POS (like \textit{VERB} and \textit{ADJ}), and potentially predicative DEP (such as complement).
\end{itemize}

In addition, for conjuncts connected by conjunctions (e.g. ``coffee and tea''), if one of them is identified as a nominal or predicative candidate, others will be included as the same type of candidate. It is possible for a constituent to be marked as both nominal and predicative, like gerunds, which is by design so that we can find the most accurate conceptualizations in the next stage. A constituent with DEP like \textit{cc}, \textit{advmod}, \textit{prep}, etc. doesn't correspond to an entity or eventuality, hence it is deemed not a candidate.
After that, a set of candidates is collected, represented as a set of subtrees, to be verified in the next stage.

\subsection{Conceptualization}

We then conceptualize candidate entities/eventualities using the concept set $\mathcal{C}$. This also verifies the previous stage, as we consider those fail to match any concept in $\mathcal{C}$ as invalid, or at least not usable. As mentioned in Section~\ref{Sec:es_conceptualization}, we first need to match a candidate to known concepts, termed \textbf{concept linking}. 


This is nevertheless a non-trivial task: In both ATOMIC events and natural language, entities/eventualities can be modified in diverse ways. The corresponding fine-grained concept may be absent in the taxonomy, which renders it impossible for further use by \textit{is-a} relations. For example, when conceptualizing the object in ``PersonX has a lovely small siamese cat,'' we surely need to ignore the determiner, but still ``[lovely small siamese cat]'' is certainly not in WordNet or Probase. Instead, only ``[siamese cat]'' can be found in Probase. 
Hence 
we define the concept linking task as \ul{finding the concept(s) from $\mathcal{C}$ that describe or cover the candidate in the most precise way}. 
When linking an eventuality, PersonX will be implicitly considered as the semantic agent of the concept, e.g. [having pet].
In this way, we try to find the most precise, low-level, or informative concepts for the mentioned entity/eventuality, with the most potential for further conceptualization into more abstract concepts.

However, this remains a challenging goal due to the variability of language. The actual concept that a constituent refers to may not match the headword or even any word within it, but depends on its meaning as a whole. For example, the constituent may involve light verbs (e.g. ``take a shower''), transparent constructions \cite{DBLP:conf/naacl/MeyersRMSZYG04} (e.g. ``a cup of tea,'' ``the city of New York''), or require deeper semantic understanding (e.g.  ``[cohabitation]'' for ``PersonX lives with her boyfriend''). 

Utilizing known \textit{is-a} relations is also elusive if we try to further conceptualize the linked concept. 
Homonymy like ``bank'' as river bank or savings bank causes some ambiguity, but more serious issues come from polysemy like ``coca cola'' as the beverage or the company, as in Figure~\ref{Fig:concept_samples}. It may even go beyond lexical ambiguity and require commonsense to handle. For the same real-world referent, conceptualization can be different due to the context. 
For example, as for 
``drink [coca cola] after eating spicy salsa,'' ``[coca cola]'' can be conceptualized into ``[iced drink],'' but in ``coca cola gains you weight'' it is no longer the case.

Therefore, the complexity of conceptualization surpasses the capacity of mere heuristic rules and requires a more complicated pipeline: We first develop heuristic linking rules to propose possible concepts. Upon these results, we leverage \textit{is-a} relations to obtain some more abstract ones. Another route is to train a neural concept generator to directly predict the concepts of the candidate. In all the three cases, the results can be imprecise. Hence we follow the principle similar to the identification stage, considering them as ways to produce good candidates with best efforts, to be later verified by a neural model, as elaborated below. 


\begin{table}[!p]
\centering
{\def\arraystretch{1.6}
\begin{tabularx}{\textwidth}{p{2.7cm}>{\RaggedRight\arraybackslash}p{3.3cm}lp{6.3cm}}
\toprule
Type                & Example & Concepts & Method\\ \midrule
Word & PersonX finds \underline{some cats} & cat & Directly use the headword, possibly lemmatized\\
{Compound/} Phrase & PersonX sees \underline{many stray cats} & stray cat & Collect compounds or phrases in nominal candidates by word matching in the constituent, subject to inflections \\
Predicate (Verb) & \underline{PersonX drinks coffee} & drinking & Directly use the gerund of the verb \\
 & PersonX says \underline{he enjoys himself} & enjoyment & Check WordNet and NOMBANK for the noun form of the verb  \\
Predicate (Adj.) & \underline{PersonX is happy} & happiness & Same as above, for a copula with adjective complement \\ \midrule
Conjunction & PersonX sees \underline{doctors and nurses} & doctor, nurse & Use concepts from both conjuncts \\ \midrule
Nominal Candidate with Classifiers & PersonX has \underline{a cup of tea} & tea & Directly return results from the accompanied argument, if the head and the preposition form a transparent construction according to NOMBANK \\
 & PersonX sees \underline{a group of people} & group, people & If an argument is connected to another one by ``of'' but not a transparent construction, both the head and accompanied argument are used. \\ \midrule
Verb Phrase & \underline{PersonX drinks} \underline{coffee} & drinking coffee & Combine the verb with its arguments \\
Phrasal Verb & \underline{PersonX gets up late} & get up & Check WordNet for combination of the verb and one or two particle in the text \\
Light Verb & \underline{PersonX gives} \underline{a speech} & giving, speech & For light verbs like \textit{give}, \textit{take}, \textit{have}, etc., the predicand can be the actual concept\\
& \underline{PersonX goes} \underline{shopping} & shopping & Directly use results from the predicand when it is a gerund\\
Aux./Raising-to-subject & \underline{PersonX seems to} \underline{be happy} & happiness & Directly use results from the predicand for verbs like \textit{seem}, \textit{appear}, \textit{used}, etc. \\
Catenative Verb & \underline{PersonX wants to} \underline{leave} & want, leave & Use both the head (superordinate verb) and results from its complement \\
Adj. + Infinitive & \underline{PersonX is likely} \underline{to leave} & leave & Directly use results from the infinitive for a copula + adjectives like \textit{likely}, \textit{going}, \textit{about}, \textit{able}, etc. as complement  \\

\bottomrule
\end{tabularx}
}

\caption{\label{Tab:word_cases} Methods to determine the linked concept(s) under various cases in the heuristic pathway, plus examples of the results, target candidates underlined. The algorithm is often recursive that performs linking based on the linking results of some sub-constituent/subtree(s). WordNet linkings are limited to nodes with matched POS depending on whether it is a nominal or predicative candidate. Different inflections of the resultant concept are all included to be inspected later.}
\end{table}

\subsubsection{Heuristic approach}
\label{Sec:rule_abs}

\paragraph{Heuristic concept linking}
In addition to simply using the headword of each candidate constituent, we also consider a number of syntactic and semantic phenomena in the constituent, and develop a recursive algorithm trying to extract the actual and precise concept based on the parsed (sub-)tree, as listed in Table~\ref{Tab:word_cases}. 

Handling predicative candidates is especially challenging, as Probase is relatively in lack of verbal nodes compared to WordNet. We nevertheless attempt to find and use the nominal form of predicative candidates by combining WordNet and NOMBANK \cite{DBLP:conf/naacl/MeyersRMSZYG04} so as to link the candidate to Probase with richer taxonomic knowledge. 
More semantic nuances like negations and modalities are left to be handled by later verification. Additional explanations of more complicated cases are available in \ref{Appendix:ec_rules}. 

\paragraph{Further conceptualization}
Given text-form linked concepts as the most precise and low-level conceptualization of the entity/eventuality, we further try to identify more abstract ones using the taxonomy. 
Regarding WordNet where nodes are not raw text but disambiguated synsets, we first feed the event to pretrained GlossBERT \cite{DBLP:conf/emnlp/HuangSQH19} (PersonX/Y/Z replaced by actual names) to determine the synset, and then pick its hypernyms.
As for Probase nodes, we pick the ten best conceptualizations ordered by mutual information, which provides relatively high-quality and relevant results. Over-general or meta concepts like ``word,'' ``noun,'' ``concept,'' etc. are excluded.


\subsubsection{Neural approach}
\label{Sec:neural_abs}
In addition to heuristics, we use neural models fine-tuned on human-annotated data to be elaborated in Section~\ref{Sec:data}. Based on PTLMs, those models are better at capturing the semantic meaning under the context for conceptualization. We carefully design the dataset aided by taxonomic knowledge, so that the models don't merely learn to induce by word co-occurrence in natural language, but also implicitly reflect taxonomic knowledge with reduced reporting bias, concept bias in particular. 
Nevertheless, neural LMs suffer from diversity and novelty issues, and the heuristic approach leverages more diverse knowledge explicitly from the taxonomy, much larger than our dataset.
Therefore, both approaches are deemed essential.

To be more specific, we use a paired event conceptualization dataset $S_a=\{(x, y)\}$, each sample formed by an event conceptualization $x$ and its binary label $y$, where each $x=(h,c)$ contains a head event $h$ with the candidate identified, plus a concept $c$. The dataset is split into \textit{trn}/\textit{dev}/\textit{tst} subsets according to the original ATOMIC partition, in which the head events in different subsets are ensured to be dissimilar.

\paragraph{Concept generation}
We first use a GPT2-based LM to generate possible concepts for the candidate in a way similar to COMET: 
Each positive sample $\bigl((h_i, c_i), y_i\bigr)$ where $y_i=1$ is formed as a sequence of tokens $t_i=\textrm{Concat}[p_i, \textrm{[GEN]}, c_i, \textrm{[EOS]}]$, with the head $h_i$ represented in a prompt $p_i$. 
The standard causal LM loss on $c_i$ is used: 
Suppose $h_i$ plus [GEN] correspond to first $m$ tokens in $t_i$ with total $n$ tokens, the loss is
\begin{equation}
L=-\sum_{t_i} \sum_{j=m+1}^n \log P(t_{i,j} \mid t_{i, <j}).
\label{eq2}
\end{equation}
 

\paragraph{Conceptualization verification}
Finally, we combine the possible event conceptualizations from all the three sources, and feed them into a neural model as an ultimate gatekeeper to filter out those not matching the context, as exemplified in Figure~\ref{Fig:method}. As a text classification task, we fine-tune RoBERTa \cite{DBLP:journals/corr/abs-1907-11692} to predict the valid probability $D_a(x_i)$
 from the first hidden vector, using binary cross-entropy loss given the label $y_i$: 
\begin{equation}
L=-\sum_{(x_i,y_i)} \bigl(y_i\log(D_a(x_i))+(1-y_i)\log(1-D_a(x_i))\bigr).
\label{eq1}
\end{equation}


\subsection{Inference verification}
\label{Sec:induce}

Based on the abstract events we produce by the verified event conceptualizations, we can now inspect all the triples of their instance events as possible abstract events. By training a neural discriminator just like the conceptualization verifier on the validity of abstract triples, a model similar to the scorer in standard KG link prediction, 
the valid inferences on the abstract events (i.e. abstract triples) are determined.

\section{Data collection}
\label{Sec:data}
To build the neural models and to collect conceptualization data for future research, we annotate a large-scale dataset on the validity of conceptualizations on events and triples based on existing events and triples from ATOMIC, aided by Probase, expected to evade concept bias without sacrificing flexibility.

\subsection{Data preparation}
\label{Sec:clean}
Before all else, we have to first clean, filter and parse the ATOMIC data to eliminate errors in the data that will the hinder following conceptualization. This is by fixing errors, parsing the sentences, and dropping those with unresolved errors, leaving 23.6K out of 24.3K ATOMIC events for further processing. 

The next step is to pick events and entities/eventualities suitable for annotation. First, we skip all events with wildcard objects (``\_''), as they are over-generalized and prevent us from leveraging more taxonomic knowledge. All following data processing and experiments are performed on this filtered \textit{ATOMIC event subset} of 18.9K events. Then, we identify and skip 3.0K events that are idiomatic. This is because Wiktionary idioms are among the sources of events in ATOMIC, while set phrases like idioms do not follow the principle of compositionality, hence conceptualization of a particular part of an idiom is usually meaningless. Next, among the rest 15.9K events, we identify 31.1K entities and eventualities as conceptualization candidates using the rules above. This set of \textit{ATOMIC conceptualization candidates} is used as the source of Abstract ATOMIC to be discussed in Section~\ref{Sec:abs_atomic}. In the end, we try to de-duplicate the candidates by removing and merging those too similar, so as to reduce costs, improve efficiency, and introduce more diversity and representativeness in the annotation. From the 18.7K candidates in 12.0K events after de-duplication, 10,000 of them are randomly picked for annotation of their event conceptualization, identified from 8,045 original ATOMIC events. More details for data preparation are available in \ref{Appendix:clean}.

\begin{figure}[t]
\centering

     \centering
     \begin{subfigure}[b]{0.475\textwidth}
         \centering
         \includegraphics[width=\textwidth]{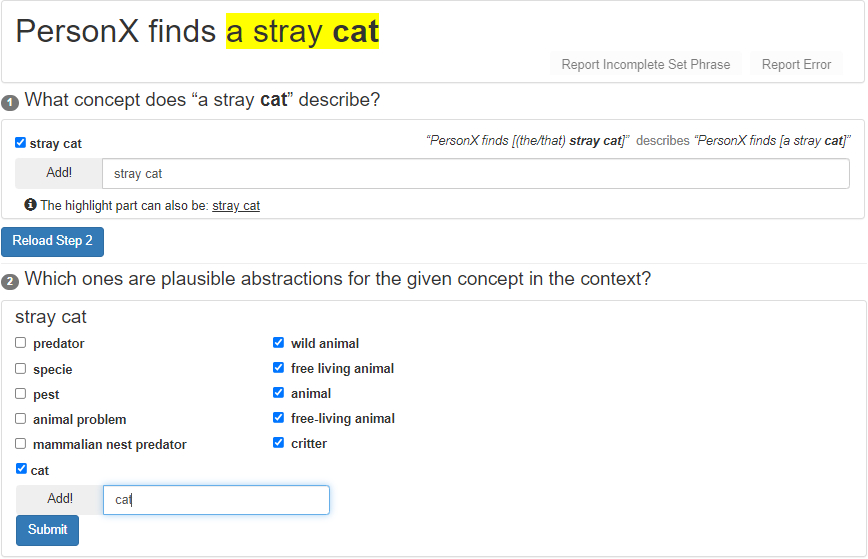}
         \caption{An example for round one annotation interface.}
         \label{Fig:annotate_a}
     \end{subfigure}
     \hfill
     \begin{subfigure}[b]{0.475\textwidth}
         \centering
         \includegraphics[width=\textwidth]{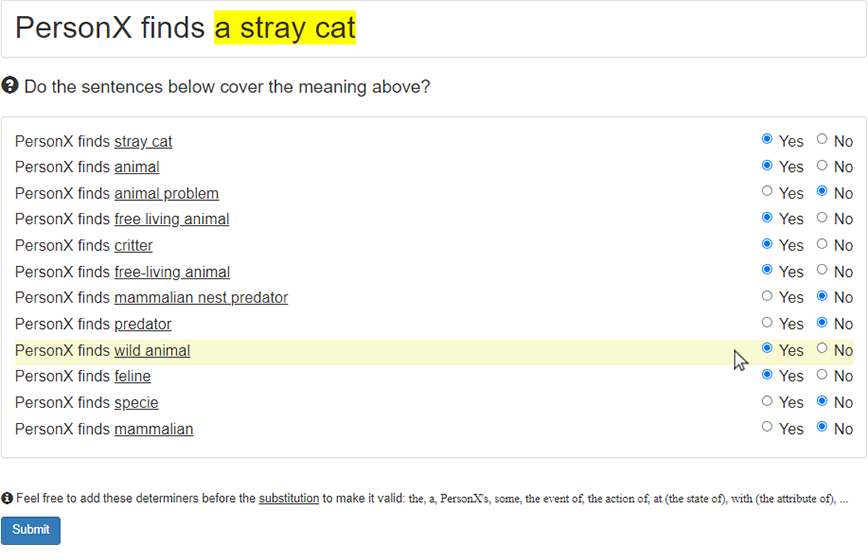}
         \caption{An example for round two annotation interface.}
         \label{Fig:annotate_b}
     \end{subfigure}
     \hfill
     \vskip\baselineskip
     \begin{subfigure}[b]{0.475\textwidth}
         \centering
         \includegraphics[width=\textwidth]{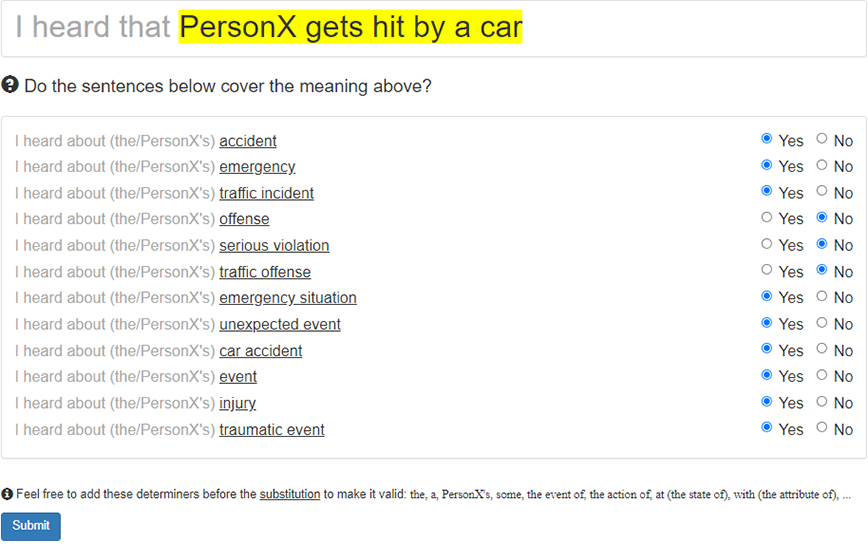}
         \caption{An example for round two annotation interface, for conceptualization of a whole event.}
         \label{Fig:annotate_c}
     \end{subfigure}
     \hfill
     \begin{subfigure}[b]{0.475\textwidth}
         \centering
         \includegraphics[width=0.8\textwidth]{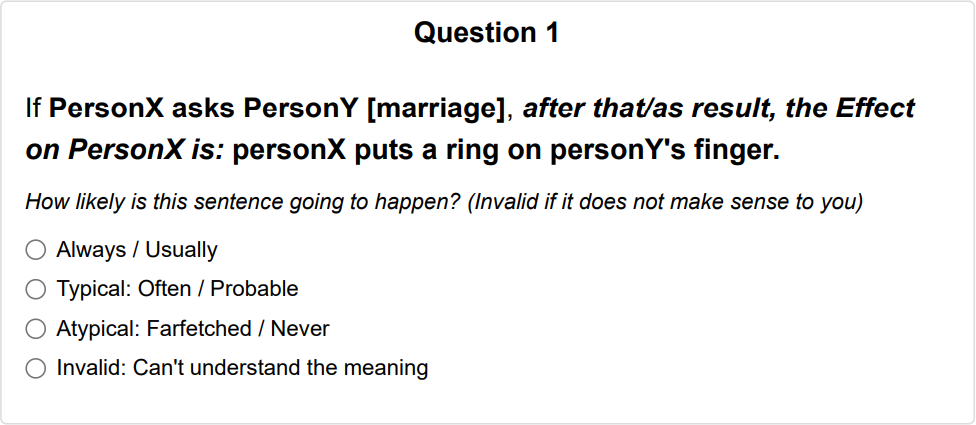}
         \vspace{3em}
         \caption{An example for a question in round three annotation interface.}
         \label{Fig:annotate_d}
     \end{subfigure}

\caption{Screenshots for the web interface in the annotation process with three rounds.
}
\label{Fig:annotate}
\end{figure}

\subsection{Crowd-sourced annotation}

We recruit human workers on Amazon Mechanical Turk for annotation. However, even with human intelligence, the task can be difficult, as discussed in Section~\ref{Sec:conceptualization}, and requires careful design, thorough guidance, stringent criteria, and annotators' painstaking work to obtain acceptable quality. We introduce the three-round annotation procedure below, while more details are available in \ref{Appendix:annotation}.

\subsubsection{Entity/eventuality-level annotation}
\paragraph{Concept linking annotation}
We ask the worker to annotate all the three level of conceptualization, starting from the entity/eventuality level upon all the 10,000 candidates, based on an interactive website as the annotation interface shown in Figure~\ref{Fig:annotate}. 
Given an event with the target candidate (entity/eventuality) highlighted, we first ask the worker to \textit{type in} some concepts that \textit{best describe} the entity as in (a). This  parallels the concept linking step. The worker can only submit a concept available in Probase, checked by the back-end instantly during typing. 
We find that workers are able to give precise or even creative responses like ``\underline{PersonX surfs the web}\footnote{The underlined part of an event indicates the target conceptualization candidate.}'' as ``internet surfing,'' ``\underline{PersonX tells PersonY's boss}'' as ``sharing information,'' and ``\underline{PersonX has always wanted to go to}'' as ``long-term goal'' or ``dream vacation'', though distinct from the expression in the sentence. Therefore the free typing approach to finding candidate concepts allows the worker to utilize their commonsense to produce rather flexible and high-quality results.

 In case an erroneous or not understandable event escapes the filtering in Section~\ref{Sec:clean}, workers should click ``Report Error." While in addition to idioms, set phrases like ``go to \underline{bed}'' do not follow compositionality as well, making conceptualization meaningless (e.g. if we conceptualize it to ``furniture'' and consider other instances like ``go to chair''). Hence they should also report when the target is part of a set phrase. 

\paragraph{Further conceptualization}
In the second step, the top ten Probase conceptualizations of each typed concept will appear below the typing section in the webpage, as in Figure~\ref{Fig:annotate} (a). The worker is asked to select the appropriate ones that \textit{cover} the meaning of the typed concept in the context. This allows the worker to consider various possible conceptualizations from the taxonomy that are often not as intuitive as BLCs, and hence alleviates concept bias mentioned in Section~\ref{Sec:conceptualization}. The worker may still type their own answers, which will then be appended to the list for that concept for future workers to choose from.

\subsubsection{Event-level conceptualization}
A second-round annotation is carried out with the goal of verifying the first-round results explicitly at the event level for better quality. Among the 10,000 candidates, those not reported as \textit{error} or \textit{set phrase component} by any annotator are used to construct event conceptualizations to be verified, leaving 8,748 targets from 7,196 different ATOMIC events. For each target, we then build possible abstract events using all the concepts annotated for it in both steps of Round 1. 
As shown in Figure~\ref{Fig:annotate} (b), we ask the worker to decide if each of the abstract events covers the meaning of the original one, i.e. if the event conceptualization is valid. Additional textual hints are given on the page when the target is a whole event as in (c). 


\subsubsection{Triple-level conceptualization}
Next, we collect possible abstract triples to annotate their validity. Specifically, we sample a subset of conceptualization candidates in a DEP-balanced way. For each corresponding head event, we include all the instance triples paired with all the derived abstract events, so as to completely depict the inferences on all the conceptualizations of an instance. We reuse the pipeline to annotate ATOMIC-like triples in \cite{DBLP:conf/emnlp/FangWCHZSH21,DBLP:journals/corr/abs-2304-10392} with an interface like Figure~\ref{Fig:annotate} (d), except that we instruct the workers to consider the validity of an abstract triple based on its instances as defined in Section~\ref{Sec:abs_knowledge}. 

\subsubsection{Quality assurance}
In all the three rounds, we go through a rigid process to recruit and educate workers. Only workers from English-speaking countries with good past records are included. Carefully written instructions along with more than 30 examples are provided for workers in each round that cover various cases in the annotation corresponding to those in Table~\ref{Tab:word_cases}. We also curate multiple sophisticated qualification tests, and inject random tests during the first 100 responses of each worker. 
Besides such gate-keeping during recruitment, quality assurance measures are also carried out during the whole annotation process: From time to time, we carefully monitor all the workers' responses, randomly select them for manual inspection, send the workers explanations when errors are found, and listen to workers' suggestions to improve the instructions and the interface. Error-prone workers are warned and then disqualified, and their annotations are discarded. 

In all three rounds, each question is annotated by five valid workers. In Round 2, an event conceptualization is determined valid if it gets $\geq$4 out of 5 votes, and invalid if it gets $\leq$1 vote. 
In Round 3, we count a positive vote if the worker responds Always/Usually or Typical, otherwise negative. With a higher possibility for the candidates to be correct, we set the valid and invalid bar to $\geq$4 and $\leq$2 votes instead. 
All annotations with the number of votes in-between are deemed \textit{indecisive} and discarded.

\subsection{Annotation results}

Statistics of the annotation are given in Table~\ref{Tab:annotation}. As for events, we retrieve 40,833 positive conceptualizations upon 7,019 ATOMIC events towards 8,545 different Probase concepts, forming 18,964 different positive abstract events. Hence each abstract event is conceptualized from 2.15 candidates on average.

\begin{table}[t]
\centering
\begin{tabular}{lcc}
\toprule
                            & Event   & Triple  \\\midrule
Total Questions             & 131,004  & 90,207   \\
Questions with Agreement    & 92,235   & 81,197   \\
Positives    & 40,833   & 65,900   \\
Positive Rate               & 44.27\% & 81.16\% \\ \midrule
Inter-annotator Agreement   & 71.4\%  & 73.0\%  \\
Manual Inspection Agreement & 87.0\%  & 86.5\%  \\
\bottomrule
\end{tabular}

\caption{\label{Tab:annotation} Statistics for the event/triple conceptualization annotation: the number of questions with agreement (i.e. not indecisive), the positive rate among agreed questions, and the agreement rate between annotators and with our manual check.}
\end{table}

Based on the positive abstract events, a total of 1,149,118 possible abstract triples are collected. Hence we have to limit the annotation to a small subset, randomly selected as mentioned above. 
In this way, 90,207 triples on 2,756 abstract events are annotated, based upon 708 entities or eventualities in 689 ATOMIC events. A total of 65,900 abstract triples are determined positive.

With a large-scale dataset collected, we then look at its quality and compute the inter-annotator agreement rate, both >70\%. Furthermore, we manually inspect 200 random samples with decisive answers from each stage, and find >85\% agreement rates with the majority vote. Besides, with a majority of triples being valid, we compute the annotation precision to identify negative samples compared to our manual inspection, which gives 70.45\%. All these results indicate that the annotation is consistent with acceptable quality.

A considerable amount of event and triple level candidates (46\% and 19\%, respectively) are annotated as negative. However, the candidate events in round 2 are proposed by annotators in round 1, and candidate triples in round 3 are valid inferences at least for some instance event. 
Therefore they are all reasonable in some sense, and can be all considered as hard negatives. The amount of such negative samples indicates the challenge of conceptualization on both levels that can not be handled by superficial induction, which highlights their value for model training. It is also noteworthy that there is an imbalance between the DEP of the annotated candidates. More details are available in \ref{Appendix:anno_stat}.

\section{Abstract KG construction}
\label{Sec:exp}

With data annotated in Section~\ref{Sec:data}, we are now able to train the models described in Section~\ref{Sec:neural_abs}. Under a supervised learning setting, models are fine-tuned from PTLMs with textual prompts. Their performances are evaluated against multiple baselines. Then we apply the three-stage abstract knowledge acquisition approach discussed in Section~\ref{Sec:methods} to ATOMIC, and harvest a large-scale abstract CKG, namely \textbf{Abstract ATOMIC}. More details of the experiments, including the prompts we use, are available in \ref{Appendix:exp_detail}.

\subsection{Conceptualization modeling}
\label{Sec:exp_abs}

\begin{table}[t]
\centering
\begin{tabular}{lcc}
\toprule
Accuracy                & \textit{dev}  & \textit{tst}  \\ \midrule
Supervised Discriminator     & 83.9 & 85.0   \\
NS  & 78.4 & 78.8 \\
Generator - Supervised       & 76.9 & 77.6 \\
\hspace{1.55cm} - GPT2 Zero-shot & 55.1 & 58.5 \\
\bottomrule
\end{tabular}

\caption{\label{Tab:stage2_models_acc} Accuracy (\%) of different conceptualization verifiers, including the \textit{Supervised Discriminator} trained on our annotated dataset, \textit{NS} trained by negative sampling, and using perplexity from a \textit{Generator} as a scorer. }
\end{table}

\begin{table}[t]
\centering
\begin{tabular}{lcccc}
\toprule
                     & \multicolumn{2}{c}{BLEU-1}  & \multicolumn{2}{c}{BLEU-2}  \\ 
                     & \textit{dev} & \textit{tst} & \textit{dev} & \textit{tst} \\ \midrule
Supervised Generator & 65.1         & 68.0         & 61.1         & 56.5         \\
GPT2 Zero-shot       & 25.0         & 20.4         & 4.8          & 2.6   \\
\bottomrule
\end{tabular}

\caption{\label{Tab:stage2_models_bleu} BLEU score (\%) of our concept generator trained on our annotated dataset, c.f. zero-shot GPT2 generations. 
}
\end{table}

Using the annotated event conceptualization validity dataset $S_a$, we can simply train the conceptualization verifier as a RoBERTa-based text classifier. While from the subset of positive samples $S_a^+$, we can train the GPT2-based neural concept generator. The models are then evaluated on the \textit{dev} and \textit{tst} subset of $S_a$, measured by accuracy or BLEU-2 score for the two types of models, respectively. In addition, we report BLEU-1 for the generator since the reference concept often contains one word only. 

In contrast, we build several baselines that do not (fully) rely on our annotated data:

1. Negative sampling (NS): As commonly used in KG completion, we build a discriminator only based on the positive subset $S_a^+$. 
For each sample in $S_a^+$ where $\bigl( x_i=(h_i,c_i), y_i=0 \bigr)$, we uniform-randomly sample five $c'_{i,k}$ from all concepts $C^+$ appeared in $S_a^+$ to build five pseudo-negative samples, forming
\begin{equation}
S_a^-=\bigl\{\bigl((h_i,c'_{i,k}), y_i=0\bigr) \mid(h_i, c_i) \in S_a^+, 1 \leq k \leq 5\bigr\},
\label{eq3}
\end{equation}
Then the discriminator is trained upon $S=S_a^+ \cup S_a^-$ similarly with binary cross-entropy loss.

2. Supervised generator: We also use the supervised generator as a discriminator by using the negated perplexity given each sample as its score, with the classification threshold tuned on \textit{dev} set. 

3. GPT2 Zero-shot: We directly feed the sample with a prompt into the pretrained GPT2 generator in a zero-shot way as a generator, or a discriminator using the perplexity score as well.

As shown in Table~\ref{Tab:stage2_models_acc} and Table~\ref{Tab:stage2_models_bleu}, both the Supervised Discriminator and Supervised Generator reach satisfying results, 
and both zero-shot models fail to produce high-quality predictions, highlighting the difficulty of our task and the necessity of our annotated data. 
Particularly, the supervised discriminator significantly outperforms the baselines, indicating the necessity of a validity dataset with negative samples annotated.

\subsection{Inference verification modeling}
\label{Sec:exp_induce}

\begin{table}[t]
\centering
\begin{tabular}{lcc}
\toprule
AUC                  & \textit{dev}  & \textit{tst}  \\ \midrule
Supervised           & 0.72 & 0.74 \\
Supervised + ATOMIC  & 0.73 & 0.76 \\
NS - ATOMIC          & 0.62 & 0.65 \\
\hspace{0.57cm}- Annotated          & 0.67 & 0.67 \\
\hspace{0.57cm}- Annotated + ATOMIC & 0.63 & 0.65 \\
Generator (ATOMIC)   & 0.49 & 0.50 \\
\bottomrule
\end{tabular}

\caption{\label{Tab:stage3_models} AUC of models discriminating the validity of abstract triples: Supervised models trained on our annotated dataset, NS models trained by negative sampling, and one scorer based on a COMET-like generator directly trained on ATOMIC . Training data are possibly augmented by ATOMIC triples.}
\end{table}

We then train and evaluate models on the set of annotated abstract triples $S_t=\{(h_i, r_i, t_i)\}$ for inference verification, again adapted from RoBERTa, 
and report results with the measure of Area Under Curve (AUC) due to imbalanced data. While we also consider several baselines and alternative methods:

1. +ATOMIC: Since the abstract triples are often similar to ATOMIC triples, and the size of ATOMIC is much larger, we attempt to mix the cleaned ATOMIC data $S_A$ created in Section~\ref{Sec:clean} into training. We select a subset $S_A^*$ based on randomly sampled heads in $S_A$ with the same number of heads with $S_t$, 
forming a 1:1-mixed data $S_{mix}=S_t \cup S_A$. 

2. Negative Sampling (NS): 
We train the model in a similar negative sampling manner, possibly combined with +ATOMIC, performing NS upon the ATOMIC samples $S_A^*$, positive samples in the annotated data $S_t^+$, or a mix of $S_{mix}^+=S_t^+ \cup S_A^*$. We form four negative samples with randomly sampled tails in $S_A^+$, $S_t^+$, or $S_{mix}^+$ per each positive one to match the ground truth proportion. 

3. Generator (ATOMIC): we directly use a COMET-like generator for classification in the same way above, which is fine-tuned from GPT2-medium on ATOMIC and reaches a good 30.3 BLEU-2 score on $S_A$.


For models trained purely on ATOMIC data, concept indicators (brackets) are removed during evaluation. 

As in Table~\ref{Tab:stage3_models}, the supervised models significantly outperform all the models that do not (fully) rely on our annotated data, while +ATOMIC brings slight improvements. The ATOMIC generator is almost ineffective. All the results above indicate that PTLMs themselves have limited capabilities on the task and our annotated dataset is crucial, which is also consistent with observations from concurrent works evaluating conceptual knowledge of PTLMs \cite{DBLP:conf/emnlp/PengWHJ0L0022,DBLP:conf/acl/WuJJXT23}.

\subsection{Abstract ATOMIC}
\label{Sec:abs_atomic}

\begin{table}[h]
\centering
\begin{tabular}{lccc}
\toprule
         & 0.7\symtilde0.8 & 0.8\symtilde0.9 & $\geq$0.9 \\ 
        \midrule
        Event Conceptualization (by Neural Concept Generator) & 10.3K & 17.7K & 171.1K \\
        Event Conceptualization (by Heuristic Concept Linker) & 8.3K & 11.5K & 81.3K \\
        Event Conceptualization (Total) & 16.7K & \underline{26.2K} & \underline{203.0K} \\
        Different Abstract Event & 4.3K & \underline{7.0K} & \underline{63.0K} \\
        Abstract Triple & 542.2K & 937.2K & \underline{2,947.9K} \\
\bottomrule
\end{tabular}
\caption{\label{Tab:extend_size} The number of abstract events, event conceptualizations, and triples built upon ATOMIC, with different scores given by the corresponding verifier. Those included in the Abstract ATOMIC are underlined, indicating the large scale of Abstract ATOMIC. Abstract events are counted according to their corresponding event conceptualization with the highest score.}
\end{table}

\begin{table}[h]
\centering
\begin{tabular}{lccc}
\toprule
         & 0.7\symtilde0.8 & 0.8\symtilde0.9 & $\geq$0.9 \\ 
        \midrule
        Event Conceptualization (by Neural Concept Generator) & 0.64 & \underline{0.72} & \underline{0.88} \\ 
        Event Conceptualization (by Heuristic Concept Linker) & 0.67 & \underline{0.74} & \underline{0.90} \\ 
        Abstract Triple & 0.41 & 0.55 & \underline{0.71} \\ 
\bottomrule
\end{tabular}
\caption{\label{Tab:extend_eval} Manually-inspected validity for event conceptualizations and abstract triples built upon ATOMIC, in different verifier score ranges. Those included in the Abstract ATOMIC are underlined, indicating a satisfying quality of Abstract ATOMIC.}
\end{table}

With the whole pipeline built and models available, we are now able to build a conceptualized version of ATOMIC for abstract knowledge that can be derived from ATOMIC under the framework in Figure~\ref{Fig:method}. 

On all the 31.1K entities or eventualities identified in ATOMIC after filtering as mentioned in Section~\ref{Sec:clean}, we apply the heuristic concept linker and the neural concept generator (with a 10-beam search), with outputs  evaluated by the supervised discriminator. Event conceptualizations with scores no less than \ul{0.8} are considered valid. We then run the supervised inference verifier for the possible abstract triples from the derived abstract events. Triples with scores no less than \ul{0.9} are picked as valid, forming our \textbf{Abstract ATOMIC} that could be used for future conceptual induction. Similarly, Abstract ATOMIC is partitioned into \textit{trn}/\textit{dev}/\textit{tst}, used in the following experiments, according to the split of the instance ATOMIC events. 

Statistics of the Abstract ATOMIC are given in Table~\ref{Tab:extend_size}, regarding the number of events (from rules or models) and triples, whose scores fall in different ranges. The underlined numbers indicate the part that is considered valid and included into Abstract ATOMIC. As demonstrated, an extensive abstract CKG with total 70.0K abstract events, 229.2K event conceptualizations, and 2.95M abstract triples is induced. 

As for the quality of this automatically-derived abstract CKG, from each score range of event conceptualizations and abstract triples, we manually inspect 100 random samples. As shown in Table~\ref{Tab:extend_eval}, the validity is satisfying, though it indicates that the modeling of conceptualization remains rather challenging. It also explains why we have to pick a high threshold (0.8 or 0.9) for an acceptable (>0.7) quality.

By comparing the total number of event conceptualizations with the number from the two routes in Table~\ref{Tab:extend_size}, we can find that both routes have their own contribution with only 22.9\% overlap. More, we notice that the heuristic linker gives broader and more diverse candidates, as the large-scale taxonomy is explicitly leveraged. But it also brings more errors, which makes the discriminator crucial. While the LM generator produces a smaller number of conceptualizations that are more coherent to the semantic context and more likely to be valid. All these findings justify our dual-route design. See \ref{Appendix:abs_atomic} for more details.

\begin{table}[t]
\centering
\begin{tabular}{lcccc}
\toprule
&  \multicolumn{2}{c}{GPT2-base} & \multicolumn{2}{c}{GPT2-medium} \\
\midrule
BLEU-2                                & \textit{dev}   & \textit{tst}   & \textit{dev}   & \textit{tst}   \\
\midrule
Baseline        & 17.4 & 17.7 & 19.8 & 19.6 \\
+Conceptualization (human) & 19.1 & 19.2 & 21.0 & 22.4 \\
\hspace{0.5cm}+Fine-tune      & 20.7 & 20.6 & 23.0 & 23.5 \\
+Conceptualization (auto)         & 19.3 & 19.1 & 19.7 & 20.0 \\
\hspace{0.5cm}+Fine-tune      & 19.5 & 19.3 & 20.9 & 21.0 \\
+Conceptualization (both)           & 18.7 & 18.5 & 21.0 & 21.6 \\
\hspace{0.5cm}+Fine-tune      & 19.0 & 19.0 & 22.4 & 22.9 \\
\bottomrule
\end{tabular}

\caption{\label{Tab:comet_aid} Performance (BLEU-2 score, \%) for COMET-like modeling on an ATOMIC subset, possibly with the human-annotated or automatically-derived abstract triples injected, and possibly then fine-tuned on the target ATOMIC subset.}
\end{table}

\section{Extrinsic evaluation}
\subsection{Concept-aided ATOMIC modeling}
\label{sec: COMET}

We now intend to explore the possible use of abstract knowledge for commonsense modeling. It is intuitive that since abstract knowledge provides a more general view, it may help inference on commonsense knowledge, particularly when there is limited knowledge known. 

Therefore, we attempt to inject abstract knowledge as a kind of data augmentation to COMET-like training. 
Given 65,900 human-annotated (\textit{human}) abstract triples linked to 688 ATOMIC instance events, we pick a subset of ATOMIC and the automatically-derived (\textit{auto}) Abstract ATOMIC corresponding to the same set of events. Then we perform COMET-like causal LM training on the ATOMIC subset, possibly mixed with the \textit{trn} split of the \textit{human} and/or \textit{auto} abstract samples. 

Since the abstract data contain more triples, the ATOMIC data is upsampled with a 1:2 ratio. 
In addition, the model can be fine-tuned on the ATOMIC subset we use after the mixed training. Experiments are carried out on both GPT2-base and GPT2-medium. We then evaluate on the whole ATOMIC dataset.

As reported in Table~\ref{Tab:comet_aid}, all models injected with abstract knowledge can significantly outperform the baseline trained merely on the original non-abstract ATOMIC, while further fine-tuning will lead to varied improvements. 
Therefore, abstract knowledge, either annotated or automatically derived, may implicitly help the modeling of situational commonsense, even without any explicit instantiation.

\begin{table*}[t]
\small
\centering
\resizebox{\linewidth}{!}{
\begin{tabular}{@{}l|c|ccccc|l@{}}
\toprule
Model & KG & a-NLI \cite{DBLP:conf/iclr/BhagavatulaBMSH20} & CSQA \cite{DBLP:conf/naacl/TalmorHLB19} & PIQA~\cite{DBLP:conf/aaai/BiskZLGC20} & SIQA~\cite{DBLP:conf/emnlp/SapRCBC19} & WG~\cite{DBLP:conf/aaai/SakaguchiBBC20} & Avg. \\ 
\midrule
Majority label  & - & 50.8 & 20.9 & 50.5 & 33.6 & 50.4 & 41.2 \\
GPT2-L~\cite{radford2019gpt2} & - & 56.5 & 41.4 & 68.9 & 44.6 & 53.2 & 52.9 \\
RoBERTa-L~\cite{DBLP:journals/corr/abs-1907-11692} & - & 65.5 & 45.0 & 67.6 & 47.3 & 57.5 & 56.6 \\
DeBERTa-v3-L~\cite{he2023debertav} & - & 59.9 & 25.4 & 44.8 & 47.8 & 50.3 & 45.6 \\
Self-talk~\cite{DBLP:conf/emnlp/ShwartzWBBC20} & - & - & 32.4 & 70.2 & 46.2 & 54.7 & - \\
COMET-DynGen~\cite{DBLP:conf/aaai/BosselutBC21} & ATOMIC & - & - & - & 50.1 & - & - \\
SMLM~\cite{DBLP:conf/emnlp/BanerjeeB20} & * & 65.3 & 38.8 & - & 48.5 & - & - \\
MICO~\cite{DBLP:conf/emnlp/SuWFZSZ22} & ATOMIC & - & 44.2 & - & 56.0 & - & - \\
STL-Adapter~\cite{DBLP:conf/naacl/KimKKAHY22} & ATOMIC & 71.3 & 66.5 & 71.1 & 64.4 & 60.3 & 66.7 \\
BUCA~\cite{DBLP:conf/acl/HeUGP23} & ATOMIC & - & 60.3 & - & 61.4 & - & - \\
\midrule
\multirow{3}{*}{RoBERTa-L + MR \scriptsize{\textit{340M}}} & ATM$_{10X}$ & 70.8 & 59.4 & 72.1 & 58.5 & 58.3 & 63.8 \\
 & ATOMIC & 70.8 & 64.2 & 72.1 & 63.1 & 59.2 & 65.9 \\
 & \textbf{Abs.ATM} & 71.4$_{\small\uparrow0.6}$ & 65.8$_{\small\uparrow1.6}$ & 72.9$_{\small\uparrow0.8}$ & 65.5$_{\small\uparrow2.4}$ & 59.7$_{\small\uparrow0.5}$ & 67.1$_{\small\uparrow1.2}$ \\
\midrule
\multirow{3}{*}{DeBERTa-v3-L + MR \scriptsize{\textit{435M}}} & ATM$_{10X}$ & 75.1 & 71.6 & \textbf{79.0} & 59.7 & 71.7 & 71.4 \\
 & ATOMIC & 76.0 & 67.0 & 78.0 & 62.1 & 76.0 & 71.8 \\
& \textbf{Abs.ATM} & \textbf{78.2}$_{\small\uparrow2.2}$ & 68.1$_{\small\uparrow1.1}$ & 78.2$_{\small\uparrow0.2}$ & 63.5$_{\small\uparrow1.4}$ & \textbf{78.3}$_{\small\uparrow2.3}$ & \textbf{73.2}$_{\small\uparrow1.4}$ \\
\midrule
\multicolumn{8}{@{}l}{\textbf{Large Language Models}} \\
GPT-3.5 (\texttt{text-davinci-003}) & - & 61.8 & 68.9 & 67.8 & 68.0 & 60.7 & 65.4 \\
ChatGPT (\texttt{gpt-3.5-turbo}) & - & 69.3 & \textbf{74.5} & 75.1 & \textbf{69.5} & 62.8 & 70.2 \\
\midrule
\multicolumn{8}{@{}l}{\textbf{Supervised Learning \& Human Performance}} \\
RoBERTa-L (Supervised) & - & 85.6 & 78.5 & 79.2 & 76.6 & 79.3 & 79.8 \\
Human Performance & - & 91.4 & 88.9 & 94.9 & 86.9 & 94.1 & 91.2 \\
 \bottomrule
\end{tabular}
}
\caption{Zero-shot evaluation results (Accuracy \%) on five commonsense QA benchmarks. -L stands for the \textit{-large} model, 
ATM$_{10X}$ for ATOMIC$_{10X}$ \cite{DBLP:conf/naacl/WestBHHJBLWC22}, and Abs.ATM for Abstract ATOMIC (i.e. our approach). 
$\uparrow$ stands for the accuracy gains by comparing models trained on Abstract ATOMIC against ATOMIC only.
The best performances are bold-faced.
}
\label{tab:zeroshot_exp_results}
\end{table*}

\subsection{Zero-shot commonsense QA}
\label{sec:zero-shot_csqa}


In addition to modeling the base CKG, we also try to explore the impact of conceptualization on the \textit{zero-shot commonsense QA} task~\cite{DBLP:conf/emnlp/ShwartzWBBC20,DBLP:conf/acl/LiWDWX20} to show the improvements abstract knowledge can bring to commonsense reasoning. 
This task assesses the model's ability to reason on unseen situations in a generalizable manner without any training on the target data~\cite{DBLP:conf/aaai/DouP22}.
For this goal, current state-of-the-art methods \cite{DBLP:conf/naacl/KimKKAHY22,CAR,CANDLE,DBLP:journals/corr/abs-2403-07398} leverage synthetic QA pairs transformed from CKG triples, where the head and relation are turned into a question using templates, and the tail serves as the answer. 
While distractors can be generated by sampling unrelated triples with the same relation, upon which PTLMs are fine-tuned using marginal ranking (MR) loss to create a general commonsense QA model, to be evaluated on various target benchmarks \cite{DBLP:conf/aaai/MaIFBNO21}.

To investigate the effects of introducing abstract commonsense knowledge, we keep such QA synthesis protocol and model training process unchanged, and ablatively study the role of abstract knowledge. 
Specifically, we use ATOMIC as the base CKG, and Abstract ATOMIC as its top-on expansion for abstract knowledge. 
As for Abstract ATOMIC, we utilize both the human-annotated and the automatically generated data. Those abstract triples are synthesized into QA pairs in combination with the original situational ATOMIC triples to train the QA model, marked by \textbf{Abs.ATM}.

We report results of various existing baselines, including vanilla PTLMs such as RoBERTa and DeBERTa~\cite{he2023debertav} as well as previous works on the task including Self-talk~\cite{DBLP:conf/emnlp/ShwartzWBBC20}, STL-Adapter~\cite{DBLP:conf/naacl/KimKKAHY22}, etc., evaluated on multiple benchmarks including a-NLI \cite{DBLP:conf/iclr/BhagavatulaBMSH20}, CSQA \cite{DBLP:conf/naacl/TalmorHLB19}, etc., as listed in Table~\ref{tab:zeroshot_exp_results}.
We then compare the results obtained by MR~\cite{DBLP:conf/aaai/MaIFBNO21} on ATOMIC and the much larger ATOMIC$_{10X}$~\cite{DBLP:conf/naacl/WestBHHJBLWC22} against those obtained on Abstract ATOMIC to study the role of introducing abstract knowledge.
As for methods that consolidate multiple CKGs, only ATOMIC is used as the knowledge source for fair comparison. 
We also report the zero-shot performance of GPT3.5~\cite{DBLP:conf/nips/Ouyang0JAWMZASR22} and ChatGPT as two more competitive baselines~\cite{CAR}.

As reported in Table~\ref{tab:zeroshot_exp_results},
among the baselines the best performance is achieved by DeBERTa-v3-Large trained with the MR approach~\cite{DBLP:conf/aaai/MaIFBNO21} on ATOMIC and ATOMIC$_{10X}$, followed by ChatGPT.
The ATOMIC$_{10X}$-based models fail to surpass the ATOMIC-based ones, suggesting the limitations of distilled commonsense knowledge from PTLMs in ATOMIC$_{10X}$.
While our top-performing system, which is based on DeBERTa-v3-large with Abstract ATOMIC, outperforms all the baselines significantly, improving performance by an average of 1.4\%, compared to the best baseline (the same model with ATOMIC).
It also champions two challenging benchmarks and outperforms ChatGPT by an average of 3\%.
These results underscore the effectiveness of using abstract commonsense knowledge to benefit downstream QA tasks.

\subsection{Transfer to ConceptNet}

With a variety of formats for knowledge representation in different CKGs, it is valuable to verify if our constructed pipeline can be applied to other CKGs beyond ATOMIC, such as ConceptNet~\cite{DBLP:conf/aaai/SpeerCH17}. 
Therefore, we examine our GPT2-based concept generator obtained in Section~\ref{Sec:exp_abs}. 
We randomly sample 100 event-like nodes from ConceptNet and prompt our model to generate plausible conceptualizations for each of them. 
To ensure consistency with the textual styles used in ATOMIC, we pre-processed the events using natural language templates as described by~\cite{DBLP:conf/emnlp/FangWCHZSH21}. 
Then, we employ the skills of three postgraduate students well-versed in commonsense reasoning to carry out expert assessments on the generated conceptualizations and determine the proportion of plausible ones. 
The findings indicate that 76\% of them are accurate, highlighting the value of generative models trained on Abstract ATOMIC annotations in comprehending and conceptualizing alternative form of knowledge in other CKGs like ConceptNet.

\section{Example studies}

\begin{table}[!t]
\centering
{\def\arraystretch{1.6}
\begin{tabularx}{\textwidth}{>{\raggedright}p{4cm}p{5.5cm}p{5.5cm}} \toprule

Event & Positive Annotation & Negative Annotation \\ \midrule
\underline{PersonX gives} \underline{birth to children} & life event, giving birth, happy event & life skill, common procedure,   routine medical procedure \\
\underline{PersonX wraps} \underline{PersonY's arms} & cuddling, hold, type of intimacy, embrace,   holding, bodily contact & bind, physical barrier,   protection, technique, cover \\
PersonX buys \underline{a new iphone} & new iphone, product, iphone, smart phone, mobile technology, device & essential technology, item \\
PersonX takes PersonY to \underline{the pound} & pound, animal housing facility,   multi animal facility, animal shelter & non profit, currency,   organization, unit, imperial unit \\
PersonX has trouble \underline{paying PersonX's bills} & pay, payment, paying bill, expense & transaction, task, simple task,   handle, routine task, service 
\\

\bottomrule

\end{tabularx}
}
\caption{\label{Tab:case_abs_anno} Examples of positive and negative annotations on event conceptualization, target entities/eventualities underlined.}
\end{table}

\begin{figure}[!t]
\centering
\includegraphics[width=0.9\textwidth]{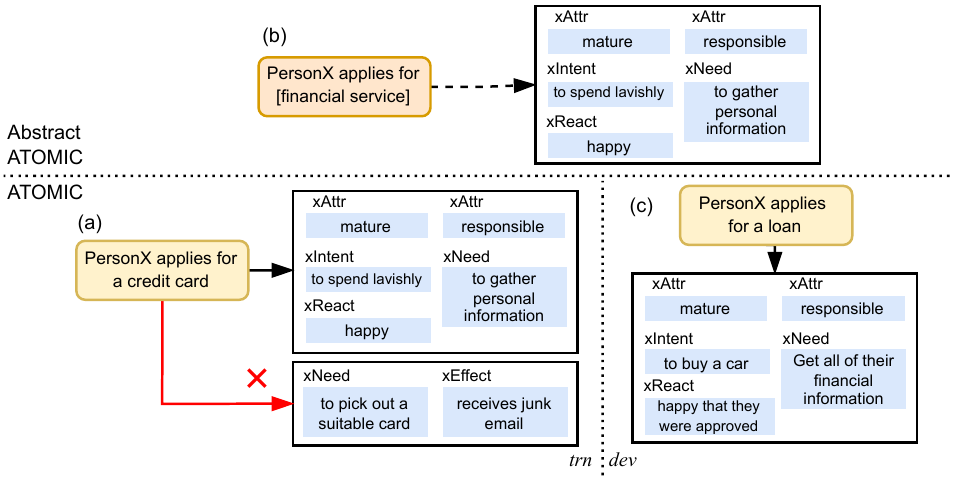}
\caption{An example of conceptual induction on the original ATOMIC based on Abstract ATOMIC. 
From an event and its triples from the ATOMIC \textit{trn} split, the generalized abstract event and its inferences are built using our pipeline. Inferences deemed not typically held for the abstract event according to our inference verifier are excluded, marked in the red edge with a cross. Then the abstract triples could be instantiated to reason about an unseen event in the \textit{dev} split.
}
\label{Fig:case_induce}
\end{figure}

In this section, we investigate some examples of the annotated and automatically derived data for event conceptualizations and abstract triples.

\paragraph{Event conceptualization}
Examples of annotated event conceptualization are given in Table~\ref{Tab:case_abs_anno}. We can find that the results are acceptable on various types of entities and eventualities, although not perfect (consider ``mobile technology'' and ``item'' for the ``a new iphone'' example). 

Examples of event conceptualizations in Abstract ATOMIC are given in Table~\ref{Tab:case_abs}, some covered in the annotated data. We can find that the methods generalize to unseen entities and eventualities of various types, and the rule-based and neural routes complement each other, often both giving adequate results, 
Sometimes the rule-based ones can be more diverse and fine-grained, as in the ``water bill'' and ``get a cold'' case. By leveraging WordNet for word sense disambiguation and nominalization, rule-based results are improved, like the ``bad day at work'' case where the verb ``have'' is linked to the sense of experience. However, in some cases, heuristics give results of lower quality that are inadequate or incomplete, particularly in non-object cases (like in the ``wrap'' case) and those requiring contexts (like in the ``bad day at work'' case). Fortunately, those incorrect results are mostly ruled out by the verifier, which features its necessity as a final gatekeeper. The generator produces fair examples in most cases, although there are still generations that seemingly simply come from textual relevance, as in the results like ``domestic pet'' in the ``pound'' case.

\paragraph{Abstract triples}
Examples from both the annotated abstract triples and Abstract ATOMIC are given in Table~\ref{Tab:case_triple}. As a more challenging task, more errors are found.
Nevertheless, many tails that hold only in specific instances are correctly excluded. For example, as for ``PersonX calls [health professional],'' tails regarding specific disciplines of health care (like those on ``vet'' and ``dentist,'' e.g. ``xIntent: to clean their teeth'') should not be included. According to our manual inspection, among 31 tails collected from the instances and deemed negative, 22.6\% fall in this category of over-specific ones. While among 115 tails classified as positive, there is a lower rate of 14.8\%. Similarly, for the more general ``PersonX calls [professional]'' case, many of the health-related tails are identified and excluded.

\begin{table}[p]
\centering
{\def\arraystretch{1.6}
\begin{tabularx}{\textwidth}{>{\raggedright}p{3cm}>{\RaggedRight\arraybackslash}p{1.6cm}p{5.1cm}p{5cm}} \toprule
Event & Linked Concepts & Heuristic-based {Conceptualizations} & Generated Conceptualizations \\ \midrule

\underline{PersonX gives birth} \underline{to children} & giving birth, birth, gift & \textbf{occasion}, \textbf{birth}, \textbf{life} \textbf{change}, \textbf{life} \textbf{event}, \textbf{happy event},  gift, business courtesy, routine medical procedure, biological process & \textbf{creation}, \textbf{birth}, \textbf{event}, \textbf{life event},  \textbf{life cycle event}, \textbf{life changing event} \\

\underline{PersonX wraps} \underline{PersonY's arms} & wrapping & \textbf{wrapping}, surgical and endovascular procedure, technique  & \textbf{close} \textbf{contact}, \textbf{physical} \textbf{touch}, \textbf{physical contact}, \textbf{action}, close\\

PersonX buys\linebreak\underline{a new iphone} & iphone, new iphone & \textbf{ios device}, \textbf{smart phone}, \textbf{mobile device}, \textbf{device},   \textbf{iphone}, \textbf{apple device}, \textbf{product}, product release, wishlist item, revolutionary product & \textbf{smart phone},   \textbf{phone}, \textbf{product}, \textbf{iphone}, \textbf{electronic   device}, \textbf{device}, \textbf{mobile device} \\

PersonX takes PersonY to \underline{the pound} & pound & \textbf{pound}, \textbf{animal management facility}, unit, measurement, poet, currency & \textbf{animal shelter}, \textbf{animal facility}, \textbf{business}, \textbf{public facility}, domestic pet, domestic animal, farm facility, place \\

PersonX has trouble \underline{paying} \underline{PersonX's bills} & pay, payment, payroll & \textbf{expense related aspect}, \textbf{financial activity}, \textbf{payment}, \textbf{expense}, banking service, financial service,   longstanding issue, cost & \textbf{paying bill},   \textbf{payment}, transaction, pay, paying bills \\ \midrule

PersonX pays \underline{PersonX's water bill} & bill, water bill & \textbf{bill}, \textbf{expense}, \textbf{basic household} \textbf{expense}, \textbf{user charge}, \textbf{utility bill}, \textbf{water bill}, document, documentary evidence & \textbf{bill}, \textbf{expense},   \textbf{water bill}, calculation, transaction \\

\underline{PersonX gets} \underline {a cold} & contracting, cold & \textbf{cold}, \textbf{common illness},   \textbf{illness}, \textbf{infection}, \textbf{minor ailment},   \textbf{respiratory infection}, \textbf{upper respiratory infection},   \textbf{viral infection}, abiotic stress, contracting, work arrangement & \textbf{cold}, \textbf{condition},   \textbf{sickness}, injury, unexpected event \\

\underline{PersonX has a} \underline{bad day at work} & experience & \textbf{experience}, criterion, meeting specific requirement, personal characteristic, personal factor,   qualification, variable & \textbf{difficulty}, \textbf{negative event}, \textbf{problem}, \textbf{unpleasant experience}, stress, unpleasant day \\
\bottomrule

\end{tabularx}
}
\caption{\label{Tab:case_abs} Examples of event conceptualizations in Abstract ATOMIC. Positive event conceptualizations according to the discriminator are marked in bold, and target conceptualization candidates are underlined.}
\end{table}

\begin{table}[p]
\centering
{\def\arraystretch{1.6}
\begin{tabularx}{\textwidth}{p{1.9cm}>{\RaggedRight\arraybackslash}p{1.9cm}p{1.3cm}p{4.7cm}p{4.6cm}} \toprule
Abstract Event & Instances & Relation & Positive Tails & Negative Tails\\ \midrule

PersonX buys [item] & \multirow{3}{1.9cm}{a new chair, a new shirt, skateboard, etc.} & xNeed & to go to the store, to open his wallet, to   go to the clothing store & to pick out a new shirt \\
 &  & xAttr & decisive, stylish, excited, rich & athletic, utilitarian, carefree \\
 &  & xEffect & gets receipt, get a coupon, style changes & person x takes skate board home \\
 \midrule

PersonX calls [health professional] & \multirow{3}{1.9cm}{the doctor, the dentist, the vet, PersonX's doctor, PersonX's dentist} & xWant & set an appointment, to ask what they should do, to ask for device, to get   a filling, to ask the doctor a question, to tell the doctor their problems & to take their pet there, to ask a question \\

 &  & xIntent & to schedule an appointment, to help pet, to be healthy, to feel better & to know about their pet, to clean their teeth, to be informed \\
 
 &  & xNeed & dial the number, find the number, look up things online, to pick up the   phone, to want them to get better, to research medical care in the area & to have a sick animal, to get PersonX's doctor's phone number, to grab a   phone \\
 
PersonX calls [professional] & \multirow{3}{1.9cm}{the doctor, plumber, exterminator, cable guy, locksmith, etc.} & oWant & set an appointment, to unclog the pipe, give an estimate & catches rodents, to get paid, to make a new key \\

 &  & xIntent & a service, to make an appointment, be healthy, to watch tv & to change a lock, to clean their teeth, to be informed, to fix some   plumbing \\
 
 &  & xEffect & a voice mail plays, admits in hospital, gets advice, waits for their   appointment & PersonX has to pay a bill for the repairman, is checked by the doctor,   ask a question \\
 
[homecoming] & \multirow{3}{1.9cm}{PersonX makes it home safely, PersonX comes back, PersonX comes to PersonY's house, etc.} & oWant & to greet PersonX, to hug him, to help him   relax, to eat out, to invite PersonX inside & To greet X, to go eat, to have dinner, to   talk to PersonX \\

 &  & xIntent & see their family, to get home, to sleep, see their family, to attend some   competition & have a break from learning, to attend the wedding, to be with someone \\
 
 &  & xReact & cozy, happy, nostalgic, relaxed & drink, ready to eat, sleepy \\

\bottomrule

\end{tabularx}
}
\caption{\label{Tab:case_triple} Examples of positive and negative abstract triples in the annotated data (the first row) and the automatically derived Abstract ATOMIC (following rows). The \textit{Instances} column shows examples of the original ATOMIC events that are the instances of the abstract event, indicated by the corresponding instances for the bracket-enclosed concept.}
\end{table}

\paragraph{Conceptual induction}
An example of conceptual induction is illustrated in Figure~\ref{Fig:case_induce}. The event ``PersonX applies for a credit card'' can be found in the original ATOMIC as in (a). Using our pipeline, an abstract event ``PersonX applies for [financial service]'' is built in Abstract ATOMIC, along with its inferences in (b). Some of the credit-card-specific inferences marked by red edges in (a) are excluded by the verifier, so that abstract triples in (b) are mostly valid in the general sense. 
There is also an event ``PersonX applies for a loan'' in the \textit{dev} set as in (c), which is a new and unseen one under the ATOMIC split. It could be conceptualized to the event as well through our pipeline. Therefore the abstract triples can be instantiated to the (c) event to create inferences like ``xAttr: responsible,'' ``xNeed: to gather personal information,'' etc. Its ground truth ATOMIC inferences are given in (c), and we can find that our inferences through the conceptual induction pipeline are identical or relevant to them.

\section{Related work}


\subsection{Commonsense knowledge acquisition}
As an indispensable component of AI, there has been plenty of work studying commonsense knowledge, for which
the primary step is to acquire the knowledge.
Below introduce the approaches for commonsense acquisition, which can be broadly classified into four types. We then discuss our choice to base our research upon the event-centric textual CKG, ATOMIC, by comparing it with other types of knowledge. 

\paragraph{Human annotation}
Human annotation has been a critical part of collecting commonsense starting from early pioneer works, possibly as logical formulae in Cyc~\cite{DBLP:journals/cacm/Lenat95CYC,Lenat1998} or textual assertions in OMCS~\cite{DBLP:conf/coopis/SinghLMLPZ02OMCS}, which is later utilized as a core part of the renowned ConceptNet~\cite{liu2004conceptnet, DBLP:conf/aaai/SpeerCH17} focused on entity-centric knowledge.
There are also attempts to annotate event-centric/situational knowledge, with focuses on social commonsense \cite{DBLP:conf/acl/SmithCSRA18,DBLP:conf/emnlp/ForbesHSSC20socialchemistry}), dialogue \cite{DBLP:conf/acl/GhosalSMMP22,DBLP:journals/corr/abs-2305-02364}, and narratives \cite{DBLP:conf/emnlp/MostafazadehKMB20}, as well as ATOMIC \cite{DBLP:conf/aaai/SapBABLRRSC19} and ATOMIC-2020 \cite{DBLP:conf/aaai/HwangBBDSBC21} with a range of dimensions of if-then reasoning on events. 
Despite their high quality, such CKGs suffer from limited scale and coverage over various entities and eventualities, as well as potential human bias~\cite{DBLP:conf/naacl/SapSVZCS22} possibly including concept bias. These drawbacks stand among our major motivations to introduce conceptualization.

\paragraph{Information extraction}
To scale up the size of CKGs, automatic extraction from large corpora based on dedicated schema and templates comes out as an alternative solution \cite{DBLP:conf/aaai/GordonDS10,DBLP:conf/wsdm/TandonMSW14,DBLP:conf/cikm/TandonMDW15}.
On top of them, ASER~\cite{zhang2022aser,zhang2019aser} proposes a unique paradigm to 
retrieve a large-scale CKG for eventualities as linguistic graphs of predicates and their arguments linked by various discourse relations. 
However, without human supervision, such methods suffer from much lower quality and larger noise despite much their larger scale. Morepver, they are particularly exposed to reporting bias, including concept bias. Hence such CKGs alone may not be ideal for our goal to acquire high-quality knowledge covering various entities and eventualities.

\paragraph{Generating from language models}
With knowledge from enormous pretraining data internalized into PTLMs, a pile of works is conducted to mine knowledge from them \cite{ davison2019commonsense,shi2021transformers,DBLP:conf/cogsci/WeirPD20,petroni2019lama},
which engenders a line of research to directly use PTLMs as commonsense KB, even as a source of knowledge to train COMET \cite{DBLP:conf/naacl/WestBHHJBLWC22}, possibly through various prompt discovery methods \cite{DBLP:journals/tacl/JiangXAN20,DBLP:conf/emnlp/ShinRLWS20,DBLP:conf/naacl/ZhongFC21}.
Furthermore, they are directly applicable to commonsense reasoning tasks \cite{DBLP:conf/acl/TamborrinoPPVN20}. 
However, as mentioned in Section~\ref{Sec:commonsense_modeling}, LMs merely based on induction from word co-occurrence may not generalise well to diverse entities and eventualities, while we explicitly incorporate taxonomy and annotated conceptual knowledge to tackle the problem.

\paragraph{Extending from existing knowledge graphs} 
A number of efforts have been made to complete or extend existing CKGs by commonsense reasoning and modeling, and our conceptual induction approach can be also viewed as one of them. Early works are mostly focused on logical reasoning \cite{DBLP:journals/jair/Davis17}, while link prediction or densification of factual KGs based on structural information using negative sampling is a well-studied task \cite{DBLP:journals/tnn/JiPCMY22}. 
As for CKGs, with the importance of semantic information in text-form nodes, NLP methods become essential \cite{DBLP:conf/acl/LiTTG16,yao2019kgbert}. More recently, there are many attempts to combine semantic and structural information \cite{DBLP:conf/acl/MoghimifarQZHB20,DBLP:conf/www/WangSLZW021,huang2021structured,DBLP:conf/acl/LovelaceNVLR20,malaviya2019ckgcompletion,DBLP:conf/ijcnn/WangWHYLK21}, 
which might be valuable for conceptualization as well. Nevertheless, we focus on text-based approaches and left it for future work. 

In addition to the densification approach that only brings new edges, there are also generative models to predict new tail nodes given text-form events \cite{DBLP:conf/conll/SaitoNAT18}, which leads to COMET \cite{DBLP:conf/acl/BosselutRSMCC19}. 
The successive work is further focused on directly using the model COMET-ATOMIC-2020 as a neural KB for commonsense inference from an event \cite{DBLP:conf/aaai/HwangBBDSBC21}. We follow a similar approach to view the extension of existing CKGs as an NLP task, with PTLMs as a key component. However, we leverage taxonomies instead of mere PTLMs to directly create more abstract events and make higher-level inferences upon them, and such abstract knowledge is essentially distinct from the original lower-level situational knowledge.

Another approach is CKG population, like TransOMCS \cite{DBLP:conf/ijcai/ZhangKSR20} and DISCOS \cite{DBLP:conf/www/FangZWSH21,DBLP:conf/emnlp/FangWCHZSH21}, which extends CKGs by transferring from external knowledge \cite{DBLP:conf/acl/JiG11}, e.g. from the larger and noisier extracted CKGs to smaller human-annotated ones. 
Attempts like CSKG further try to combine existing CKGs altogether \cite{DBLP:journals/kbs/IlievskiOMZMS21,DBLP:conf/esws/IlievskiSZ21}, which have been shown promising for downstream tasks \cite{DBLP:conf/aaai/MaIFBNO21}. They are also close to our approach to extend existing CKGs with other KGs, but we are specifically motivated by human conceptualization and thus focused on utilizing external taxonomic knowledge. 

\paragraph{Comparison with other types of knowledge}
\label{Sec:compare_knowledge}
We follow ATOMIC to study the \textbf{event-centric commonsense} represented by rather free-form texts in an intuition-based approach. As mentioned in Section~\ref{Sec:ATOMIC_approach}, this is distinct from Cyc-like \textbf{symbolic reasoning} \cite{DBLP:journals/cacm/Lenat95CYC,gordon2017formal},
\textbf{factual knowledge} defined in traditional KBs like DBPedia \cite{DBLP:conf/semweb/AuerBKLCI07dbpedia}, Freebase \cite{DBLP:conf/sigmod/BollackerEPST08}, and YAGO \cite{DBLP:conf/www/SuchanekKW07}, as well as \textbf{entity-centric commonsense} like ConceptNet \cite{liu2004conceptnet, DBLP:conf/coopis/SinghLMLPZ02OMCS, DBLP:conf/wsdm/TandonMSW14,DBLP:conf/semweb/IlievskiSS20}. 
It is also different from \textbf{script knowledge} \cite{schank1975scripts, DBLP:conf/acl/RegneriKP10}, which represents ``a standardized sequence of events that
describes some stereotypical human activity'' \cite{barr1981handbook}. 
Although both centered on events, script knowledge focuses on the \textit{sequence} of events that leads to a potential script, while ATOMIC-like CKGs are \textit{graphs} defined by pairwise relations, which highlight diverse possibilities of outbound edges for an event.
Also, script knowledge is focused on temporal relations between events, whereas ATOMIC covers a broad range of dimensions including intentions, prerequisites, attributes of the agent, etc. 
Although there are efforts to distill pairwise relations among events from script chains, such as temporal event relation extraction \cite{DBLP:conf/acl/RothWN18} and script knowledge graph construction \cite{DBLP:conf/ijcai/LiDL18}, the focus is restricted to the temporal ordering of events without considering broader and more fine-grained commonsense inferences.

As discussed in Section~\ref{Sec:ATOMIC_approach}, the choice of the ATOMIC approach is backed by multiple theoretical advantages. While from practical aspects, ATOMIC enjoys relatively high quality compared to many other automatically extracted event-centric CKGs, hence desirable as the foundation for dataset annotation and CKG extension with optimal quality. Furthermore, being event-centric and represented in texts, ATOMIC knowledge can be flexibly applied or transformed to suit various commonsense-related tasks over natural texts. 
For example, fine-tuning a QA model on ATOMIC leads to more robust performance on multiple commonsense QA benchmarks~\cite{DBLP:conf/aaai/MaIFBNO21,DBLP:conf/naacl/KimKKAHY22}, and training COMET on ATOMIC results in more resilient and diverse commonsense inference generation compared to ConceptNet \cite{DBLP:conf/aaai/HwangBBDSBC21}.
ATOMIC has also demonstrated its usefulness in other commonsense-related tasks such as empathy dialogue generation~\cite{DBLP:conf/aaai/SabourZH22} and story generation~\cite{DBLP:journals/tacl/GuanHHZZ20}. Such superior downstream performance further justifies our choice to use ATOMIC.


\subsection{Conceptualization}

Conceptualization, as a significant component of human intelligence~\cite{murphy2004concept1}, has been investigated before under various contexts, including NLP, commonsense knowledge, as well as symbolic reasoning, Also, concept linking is necessary for the conceptualization from text. Below introduce these previous works.

\paragraph{Conceptualization on natural language} 
Our work is based on conceptualization upon entities and eventualities represented by natural texts, which has been investigated in multiple previous works. 
Taxonomies like WordNet~\cite{miller1998wordnet} and Probase~\cite{wu2012probase} collect \textit{is-a}  relations between words, and hence provide context-free conceptualization on entity-level semantics, which is essential for more complicated scenarios like our case.  
Based on that, an early line of work is particularly interested in conceptualizing verb arguments into WordNet nodes of appropriate abstractness (namely BLCs) to model selectional preference by co-occurrence statistics \cite{DBLP:conf/acl/Resnik92,DBLP:conf/naacl/Resnik93,resnik1997selectional}, using various heuristics or statistic models \cite{DBLP:journals/coling/LiA98,DBLP:journals/coling/ClarkW02,DBLP:conf/starsem/SeaghdhaK12,DBLP:conf/mtsr/HollinkBO20}. While they are all based on the strict WordNet hierarchy, and do not match our goal to cover a broader set of non-BLC concepts.

Similar to our choice as discussed in Section~\ref{Sec:ckg}, the more flexible non-hierarchical Probase with more fine-grained concepts and informative \textit{is-a} edges has been leveraged to conceptualize verb arguments \cite{DBLP:conf/aaai/ParkHW16,DBLP:conf/aaai/GongZZ16}, conceptualize entities in short texts like search queries, or further represent the meaning of the query by concepts \cite{song2011concept,DBLP:conf/cikm/SongWCW14,DBLP:conf/ijcai/SongWW15}.
In addition to all these context-free cases, 
contextualized conceptualization, as emphasized in our framework, is also carried out by levering the topic \cite{DBLP:conf/ijcai/KimWO13} and relations between entities and verbs/attributes \cite{DBLP:conf/ijcai/WangZWMW15} as well as other entities \cite{DBLP:conf/icde/HuaWWZZ15} within the text. 

All methods above are based on heuristics, statistics, or at most traditional machine learning techniques.
While recently, 
\cite{DBLP:conf/aaai/ChenHLXJ19} leverages neural models to classify entities in texts towards Probase concepts.
Nevertheless, all those works are substantially different from ours as they are 1) not aimed at capturing commonsense knowledge; 2) not covering the general sense of conceptualization upon all eventualities but restricted to nominal entities; and 3) not based on modern PTLMs with sophisticated semantic understanding for contextualized conceptualization. While very recently, there are attempts to examine PTLMs' conceptual abilities \cite{DBLP:conf/emnlp/PengWHJ0L0022,DBLP:conf/acl/WuJJXT23}, which discover that current PTLMs have limited capabilities in memorizing \textit{is-a} relations, understanding properties of concepts, and particularly, performing conceptualization within the context, even after fine-tuning on labeled data. This supports our choice to combine PTLMs, taxonomy, and annotated data in the whole pipeline so that we could reach satisfying results.

\paragraph{Conceptualization for commonsense knowledge} Unsurprisingly, conceptualization has been applied to commonsense knowledge as well, such as works to summarize abstract knowledge from KNEXT using WordNet \cite{DBLP:conf/eacl/DurmeMS09}, and to instantiate arguments to induce more specific triples \cite{DBLP:conf/flairs/BlancoCM11}, which are most similar to our work, though both before the emergence of current neural NLP techniques and flexible event-centric CKGs. Other explorations include matching concepts in software specifications to identify ambiguities \cite{DBLP:journals/corr/abs-2103-11302}, and to harvest relations between nominal BLCs based on patterns upon texts \cite{DBLP:conf/ranlp/Barbu09}.

More recently, VoCSK induces plausible eventualities by summarizing arguments of verb phrases to BLCs by the minimum description length using Probase \cite{DBLP:conf/icde/LiuZWWJZXX20,liu2022vocsk}. Similarly, \cite{DBLP:conf/naacl/PoradaSTC21} measures the consistency of neural models given texts with arguments in different abstractness and attempts to improve consistency and to leverage BLCs for scoring eventualities. Hence they are different from our work, which considers all types of entities and eventualities towards concepts of varied abstractness, including non-BLCs. 
While GenericsKB directly extracts and annotates statements or generics that apply to all instances of certain central concepts, similar to abstract events in our framework, but more focused on common factual knowledge instead of commonsense \cite{DBLP:journals/corr/abs-2005-00660}. \cite{DBLP:journals/corr/abs-2212-09246} instead directly prompts and controls PTLMs to generate generic statements about a set of concepts. Both of them are limited to generics that may naturally appear in language and thus suffer from reporting bias, while the abstract events that are more general but less linguistically natural are covered in our approach.
\cite{DBLP:journals/corr/abs-2205-11658} defines instantiations and exceptions (part of them being invalid instances in our words) of generics in logical forms. Their approach is similar to our framework but in a different direction that identifies concepts from 653 known textual generics by parsing, replacing the concepts with instantiations (instead of conceptualizations in our framework), and then scoring the candidate instantiations and exceptions based on PTLMs. Focused on the instantiations and exceptions of known generics, the method is unlikely to cover the broad range of situations in the world.

More importantly, the works above are all limited to conceptualization within an event, without touching high-level conceptualization upon non-nominal concepts, such as clauses and the whole event, as well as the even higher-level relations between those abstract events (i.e., abstract triples). 
ASER 2.0 makes a significant step forward to include abstract triples based on Probase statistics, but is still limited to the conceptualization of arguments within eventualities \cite{zhang2022aser}. Several recent works cover conceptual relations between eventualities: ELG harvests \textit{is-a} relations between whole non-abstract eventualities \cite{DBLP:journals/corr/abs-1907-08015}, \cite{DBLP:conf/akbc/YuZSNS20} attempts to identify entailment relations between ASER eventualities, and
FolkScope utilizes conceptualization to extract the abstract intention behind various event types common in online shopping experiences \cite{DBLP:conf/acl/YuWLBSLG0Y23}. 
All these methods above reflect some aspects of the conceptualization approach presented in this paper, but none of them systematically analyze and implement the whole human conceptual induction process.

\paragraph{Machine reasoning through concepts} Instead of approaches mentioned above centered around natural language, reasoning based on concepts and taxonomic relations has been heavily discussed in the early symbolic methods for AI, particularly on commonsense reasoning. 
For example, \cite{DBLP:journals/ai/GiunchigliaW92} defines abstraction on formal systems for theorem proving, \cite{davis2015commonsense} introduces inheritance and default inheritance of attributes of entities based on description logics \cite{baader2004description}, while \cite{ramachandran2005first} represents ResearchCyc into first-order logic with quantifiers.
Particularly relevant, inheritance reasoning is a type of non-monotonic reasoning to draw conclusions about the properties of objects based on their inheritance relationships \cite{hanks1986default, brewka1987logic, horty1988mixing} through methods like shortest path reasoning~\cite{touretzky1984implicit} and most path reasoning~\cite{vogel1996human}. They focus on determining incorrect reasoning paths of exceptions, for example, the exception of ``penguins don't fly'' versus the general default inheritance of ``birds can fly''~\cite{DBLP:journals/corr/abs-2205-11658}.
Instead, we focus on the inverse process of conceptualizing an instance. 

Nevertheless, as discussed in Section~\ref{Sec:intuition}, due to the complexity and defeasibility of the real world and human cognition of this world, it is elusive to represent all such commonsense knowledge in formalized forms for symbolic reasoning, and this is the same for such formalized reasoning approaches. 
This is also supported by the recent researches that already apply our approach for commonsense modelling and reasoning on automatically constructing new training data based on our Abstract ATOMIC using one-hop conceptualization inference, such as in few-shot COMET~\cite{CAT} and zero-shot commonsense QA~\cite{CAR}.

\paragraph{Concept linking} 
Concept linking is a non-trivial task, and previous attempts of conceptualization on texts are mostly based on word matching, possibly with additional heuristics \cite{DBLP:conf/emnlp/LinCCR19}. There are also works restricted to a limited set of strict ontologies using heuristic or statistic methods \cite{DBLP:conf/www/BrauerHHLNB10,DBLP:conf/aaai/YatesGF15}. For the more general case on ConceptNet, heuristics have been developed on word chunks \cite{DBLP:conf/www/RajagopalCOK13} and linguistic features on the dependency tree \cite{DBLP:conf/cicling/PoriaAGHH14,DBLP:conf/eacl/BeckerKF21}.
As for concept linking to Probase, there are methods to use the longest covering concept \cite{DBLP:conf/cikm/SongWCW14} or statistic scoring \cite{DBLP:conf/icde/HuaWWZZ15}, and YAGO or DBPedia entities have been used as intermediate proxies \cite{DBLP:conf/cikm/ChenLXX18}. 
Besides, ComFact \cite{DBLP:conf/emnlp/GaoHKWMB22} annotates data on relevant retrieved facts based on sentence embedding similarities, while they focus on fact instead of concept linking on dialogue systems. 
Different from previous methods,
our work reaches better and more generalized concept linking by a set of sophisticated rules covering both nominal and verbal candidates, combined with strong generative and discriminative models. 

The task is also related to ultra-fine entity typing \cite{DBLP:conf/acl/LevyZCC18}, which can be achieved by various methods such as leveraging box embeddings \cite{DBLP:conf/acl/OnoeBMD20}, weak supervision from PTLMs \cite{DBLP:conf/acl/DaiSW20}, and label reasoning \cite{DBLP:conf/emnlp/LiuLXH0W21}. But instead of typing a nominal entity into free-form phrases, our goal is to link the nominal or verbal candidates to pre-defined textual concepts, so that we can leverage known taxonomic knowledge for conceptual induction.

\section{Conclusion and future work}

In this paper, we formulate the task of conceptual induction on commonsense knowledge graphs (CKGs) and design a pipeline to replicate the human conceptual induction, so that we can apply known knowledge to unseen entities and eventualities. A large-scale dataset for the validity of conceptualizations of entities and eventualities is annotated. Upon that, neural models are trained as part of the pipeline. Combined with heuristic rules, we create a large abstract CKG, and demonstrate its application in downstream tasks.

Considering the limitations of our work, future work will be focused on three aspects:

1. To improve the quality of the data, the pipeline, and the inferred abstract knowledge. As a crowd-sourced dataset, our annotated data inevitably contain errors, and the difficulty of our task amplifies this situation.
Hence it is important to further clarify the annotation criterion and clean the data accordingly. As for our pipeline, recently, large language models have been introduced to commonsense reasoning, along with various parameter-efficient training methods. Improvements in the neural model and fine-tuning strategy, as well as better formalization, might be incorporated to capture abstract knowledge with higher quality. 

2. To depict more detailed aspects in commonsense modeling and conceptualization, as various more sophisticated cases of conceptualization are yet to be explored. For example, modality, negations, and other modifiers in the context make subtle differences in the concept and inferences, while in our work, they are either ignored or conceptualized as a whole without more finer-grained analysis, such as inspecting the relationship between a concept and its negation. Also, groups of concepts with certain logical or quantifying relationships (e.g. ``doctor and nurse,'' ``a cup of tea'') are common in our language but not precisely handled in our implementation. The case is similar for event groups. An example is that one will eat only if there is food \textit{and} the person is hungry. Furthermore, more in-depth analysis of the inferences, such as probabilities for the consequences and exceptions among instances (like ``penguins can't fly'') will be desirable.

3. To apply the method in more scenarios, as the value and potential applications of our method are not fully explored in this paper. The framework can be applied to other CKGs, for example, to perform conceptualization and to form the abstract version of discourse-based commonsense knowledge as in ASER. Also, since abstract knowledge is shown to be helpful for commonsense modeling, efforts can be made to develop better methods to utilize the knowledge. For example, abstract knowledge collected through our pipeline might be directly applicable to commonsense reasoning tasks, possibly through instantiations.

\section*{Acknowledgements}
The authors of this paper were supported by the NSFC Fund (U20B2053) from the NSFC of China, as well as the RIF (R6020-19 and R6021-20) and the GRF (16211520 and 16205322) from the RGC of Hong Kong. We also thank support from the UGC Research Matching Grants (RMGS20EG01-D, RMGS20CR11, RMGS20CR12, RMGS20EG19, RMGS20EG21, RMGS23CR05, RMGS23EG08).

\clearpage
\bibliography{mybibfile}

\begin{thebibliography}{100}
\expandafter\ifx\csname url\endcsname\relax
  \def\url#1{\texttt{#1}}\fi
\expandafter\ifx\csname urlprefix\endcsname\relax\def\urlprefix{URL }\fi
\expandafter\ifx\csname href\endcsname\relax
  \def\href#1#2{#2} \def\path#1{#1}\fi

\bibitem{DBLP:conf/aaai/SapBABLRRSC19}
M.~Sap, R.~L. Bras, E.~Allaway, C.~Bhagavatula, N.~Lourie, H.~Rashkin, B.~Roof,
  N.~A. Smith, Y.~Choi, {ATOMIC:} {A}n atlas of machine commonsense for if-then
  reasoning, in: The Thirty-Third {AAAI} Conference on Artificial Intelligence,
  AAAI 2019, Honolulu, Hawaii, USA, 2019, pp. 3027--3035.

\bibitem{DBLP:conf/emnlp/ForbesHSSC20socialchemistry}
M.~Forbes, J.~D. Hwang, V.~Shwartz, M.~Sap, Y.~Choi, {Social Chemistry 101}:
  {L}earning to reason about social and moral norms, in: Proceedings of the
  2020 Conference on Empirical Methods in Natural Language Processing, {EMNLP}
  2020, Online, 2020, pp. 653--670.

\bibitem{DBLP:conf/emnlp/MostafazadehKMB20}
N.~Mostafazadeh, A.~Kalyanpur, L.~Moon, D.~W. Buchanan, L.~Berkowitz, O.~Biran,
  J.~Chu{-}Carroll, {GLUCOSE:} generalized and contextualized story
  explanations, in: Proceedings of the 2020 Conference on Empirical Methods in
  Natural Language Processing, {EMNLP} 2020, Online, 2020, pp. 4569--4586.

\bibitem{zhang2022aser}
H.~Zhang, X.~Liu, H.~Pan, H.~Ke, J.~Ou, T.~Fang, Y.~Song, {ASER}: {T}owards
  large-scale commonsense knowledge acquisition via higher-order selectional
  preference over eventualities, Artif. Intell. 309 (2022) 103740.

\bibitem{DBLP:conf/www/FangZWSH21}
T.~Fang, H.~Zhang, W.~Wang, Y.~Song, B.~He, {DISCOS:} {B}ridging the gap
  between discourse knowledge and commonsense knowledge, in: {WWW} '21: The Web
  Conference 2021, Virtual Event / Ljubljana, Slovenia, {ACM} / {IW3C2}, 2021,
  pp. 2648--2659.

\bibitem{DBLP:conf/aaai/YoungCCZBH18}
T.~Young, E.~Cambria, I.~Chaturvedi, H.~Zhou, S.~Biswas, M.~Huang, Augmenting
  end-to-end dialogue systems with commonsense knowledge, in: Proceedings of
  the Thirty-Second {AAAI} Conference on Artificial Intelligence, AAAI 2018,
  New Orleans, Louisiana, USA, 2018, pp. 4970--4977.

\bibitem{DBLP:journals/corr/abs-2212-10465}
H.~Kim, J.~Hessel, L.~Jiang, P.~West, X.~Lu, Y.~Yu, P.~Zhou, R.~L. Bras,
  M.~Alikhani, G.~Kim, M.~Sap, Y.~Choi, {SODA:} million-scale dialogue
  distillation with social commonsense contextualization, in: Proceedings of
  the 2023 Conference on Empirical Methods in Natural Language Processing,
  {EMNLP} 2023, Singapore, 2023, pp. 12930--12949.

\bibitem{DBLP:conf/aaai/SabourZH22}
S.~Sabour, C.~Zheng, M.~Huang, {CEM:} {C}ommonsense-aware empathetic response
  generation, in: Thirty-Sixth {AAAI} Conference on Artificial Intelligence,
  AAAI 2022, Virtual Event, 2022, pp. 11229--11237.

\bibitem{DBLP:conf/aaai/GabrielBSBFC21}
S.~Gabriel, C.~Bhagavatula, V.~Shwartz, R.~L. Bras, M.~Forbes, Y.~Choi,
  Paragraph-level commonsense transformers with recurrent memory, in:
  Thirty-Fifth {AAAI} Conference on Artificial Intelligence, {AAAI} 2021,
  Virtual Event, 2021, pp. 12857--12865.

\bibitem{DBLP:conf/aaai/ChenCY19}
J.~Chen, J.~Chen, Z.~Yu, Incorporating structured commonsense knowledge in
  story completion, in: The Thirty-Third {AAAI} Conference on Artificial
  Intelligence, {AAAI} 2019, Honolulu, Hawaii, USA, 2019, pp. 6244--6251.

\bibitem{DBLP:conf/acl/ZhouGSPZJ21}
Y.~Zhou, X.~Geng, T.~Shen, J.~Pei, W.~Zhang, D.~Jiang, Modeling event-pair
  relations in external knowledge graphs for script reasoning, in: Findings of
  the Association for Computational Linguistics: {ACL/IJCNLP} 2021, Online
  Event, 2021, pp. 4586--4596.

\bibitem{DBLP:conf/coling/LvZH20}
S.~Lv, F.~Zhu, S.~Hu, Integrating external event knowledge for script learning,
  in: Proceedings of the 28th International Conference on Computational
  Linguistics, {COLING} 2020, Barcelona, Spain (Online), 2020, pp. 306--315.

\bibitem{DBLP:conf/emnlp/LinCCR19}
B.~Y. Lin, X.~Chen, J.~Chen, X.~Ren, {KagNet}: {K}nowledge-aware graph networks
  for commonsense reasoning, in: Proceedings of the 2019 Conference on
  Empirical Methods in Natural Language Processing and the 9th International
  Joint Conference on Natural Language Processing, {EMNLP-IJCNLP} 2019, Hong
  Kong, China, 2019, pp. 2829--2839.

\bibitem{DBLP:conf/aaai/MaIFBNO21}
K.~Ma, F.~Ilievski, J.~Francis, Y.~Bisk, E.~Nyberg, A.~Oltramari,
  Knowledge-driven data construction for zero-shot evaluation in commonsense
  question answering, in: Thirty-Fifth {AAAI} Conference on Artificial
  Intelligence, AAAI 2021, Virtual Event, 2021, pp. 13507--13515.

\bibitem{DBLP:conf/emnlp/ShwartzWBBC20}
V.~Shwartz, P.~West, R.~L. Bras, C.~Bhagavatula, Y.~Choi, Unsupervised
  commonsense question answering with self-talk, in: Proceedings of the 2020
  Conference on Empirical Methods in Natural Language Processing, {EMNLP} 2020,
  Online, 2020, pp. 4615--4629.

\bibitem{DBLP:conf/naacl/KimKKAHY22}
Y.~J. Kim, B.~Kwak, Y.~Kim, R.~K. Amplayo, S.~Hwang, J.~Yeo, Modularized
  transfer learning with multiple knowledge graphs for zero-shot commonsense
  reasoning, in: Proceedings of the 2022 Conference of the North American
  Chapter of the Association for Computational Linguistics: Human Language
  Technologies, {NAACL-HLT} 2022, Seattle, WA, USA, 2022, pp. 2244--2257.

\bibitem{DBLP:conf/aaai/HwangBBDSBC21}
J.~D. Hwang, C.~Bhagavatula, R.~L. Bras, J.~Da, K.~Sakaguchi, A.~Bosselut,
  Y.~Choi, {(Comet-) Atomic} 2020: {O}n symbolic and neural commonsense
  knowledge graphs, in: The Thirty-Fifth {AAAI} Conference on Artificial
  Intelligence, {AAAI} 2021, Virtual Event, 2021, pp. 6384--6392.

\bibitem{DBLP:conf/aaai/SpeerCH17}
R.~Speer, J.~Chin, C.~Havasi, {ConceptNet} 5.5: {A}n open multilingual graph of
  general knowledge, in: Proceedings of the Thirty-First {AAAI} Conference on
  Artificial Intelligence, AAAI 2017, San Francisco, California, {USA}, 2017,
  pp. 4444--4451.

\bibitem{wu2012probase}
W.~Wu, H.~Li, H.~Wang, K.~Q. Zhu, Probase: {A} probabilistic taxonomy for text
  understanding, in: Proceedings of the {ACM} {SIGMOD} International Conference
  on Management of Data, {SIGMOD} 2012, Scottsdale, AZ, USA, {ACM}, 2012, pp.
  481--492.

\bibitem{DBLP:conf/acl/BosselutRSMCC19}
A.~Bosselut, H.~Rashkin, M.~Sap, C.~Malaviya, A.~Celikyilmaz, Y.~Choi, {COMET:}
  {C}ommonsense transformers for automatic knowledge graph construction, in:
  Proceedings of the 57th Conference of the Association for Computational
  Linguistics, {ACL} 2019, Florence, Italy, 2019, pp. 4762--4779.

\bibitem{radford2019gpt2}
A.~Radford, J.~Wu, R.~Child, D.~Luan, D.~Amodei, I.~Sutskever, Language models
  are unsupervised multitask learners, Tech. rep., OpenAI (2019).

\bibitem{DBLP:conf/acl/LewisLGGMLSZ20}
M.~Lewis, Y.~Liu, N.~Goyal, M.~Ghazvininejad, A.~Mohamed, O.~Levy, V.~Stoyanov,
  L.~Zettlemoyer, {BART:} {D}enoising sequence-to-sequence pre-training for
  natural language generation, translation, and comprehension, in: Proceedings
  of the 58th Annual Meeting of the Association for Computational Linguistics,
  {ACL} 2020, Online, 2020, pp. 7871--7880.

\bibitem{DBLP:conf/acl/WangICR21}
P.~Wang, F.~Ilievski, M.~Chen, X.~Ren, Do language models perform generalizable
  commonsense inference?, in: Findings of the Association for Computational
  Linguistics: {ACL/IJCNLP} 2021, Online Event, 2021, pp. 3681--3688.

\bibitem{DBLP:conf/emnlp/ZhouKLLHPR21}
P.~Zhou, R.~Khanna, S.~Lee, B.~Y. Lin, D.~Ho, J.~Pujara, X.~Ren, {RICA:}
  {E}valuating robust inference capabilities based on commonsense axioms, in:
  Proceedings of the 2021 Conference on Empirical Methods in Natural Language
  Processing, {EMNLP} 2021, Virtual Event / Punta Cana, Dominican Republic,
  2021, pp. 7560--7579.

\bibitem{murphy2004concept1}
G.~Murphy, Introduction, in: The big book of concepts, MIT press, 2004, Ch.~1,
  pp. 1--10.

\bibitem{tenenbaum2011grow}
J.~B. Tenenbaum, C.~Kemp, T.~L. Griffiths, N.~D. Goodman, How to grow a mind:
  {S}tatistics, structure, and abstraction, Science 331~(6022) (2011)
  1279--1285.

\bibitem{murphy2004concept8}
G.~Murphy, Induction, in: The big book of concepts, MIT press, 2004, Ch.~8, pp.
  243--270.

\bibitem{ramachandran2005first}
D.~Ramachandran, P.~Reagan, K.~Goolsbey, First-orderized {R}esearch{C}yc:
  {E}xpressivity and efficiency in a common-sense ontology, in: AAAI workshop
  on contexts and ontologies: theory, practice and applications, 2005, pp.
  33--40.

\bibitem{DBLP:conf/naacl/PoradaSTC21}
I.~Porada, K.~Suleman, A.~Trischler, J.~C.~K. Cheung, Modeling event
  plausibility with consistent conceptual abstraction, in: Proceedings of the
  2021 Conference of the North American Chapter of the Association for
  Computational Linguistics: Human Language Technologies, {NAACL-HLT} 2021,
  Online, 2021, pp. 1732--1743.

\bibitem{bach1986algebra}
E.~Bach, The algebra of events, Linguistics and philosophy (1986) 5--16.

\bibitem{mourelatos1978events}
A.~P. Mourelatos, Events, processes, and states, Linguistics and philosophy 2
  (1978) 415--434.

\bibitem{liu2004conceptnet}
H.~Liu, P.~Singh, {ConceptNet}—{a} practical commonsense reasoning tool-kit,
  BT technology journal 22~(4) (2004) 211--226.

\bibitem{DBLP:journals/jair/Davis17}
E.~Davis, Logical formalizations of commonsense reasoning: {A} survey, J.
  Artif. Intell. Res. 59 (2017) 651--723.

\bibitem{DBLP:conf/aaai/GordonDS10}
J.~Gordon, B.~V. Durme, L.~K. Schubert, Learning from the {W}eb: Extracting
  general world knowledge from noisy text, in: Collaboratively-Built Knowledge
  Sources and Artificial Intelligence, Papers from the 2010 {AAAI} Workshop,
  Atlanta, Georgia, USA, Vol. {WS-10-02} of {AAAI} Technical Report, {AAAI},
  2010.

\bibitem{choi2022curious}
Y.~Choi, The curious case of commonsense intelligence, Daedalus 151~(2) (2022)
  139--155.

\bibitem{grice1975logic}
H.~P. Grice, Logic and conversation, in: Speech acts, Brill, 1975, pp. 41--58.

\bibitem{DBLP:conf/cikm/GordonD13}
J.~Gordon, B.~V. Durme, Reporting bias and knowledge acquisition, in:
  Proceedings of the 2013 workshop on Automated Knowledge Base Construction,
  AKBC@CIKM 13, San Francisco, California, USA, {ACM}, 2013, pp. 25--30.

\bibitem{DBLP:conf/coling/ShwartzC20}
V.~Shwartz, Y.~Choi, Do neural language models overcome reporting bias?, in:
  Proceedings of the 28th International Conference on Computational
  Linguistics, {COLING} 2020, Barcelona, Spain (Online), International
  Committee on Computational Linguistics, 2020, pp. 6863--6870.

\bibitem{DBLP:conf/semweb/AuerBKLCI07dbpedia}
S.~Auer, C.~Bizer, G.~Kobilarov, J.~Lehmann, R.~Cyganiak, Z.~G. Ives,
  {DBpedia}: {A} nucleus for a web of open data, in: The Semantic Web, 6th
  International Semantic Web Conference, 2nd Asian Semantic Web Conference,
  {ISWC} 2007 + {ASWC} 2007, Busan, Korea, Vol. 4825 of Lecture Notes in
  Computer Science, Springer, 2007, pp. 722--735.

\bibitem{DBLP:journals/cacm/Lenat95CYC}
D.~B. Lenat, {CYC:} {A} large-scale investment in knowledge infrastructure,
  Commun. {ACM} 38~(11) (1995) 32--38.

\bibitem{gordon2017formal}
A.~S. Gordon, J.~R. Hobbs, A formal theory of commonsense psychology: How
  people think people think, Cambridge University Press, 2017.

\bibitem{miller1998wordnet}
G.~A. Miller, WordNet: {A}n electronic lexical database, MIT press, 1998.

\bibitem{devlin2018bert}
J.~Devlin, M.~Chang, K.~Lee, K.~Toutanova, {BERT:} {P}re-training of deep
  bidirectional transformers for language understanding, in: Proceedings of the
  2019 Conference of the North American Chapter of the Association for
  Computational Linguistics: Human Language Technologies, {NAACL-HLT} 2019,
  Minneapolis, MN, USA, 2019, pp. 4171--4186.

\bibitem{radford2018improving}
A.~Radford, K.~Narasimhan, T.~Salimans, I.~Sutskever, Improving language
  understanding by generative pre-training, Tech. rep., OpenAI (2018).

\bibitem{brown2020language}
T.~B. Brown, B.~Mann, N.~Ryder, M.~Subbiah, J.~Kaplan, P.~Dhariwal,
  A.~Neelakantan, P.~Shyam, G.~Sastry, A.~Askell, S.~Agarwal,
  A.~Herbert{-}Voss, G.~Krueger, T.~Henighan, R.~Child, A.~Ramesh, D.~M.
  Ziegler, J.~Wu, C.~Winter, C.~Hesse, M.~Chen, E.~Sigler, M.~Litwin, S.~Gray,
  B.~Chess, J.~Clark, C.~Berner, S.~McCandlish, A.~Radford, I.~Sutskever,
  D.~Amodei, Language models are few-shot learners, in: Advances in Neural
  Information Processing Systems 33, NeurIPS 2020, virtual, 2020.

\bibitem{davison2019commonsense}
J.~Davison, J.~Feldman, A.~M. Rush, Commonsense knowledge mining from
  pretrained models, in: Proceedings of the 2019 Conference on Empirical
  Methods in Natural Language Processing and the 9th International Joint
  Conference on Natural Language Processing, {EMNLP-IJCNLP} 2019, Hong Kong,
  China, 2019, pp. 1173--1178.

\bibitem{shi2021transformers}
H.~Shi, P.~Wolff, What transformers might know about the physical world: {T}5
  and the origins of knowledge, in: Proceedings of the 43th Annual Meeting of
  the Cognitive Science Society, CogSci 2021, virtual, 2021.

\bibitem{DBLP:conf/cogsci/WeirPD20}
N.~Weir, A.~Poliak, B.~V. Durme, Probing neural language models for human tacit
  assumptions, in: Proceedings of the 42th Annual Meeting of the Cognitive
  Science Society, CogSci 2020, virtual, 2020.

\bibitem{DBLP:conf/iclr/WangSMHLB19}
A.~Wang, A.~Singh, J.~Michael, F.~Hill, O.~Levy, S.~R. Bowman, {GLUE:} {A}
  multi-task benchmark and analysis platform for natural language
  understanding, in: 7th International Conference on Learning Representations,
  {ICLR} 2019, New Orleans, LA, USA, 2019.

\bibitem{DBLP:conf/emnlp/DuDLL19}
L.~Du, X.~Ding, T.~Liu, Z.~Li, Modeling event background for if-then
  commonsense reasoning using context-aware variational autoencoder, in:
  Proceedings of the 2019 Conference on Empirical Methods in Natural Language
  Processing and the 9th International Joint Conference on Natural Language
  Processing, {EMNLP-IJCNLP} 2019, Hong Kong, China, 2019, pp. 2682--2691.

\bibitem{DBLP:journals/corr/abs-2008-05925}
C.~Wang, J.~Wu, L.~Liu, Y.~Zhang, Commonsense knowledge graph reasoning by
  selection or generation? {W}hy?, CoRR abs/2008.05925.

\bibitem{DBLP:conf/cogsci/ForbesHC19}
M.~Forbes, A.~Holtzman, Y.~Choi, Do neural language representations learn
  physical commonsense?, in: Proceedings of the 41th Annual Meeting of the
  Cognitive Science Society , CogSci 2019, Montreal, Canada, 2019, pp.
  1753--1759.

\bibitem{da2019cracking}
J.~Da, J.~Kasai, Cracking the contextual commonsense code: {U}nderstanding
  commonsense reasoning aptitude of deep contextual representations, in:
  Proceedings of the First Workshop on Commonsense Inference in Natural
  Language Processing, 2019, pp. 1--12.

\bibitem{DBLP:journals/corr/abs-2203-08452}
Q.~He, S.~Cheng, Z.~Li, R.~Xie, Y.~Xiao, Can pre-trained language models
  interpret similes as smart as human?, in: Proceedings of the 60th Annual
  Meeting of the Association for Computational Linguistics, {ACL} 2022, Dublin,
  Ireland, 2022, pp. 7875--7887.

\bibitem{CAR}
W.~Wang, T.~Fang, W.~Ding, B.~Xu, X.~Liu, Y.~Song, A.~Bosselut, {CAR:}
  conceptualization-augmented reasoner for zero-shot commonsense question
  answering, in: Findings of the Association for Computational Linguistics:
  {EMNLP} 2023, Singapore, 2023, pp. 13520--13545.

\bibitem{DBLP:journals/tacl/Ettinger20}
A.~Ettinger, What {BERT} is not: {L}essons from a new suite of psycholinguistic
  diagnostics for language models, Trans. Assoc. Comput. Linguistics 8 (2020)
  34--48.

\bibitem{DBLP:conf/acl/KassnerS20}
N.~Kassner, H.~Sch{\"{u}}tze, Negated and misprimed probes for pretrained
  language models: {B}irds can talk, but cannot fly, in: Proceedings of the
  58th Annual Meeting of the Association for Computational Linguistics, {ACL}
  2020, Online, 2020, pp. 7811--7818.

\bibitem{DBLP:conf/emnlp/PaikARK21}
C.~Paik, S.~Aroca{-}Ouellette, A.~Roncone, K.~Kann, The world of an octopus:
  {H}ow reporting bias influences a language model's perception of color, in:
  Proceedings of the 2021 Conference on Empirical Methods in Natural Language
  Processing, {EMNLP} 2021, Virtual Event / Punta Cana, Dominican Republic,
  2021, pp. 823--835.

\bibitem{DBLP:conf/emnlp/PengWHJ0L0022}
H.~Peng, X.~Wang, S.~Hu, H.~Jin, L.~Hou, J.~Li, Z.~Liu, Q.~Liu, {COPEN:}
  {P}robing conceptual knowledge in pre-trained language models, in:
  Proceedings of the 2022 Conference on Empirical Methods in Natural Language
  Processing, {EMNLP} 2022, Abu Dhabi, United Arab Emirates, 2022, pp.
  5015--5035.

\bibitem{DBLP:conf/acl/WuJJXT23}
W.~Wu, C.~Jiang, Y.~Jiang, P.~Xie, K.~Tu, Do {PLMs} know and understand
  ontological knowledge?, in: Proceedings of the 61st Annual Meeting of the
  Association for Computational Linguistics, {ACL} 2023, Toronto, Canada, 2023,
  pp. 3080--3101.

\bibitem{DBLP:journals/aim/ForbusH17}
K.~D. Forbus, T.~Hinrich, Analogy and relational representations in the
  companion cognitive architecture, {AI} Mag. 38~(4) (2017) 34--42.

\bibitem{murphy2004concept3}
G.~Murphy, Theories, in: The big book of concepts, MIT press, 2004, Ch.~3, pp.
  41--72.

\bibitem{murphy2004concept2}
G.~Murphy, Typicality and the classical view of categories, in: The big book of
  concepts, MIT press, 2004, Ch.~2, pp. 11--40.

\bibitem{murphy2004concept7}
G.~Murphy, Taxonomic organization and the basic level of concepts, in: The big
  book of concepts, MIT press, 2004, Ch.~7, pp. 199--242.

\bibitem{minsky1980k}
M.~Minsky, {K-Lines: A} theory of memory, Cognitive science 4~(2) (1980)
  117--133.

\bibitem{DBLP:journals/jis/Hajibayova13}
L.~Hajibayova, Basic-level categories: {A} review, J. Inf. Sci. 39~(5) (2013)
  676--687.

\bibitem{sep-meaning}
J.~Speaks, {Theories of Meaning}, in: The {Stanford} Encyclopedia of
  Philosophy, {S}pring 2021 Edition, Metaphysics Research Lab, Stanford
  University, 2021.

\bibitem{davidson1967logical}
D.~Davidson, N.~Rescher, The logical form of action sentences, 1967 (1967)
  105--122.

\bibitem{DBLP:journals/ai/GiunchigliaW92}
F.~Giunchiglia, T.~Walsh, A theory of abstraction, Artif. Intell. 57~(2-3)
  (1992) 323--389.

\bibitem{davis2015commonsense}
E.~Davis, G.~Marcus, Commonsense reasoning and commonsense knowledge in
  artificial intelligence, Communications of the ACM 58~(9) (2015) 92--103.

\bibitem{DBLP:conf/eacl/DurmeMS09}
B.~V. Durme, P.~Michalak, L.~K. Schubert, Deriving generalized knowledge from
  corpora using wordnet abstraction, in: {EACL} 2009, 12th Conference of the
  European Chapter of the Association for Computational Linguistics, Athens,
  Greece, 2009, pp. 808--816.

\bibitem{honnibal2017spacy}
M.~Honnibal, I.~Montani, spacy 2: {N}atural language understanding with {B}loom
  embeddings, convolutional neural networks and incremental parsing.

\bibitem{DBLP:conf/naacl/MeyersRMSZYG04}
A.~L. Meyers, R.~Reeves, C.~Macleod, R.~Szekely, V.~Zielinska, B.~Young,
  R.~Grishman, The nombank project: {A}n interim report, in: Proceedings of the
  Workshop Frontiers in Corpus Annotation@HLT-NAACL 2004, Boston, MA, USA,
  2004.

\bibitem{DBLP:conf/emnlp/HuangSQH19}
L.~Huang, C.~Sun, X.~Qiu, X.~Huang, {GlossBERT: BERT} for word sense
  disambiguation with gloss knowledge, in: Proceedings of the 2019 Conference
  on Empirical Methods in Natural Language Processing and the 9th International
  Joint Conference on Natural Language Processing, {EMNLP-IJCNLP} 2019, Hong
  Kong, China, 2019, pp. 3507--3512.

\bibitem{DBLP:journals/corr/abs-1907-11692}
Y.~Liu, M.~Ott, N.~Goyal, J.~Du, M.~Joshi, D.~Chen, O.~Levy, M.~Lewis,
  L.~Zettlemoyer, V.~Stoyanov, {RoBERTa}: {A} robustly optimized {BERT}
  pretraining approach, CoRR abs/1907.11692.

\bibitem{DBLP:conf/emnlp/FangWCHZSH21}
T.~Fang, W.~Wang, S.~Choi, S.~Hao, H.~Zhang, Y.~Song, B.~He, Benchmarking
  commonsense knowledge base population with an effective evaluation dataset,
  in: Proceedings of the 2021 Conference on Empirical Methods in Natural
  Language Processing, {EMNLP} 2021, Virtual Event / Punta Cana, Dominican
  Republic, 2021, pp. 8949--8964.

\bibitem{DBLP:journals/corr/abs-2304-10392}
T.~Fang, Q.~V. Do, S.~Choi, W.~Wang, Y.~Song, {CKBP} v2: An expert-annotated
  evaluation set for commonsense knowledge base population, CoRR
  abs/2304.10392.

\bibitem{DBLP:conf/iclr/BhagavatulaBMSH20}
C.~Bhagavatula, R.~L. Bras, C.~Malaviya, K.~Sakaguchi, A.~Holtzman, H.~Rashkin,
  D.~Downey, W.~Yih, Y.~Choi, Abductive commonsense reasoning, in: 8th
  International Conference on Learning Representations, {ICLR} 2020, Addis
  Ababa, Ethiopia, 2020.

\bibitem{DBLP:conf/naacl/TalmorHLB19}
A.~Talmor, J.~Herzig, N.~Lourie, J.~Berant, Commonsense{QA}: {A} question
  answering challenge targeting commonsense knowledge, in: Proceedings of the
  2019 Conference of the North American Chapter of the Association for
  Computational Linguistics: Human Language Technologies, {NAACL-HLT} 2019,
  Minneapolis, MN, USA, 2019, pp. 4149--4158.

\bibitem{DBLP:conf/aaai/BiskZLGC20}
Y.~Bisk, R.~Zellers, R.~L. Bras, J.~Gao, Y.~Choi, {PIQA:} {R}easoning about
  physical commonsense in natural language, in: The Thirty-Fourth {AAAI}
  Conference on Artificial Intelligence, {AAAI} 2020, New York, NY, USA, 2020,
  2020, pp. 7432--7439.

\bibitem{DBLP:conf/emnlp/SapRCBC19}
M.~Sap, H.~Rashkin, D.~Chen, R.~L. Bras, Y.~Choi, {Social IQa}: {C}ommonsense
  reasoning about social interactions, in: K.~Inui, J.~Jiang, V.~Ng, X.~Wan
  (Eds.), Proceedings of the 2019 Conference on Empirical Methods in Natural
  Language Processing and the 9th International Joint Conference on Natural
  Language Processing, {EMNLP-IJCNLP} 2019, Hong Kong, China, 2019, 2019, pp.
  4462--4472.

\bibitem{DBLP:conf/aaai/SakaguchiBBC20}
K.~Sakaguchi, R.~L. Bras, C.~Bhagavatula, Y.~Choi, Wino{G}rande: {A}n
  adversarial winograd schema challenge at scale, in: The Thirty-Fourth {AAAI}
  Conference on Artificial Intelligence, {AAAI} 2020, New York, NY, USA, 2020,
  2020, pp. 8732--8740.

\bibitem{he2023debertav}
P.~He, J.~Gao, W.~Chen, {DeBERTaV3}: {I}mproving {DeBERTa} using
  {ELECTRA}-style pre-training with gradient-disentangled embedding sharing,
  in: The Eleventh International Conference on Learning Representations, {ICLR}
  2023, Kigali, Rwanda, 2023.

\bibitem{DBLP:conf/aaai/BosselutBC21}
A.~Bosselut, R.~L. Bras, Y.~Choi, Dynamic neuro-symbolic knowledge graph
  construction for zero-shot commonsense question answering, in: Thirty-Fifth
  {AAAI} Conference on Artificial Intelligence, AAAI 2021, Virtual Event, 2021,
  pp. 4923--4931.

\bibitem{DBLP:conf/emnlp/BanerjeeB20}
P.~Banerjee, C.~Baral, Self-supervised knowledge triplet learning for zero-shot
  question answering, in: Proceedings of the 2020 Conference on Empirical
  Methods in Natural Language Processing, {EMNLP} 2020, Online, 2020, pp.
  151--162.

\bibitem{DBLP:conf/emnlp/SuWFZSZ22}
Y.~Su, Z.~Wang, T.~Fang, H.~Zhang, Y.~Song, T.~Zhang, {MICO:} {A}
  multi-alternative contrastive learning framework for commonsense knowledge
  representation, in: Findings of the Association for Computational
  Linguistics: {EMNLP} 2022, Abu Dhabi, United Arab Emirates, 2022, pp.
  1339--1351.

\bibitem{DBLP:conf/acl/HeUGP23}
J.~He, S.~C.~L. U, V.~Guti{\'{e}}rrez{-}Basulto, J.~Z. Pan, {BUCA:} {A} binary
  classification approach to unsupervised commonsense question answering, in:
  Proceedings of the 61st Annual Meeting of the Association for Computational
  Linguistics, {ACL} 2023, Toronto, Canada, 2023, pp. 376--387.

\bibitem{DBLP:conf/naacl/WestBHHJBLWC22}
P.~West, C.~Bhagavatula, J.~Hessel, J.~D. Hwang, L.~Jiang, R.~L. Bras, X.~Lu,
  S.~Welleck, Y.~Choi, Symbolic knowledge distillation: from general language
  models to commonsense models, in: Proceedings of the 2022 Conference of the
  North American Chapter of the Association for Computational Linguistics:
  Human Language Technologies, {NAACL-HLT} 2022, Seattle, WA, USA, 2022, pp.
  4602--4625.

\bibitem{DBLP:conf/acl/LiWDWX20}
Z.~Li, W.~Wang, L.~Dong, F.~Wei, K.~Xu, Harvesting and refining question-answer
  pairs for unsupervised {QA}, in: Proceedings of the 58th Annual Meeting of
  the Association for Computational Linguistics, {ACL} 2020, Online, 2020, pp.
  6719--6728.

\bibitem{DBLP:conf/aaai/DouP22}
Z.~Dou, N.~Peng, Zero-shot commonsense question answering with cloze
  translation and consistency optimization, in: Thirty-Sixth {AAAI} Conference
  on Artificial Intelligence, AAAI 2022, Virtual Event, 2022, pp. 10572--10580.

\bibitem{CANDLE}
W.~Wang, T.~Fang, C.~Li, H.~Shi, W.~Ding, B.~Xu, Z.~Wang, J.~Bai, X.~Liu,
  J.~Cheng, C.~Chan, Y.~Song, {CANDLE:} {I}terative conceptualization and
  instantiation distillation from large language models for commonsense
  reasoning, CoRR abs/2401.07286.

\bibitem{DBLP:journals/corr/abs-2403-07398}
T.~Fang, Z.~Chen, Y.~Song, A.~Bosselut, Complex reasoning over logical queries
  on commonsense knowledge graphs, CoRR abs/2403.07398.

\bibitem{DBLP:conf/nips/Ouyang0JAWMZASR22}
L.~Ouyang, J.~Wu, X.~Jiang, D.~Almeida, C.~L. Wainwright, P.~Mishkin, C.~Zhang,
  S.~Agarwal, K.~Slama, A.~Ray, J.~Schulman, J.~Hilton, F.~Kelton, L.~Miller,
  M.~Simens, A.~Askell, P.~Welinder, P.~F. Christiano, J.~Leike, R.~Lowe,
  Training language models to follow instructions with human feedback, in:
  Advances in Neural Information Processing Systems 35, NeurIPS 2022, New
  Orleans, LA, USA, 2022.

\bibitem{Lenat1998}
D.~Lenat, The dimensions of context space, Technical report, Cycorp (1998).

\bibitem{DBLP:conf/coopis/SinghLMLPZ02OMCS}
P.~Singh, T.~Lin, E.~T. Mueller, G.~Lim, T.~Perkins, W.~L. Zhu, {Open Mind
  Common Sense}: {K}nowledge acquisition from the general public, in: 2002
  Confederated International Conferences DOA, CoopIS and {ODBASE} , Irvine,
  California, USA, Vol. 2519 of Lecture Notes in Computer Science, Springer,
  2002, pp. 1223--1237.

\bibitem{DBLP:conf/acl/SmithCSRA18}
H.~Rashkin, M.~Sap, E.~Allaway, N.~A. Smith, Y.~Choi, Event2mind: {C}ommonsense
  inference on events, intents, and reactions, in: Proceedings of the 56th
  Annual Meeting of the Association for Computational Linguistics, {ACL} 2018,
  Melbourne, Australia, 2018, pp. 463--473.

\bibitem{DBLP:conf/acl/GhosalSMMP22}
D.~Ghosal, S.~Shen, N.~Majumder, R.~Mihalcea, S.~Poria, {CICERO:} {A} dataset
  for contextualized commonsense inference in dialogues, in: Proceedings of the
  60th Annual Meeting of the Association for Computational Linguistics, {ACL}
  2022, Dublin, Ireland, 2022, pp. 5010--5028.

\bibitem{DBLP:journals/corr/abs-2305-02364}
S.~Gao, B.~Borges, S.~Oh, D.~Bayazit, S.~Kanno, H.~Wakaki, Y.~Mitsufuji,
  A.~Bosselut, {PeaCoK}: {P}ersona commonsense knowledge for consistent and
  engaging narratives, in: Proceedings of the 61st Annual Meeting of the
  Association for Computational Linguistics, {ACL} 2023, Toronto, Canada, 2023,
  pp. 6569--6591.

\bibitem{DBLP:conf/naacl/SapSVZCS22}
M.~Sap, S.~Swayamdipta, L.~Vianna, X.~Zhou, Y.~Choi, N.~A. Smith, Annotators
  with attitudes: {H}ow annotator beliefs and identities bias toxic language
  detection, in: Proceedings of the 2022 Conference of the North American
  Chapter of the Association for Computational Linguistics: Human Language
  Technologies, {NAACL-HLT} 2022, Seattle, WA, USA, 2022, pp. 5884--5906.

\bibitem{DBLP:conf/wsdm/TandonMSW14}
N.~Tandon, G.~de~Melo, F.~M. Suchanek, G.~Weikum, {WebChild}: {H}arvesting and
  organizing commonsense knowledge from the web, in: Seventh {ACM}
  International Conference on Web Search and Data Mining, {WSDM} 2014, New
  York, USA, {ACM}, 2014, pp. 523--532.

\bibitem{DBLP:conf/cikm/TandonMDW15}
N.~Tandon, G.~de~Melo, A.~De, G.~Weikum, Knowlywood: {M}ining activity
  knowledge from hollywood narratives, in: Proceedings of the 24th {ACM}
  International Conference on Information and Knowledge Management, {CIKM}
  2015, Melbourne, Australia, {ACM}, 2015, pp. 223--232.

\bibitem{zhang2019aser}
H.~Zhang, X.~Liu, H.~Pan, Y.~Song, C.~W. Leung, {ASER:} {A} large-scale
  eventuality knowledge graph, in: {WWW} '20: The Web Conference 2020, Taipei,
  Taiwan, {ACM} / {IW3C2}, 2020, pp. 201--211.

\bibitem{petroni2019lama}
F.~Petroni, T.~Rockt{\"{a}}schel, S.~Riedel, P.~S.~H. Lewis, A.~Bakhtin, Y.~Wu,
  A.~H. Miller, Language models as knowledge bases?, in: Proceedings of the
  2019 Conference on Empirical Methods in Natural Language Processing and the
  9th International Joint Conference on Natural Language Processing,
  {EMNLP-IJCNLP} 2019, Hong Kong, China, 2019, pp. 2463--2473.

\bibitem{DBLP:journals/tacl/JiangXAN20}
Z.~Jiang, F.~F. Xu, J.~Araki, G.~Neubig, How can we know what language models
  know, Trans. Assoc. Comput. Linguistics 8 (2020) 423--438.

\bibitem{DBLP:conf/emnlp/ShinRLWS20}
T.~Shin, Y.~Razeghi, R.~L.~L. IV, E.~Wallace, S.~Singh, {AutoPrompt}:
  {E}liciting knowledge from language models with automatically generated
  prompts, in: Proceedings of the 2020 Conference on Empirical Methods in
  Natural Language Processing, {EMNLP} 2020, Online, 2020, pp. 4222--4235.

\bibitem{DBLP:conf/naacl/ZhongFC21}
Z.~Zhong, D.~Friedman, D.~Chen, Factual probing is {[MASK]:} learning vs.
  learning to recall, in: Proceedings of the 2021 Conference of the North
  American Chapter of the Association for Computational Linguistics: Human
  Language Technologies, {NAACL-HLT} 2021, Online, 2021, pp. 5017--5033.

\bibitem{DBLP:conf/acl/TamborrinoPPVN20}
A.~Tamborrino, N.~Pellican{\`{o}}, B.~Pannier, P.~Voitot, L.~Naudin,
  Pre-training is (almost) all you need: {A}n application to commonsense
  reasoning, in: Proceedings of the 58th Annual Meeting of the Association for
  Computational Linguistics, {ACL} 2020, Online, 2020, pp. 3878--3887.

\bibitem{DBLP:journals/tnn/JiPCMY22}
S.~Ji, S.~Pan, E.~Cambria, P.~Marttinen, P.~S. Yu, A survey on knowledge
  graphs: Representation, acquisition, and applications, {IEEE} Trans. Neural
  Networks Learn. Syst. 33~(2) (2022) 494--514.

\bibitem{DBLP:conf/acl/LiTTG16}
X.~Li, A.~Taheri, L.~Tu, K.~Gimpel, Commonsense knowledge base completion, in:
  Proceedings of the 54th Annual Meeting of the Association for Computational
  Linguistics, {ACL} 2016, Berlin, Germany, 2016.

\bibitem{yao2019kgbert}
L.~Yao, C.~Mao, Y.~Luo, {KG-BERT:} {BERT} for knowledge graph completion, CoRR
  abs/1909.03193.

\bibitem{DBLP:conf/acl/MoghimifarQZHB20}
F.~Moghimifar, L.~Qu, T.~Y. Zhuo, G.~Haffari, M.~Baktashmotlagh,
  Neural-symbolic commonsense reasoner with relation predictors, in:
  Proceedings of the 59th Annual Meeting of the Association for Computational
  Linguistics and the 11th International Joint Conference on Natural Language
  Processing, {ACL/IJCNLP} 2021, Virtual Event, 2021, pp. 797--802.

\bibitem{DBLP:conf/www/WangSLZW021}
B.~Wang, T.~Shen, G.~Long, T.~Zhou, Y.~Wang, Y.~Chang, Structure-augmented text
  representation learning for efficient knowledge graph completion, in: {WWW}
  '21: The Web Conference 2021, Virtual Event / Ljubljana, Slovenia, {ACM} /
  {IW3C2}, 2021, pp. 1737--1748.

\bibitem{huang2021structured}
J.~Huang, Y.~Du, S.~Tao, K.~Xu, P.~Xie, Structured self-supervised pretraining
  for commonsense knowledge graph completion, Trans. Assoc. Comput. Linguistics
  9 (2021) 1268--1284.

\bibitem{DBLP:conf/acl/LovelaceNVLR20}
J.~Lovelace, D.~Newman{-}Griffis, S.~Vashishth, J.~F. Lehman, C.~P. Ros{\'{e}},
  Robust knowledge graph completion with stacked convolutions and a student
  re-ranking network, in: Proceedings of the 59th Annual Meeting of the
  Association for Computational Linguistics and the 11th International Joint
  Conference on Natural Language Processing, {ACL/IJCNLP} 2021, Virtual Event,
  2021, pp. 1016--1029.

\bibitem{malaviya2019ckgcompletion}
C.~Malaviya, C.~Bhagavatula, A.~Bosselut, Y.~Choi, Exploiting structural and
  semantic context for commonsense knowledge base completion, CoRR
  abs/1910.02915.

\bibitem{DBLP:conf/ijcnn/WangWHYLK21}
B.~Wang, G.~Wang, J.~Huang, J.~You, J.~Leskovec, C.~J. Kuo, Inductive learning
  on commonsense knowledge graph completion, in: International Joint Conference
  on Neural Networks, {IJCNN} 2021, Shenzhen, China, {IEEE}, 2021, pp. 1--8.

\bibitem{DBLP:conf/conll/SaitoNAT18}
I.~Saito, K.~Nishida, H.~Asano, J.~Tomita, Commonsense knowledge base
  completion and generation, in: Proceedings of the 22nd Conference on
  Computational Natural Language Learning, CoNLL 2018, Brussels, Belgium, 2018,
  pp. 141--150.

\bibitem{DBLP:conf/ijcai/ZhangKSR20}
H.~Zhang, D.~Khashabi, Y.~Song, D.~Roth, {TransOMCS}: {F}rom linguistic graphs
  to commonsense knowledge, in: Proceedings of the Twenty-Ninth International
  Joint Conference on Artificial Intelligence, {IJCAI} 2020, 2020, pp.
  4004--4010.

\bibitem{DBLP:conf/acl/JiG11}
H.~Ji, R.~Grishman, Knowledge base population: {S}uccessful approaches and
  challenges, in: The 49th Annual Meeting of the Association for Computational
  Linguistics: Human Language Technologies, ACL-HLT 2011, Portland, Oregon,
  {USA}, 2011, pp. 1148--1158.

\bibitem{DBLP:journals/kbs/IlievskiOMZMS21}
F.~Ilievski, A.~Oltramari, K.~Ma, B.~Zhang, D.~L. McGuinness, P.~A. Szekely,
  Dimensions of commonsense knowledge, Knowl. Based Syst. 229 (2021) 107347.

\bibitem{DBLP:conf/esws/IlievskiSZ21}
F.~Ilievski, P.~A. Szekely, B.~Zhang, {CSKG:} the commonsense knowledge graph,
  in: The Semantic Web - 18th International Conference, {ESWC} 2021, Virtual
  Event, Vol. 12731 of Lecture Notes in Computer Science, Springer, 2021, pp.
  680--696.

\bibitem{DBLP:conf/sigmod/BollackerEPST08}
K.~D. Bollacker, C.~Evans, P.~K. Paritosh, T.~Sturge, J.~Taylor, Freebase: {A}
  collaboratively created graph database for structuring human knowledge, in:
  Proceedings of the {ACM} {SIGMOD} International Conference on Management of
  Data, {SIGMOD} 2008, Vancouver, BC, Canada, {ACM}, 2008, pp. 1247--1250.

\bibitem{DBLP:conf/www/SuchanekKW07}
F.~M. Suchanek, G.~Kasneci, G.~Weikum, Yago: {A} core of semantic knowledge,
  in: Proceedings of the 16th International Conference on World Wide Web, {WWW}
  2007, Banff, Alberta, Canada, {ACM}, 2007, pp. 697--706.

\bibitem{DBLP:conf/semweb/IlievskiSS20}
F.~Ilievski, P.~A. Szekely, D.~Schwabe, Commonsense knowledge in wikidata, in:
  Proceedings of the 1st Wikidata Workshop (Wikidata 2020) co-located with 19th
  International Semantic Web Conference(OPub 2020), Virtual Conference, Vol.
  2773 of {CEUR} Workshop Proceedings, CEUR-WS.org, 2020.

\bibitem{schank1975scripts}
R.~C. Schank, R.~P. Abelson, Scripts, plans and knowledge, in: Advance Papers
  of the Fourth International Joint Conference on Artificial Intelligence,
  Tbilisi, Georgia, USSR, 1975, pp. 151--157.

\bibitem{DBLP:conf/acl/RegneriKP10}
M.~Regneri, A.~Koller, M.~Pinkal, Learning script knowledge with web
  experiments, in: {ACL} 2010, Proceedings of the 48th Annual Meeting of the
  Association for Computational Linguistics, Uppsala, Sweden, 2010, pp.
  979--988.

\bibitem{barr1981handbook}
A.~Barr, E.~A. Feigenbaum, P.~R. Cohen, The handbook of artificial
  intelligence, Vol.~3, HeurisTech Press, 1981.

\bibitem{DBLP:conf/acl/RothWN18}
Q.~Ning, H.~Wu, D.~Roth, A multi-axis annotation scheme for event temporal
  relations, in: Proceedings of the 56th Annual Meeting of the Association for
  Computational Linguistics, {ACL} 2018, Melbourne, Australia, 2018, pp.
  1318--1328.

\bibitem{DBLP:conf/ijcai/LiDL18}
Z.~Li, X.~Ding, T.~Liu, Constructing narrative event evolutionary graph for
  script event prediction, in: Proceedings of the Twenty-Seventh International
  Joint Conference on Artificial Intelligence, {IJCAI} 2018, Stockholm, Sweden,
  ijcai.org, 2018, pp. 4201--4207.

\bibitem{DBLP:journals/tacl/GuanHHZZ20}
J.~Guan, F.~Huang, M.~Huang, Z.~Zhao, X.~Zhu, A knowledge-enhanced pretraining
  model for commonsense story generation, Trans. Assoc. Comput. Linguistics 8
  (2020) 93--108.

\bibitem{DBLP:conf/acl/Resnik92}
P.~Resnik, A class-based approach to lexical discovery, in: 30th Annual Meeting
  of the Association for Computational Linguistics, ACL 1992, Newark, Deleware,
  USA, 1992, pp. 327--329.

\bibitem{DBLP:conf/naacl/Resnik93}
P.~Resnik, Semantic classes and syntactic ambiguity, in: Human Language
  Technology: Proceedings of a Workshop Held at Plainsboro, New Jersey, USA,
  Morgan Kaufmann, 1993.

\bibitem{resnik1997selectional}
P.~Resnik, Selectional preference and sense disambiguation, in: Tagging Text
  with Lexical Semantics: Why, What, and How?, 1997.

\bibitem{DBLP:journals/coling/LiA98}
H.~Li, N.~Abe, Generalizing case frames using a thesaurus and the {MDL}
  principle, Comput. Linguistics 24~(2) (1998) 217--244.

\bibitem{DBLP:journals/coling/ClarkW02}
S.~Clark, D.~J. Weir, Class-based probability estimation using a semantic
  hierarchy, Comput. Linguistics 28~(2) (2002) 187--206.

\bibitem{DBLP:conf/starsem/SeaghdhaK12}
D.~{\'{O}}. S{\'{e}}aghdha, A.~Korhonen, Modelling selectional preferences in a
  lexical hierarchy, in: Proceedings of the First Joint Conference on Lexical
  and Computational Semantics, *SEM 2012, Montr{\'{e}}al, Canada, 2012, pp.
  170--179.

\bibitem{DBLP:conf/mtsr/HollinkBO20}
L.~Hollink, A.~Bilgin, J.~van Ossenbruggen, Predicting the basic level in a
  hierarchy of concepts, in: Metadata and Semantic Research - 14th
  International Conference, {MTSR} 2020, Madrid, Spain, Vol. 1355 of
  Communications in Computer and Information Science, Springer, 2020, pp.
  22--34.

\bibitem{DBLP:conf/aaai/ParkHW16}
J.~Park, S.~Hwang, H.~Wang, Fine-grained semantic conceptualization of
  framenet, in: Proceedings of the Thirtieth {AAAI} Conference on Artificial
  Intelligence, AAAI 2016, Phoenix, Arizona, {USA}, 2016, pp. 2638--2644.

\bibitem{DBLP:conf/aaai/GongZZ16}
Y.~Gong, K.~Zhao, K.~Q. Zhu, Representing verbs as argument concepts, in:
  Proceedings of the Thirtieth {AAAI} Conference on Artificial Intelligence,
  AAAI 2016, Phoenix, Arizona, {USA}, 2016, pp. 2615--2621.

\bibitem{song2011concept}
Y.~Song, H.~Wang, Z.~Wang, H.~Li, W.~Chen, Short text conceptualization using a
  probabilistic knowledgebase, in: {IJCAI} 2011, Proceedings of the 22nd
  International Joint Conference on Artificial Intelligence, Barcelona,
  Catalonia, Spain, 2011, pp. 2330--2336.

\bibitem{DBLP:conf/cikm/SongWCW14}
Y.~Song, H.~Wang, W.~Chen, S.~Wang, Transfer understanding from head queries to
  tail queries, in: Proceedings of the 23rd {ACM} International Conference on
  Conference on Information and Knowledge Management, {CIKM} 2014, Shanghai,
  China, {ACM}, 2014, pp. 1299--1308.

\bibitem{DBLP:conf/ijcai/SongWW15}
Y.~Song, S.~Wang, H.~Wang, Open domain short text conceptualization: {A}
  generative + descriptive modeling approach, in: Proceedings of the
  Twenty-Fourth International Joint Conference on Artificial Intelligence,
  {IJCAI} 2015, Buenos Aires, Argentina, 2015, pp. 3820--3826.

\bibitem{DBLP:conf/ijcai/KimWO13}
D.~Kim, H.~Wang, A.~H. Oh, Context-dependent conceptualization, in: {IJCAI}
  2013, Proceedings of the 23rd International Joint Conference on Artificial
  Intelligence, Beijing, China, 2013, pp. 2654--2661.

\bibitem{DBLP:conf/ijcai/WangZWMW15}
Z.~Wang, K.~Zhao, H.~Wang, X.~Meng, J.~Wen, Query understanding through
  knowledge-based conceptualization, in: Proceedings of the Twenty-Fourth
  International Joint Conference on Artificial Intelligence, {IJCAI} 2015,
  Buenos Aires, Argentina, 2015, pp. 3264--3270.

\bibitem{DBLP:conf/icde/HuaWWZZ15}
W.~Hua, Z.~Wang, H.~Wang, K.~Zheng, X.~Zhou, Short text understanding through
  lexical-semantic analysis, in: 31st {IEEE} International Conference on Data
  Engineering, {ICDE} 2015, Seoul, South Korea, {IEEE}, 2015, pp. 495--506.

\bibitem{DBLP:conf/aaai/ChenHLXJ19}
J.~Chen, Y.~Hu, J.~Liu, Y.~Xiao, H.~Jiang, Deep short text classification with
  knowledge powered attention, in: The Thirty-Third {AAAI} Conference on
  Artificial Intelligence, {AAAI} 2019, Honolulu, Hawaii, USA, 2019, pp.
  6252--6259.

\bibitem{DBLP:conf/flairs/BlancoCM11}
E.~Blanco, H.~C. Cankaya, D.~I. Moldovan, Commonsense knowledge extraction
  using concepts properties, in: Proceedings of the Twenty-Fourth International
  Florida Artificial Intelligence Research Society Conference, Palm Beach,
  Florida, {USA}, 2011.

\bibitem{DBLP:journals/corr/abs-2103-11302}
O.~Emebo, A.~S. Varde, O.~J. Daramola, Common sense knowledge, ontology and
  text mining for implicit requirements, in: DMIN 2016: Proceedings of the 2016
  International Conference on Data Mining, {CSREA}, 1992.

\bibitem{DBLP:conf/ranlp/Barbu09}
E.~Barbu, Acquisition of common sense knowledge for basic level concepts, in:
  Recent Advances in Natural Language Processing, {RANLP} 2009, Borovets,
  Bulgaria, 2009, pp. 23--27.

\bibitem{DBLP:conf/icde/LiuZWWJZXX20}
J.~Liu, Y.~Zhou, D.~Wu, C.~Wang, H.~Jiang, S.~Zhang, B.~Xu, Y.~Xiao, Mining
  verb-oriented commonsense knowledge, in: 36th {IEEE} International Conference
  on Data Engineering, {ICDE} 2020, Dallas, Texas, USA, {IEEE}, 2020, pp.
  1830--1833.

\bibitem{liu2022vocsk}
J.~Liu, T.~Chen, C.~Wang, J.~Liang, L.~Chen, Y.~Xiao, Y.~Chen, K.~Jin, {VoCSK}:
  {V}erb-oriented commonsense knowledge mining with taxonomy-guided induction,
  Artif. Intell. 310 (2022) 103744.

\bibitem{DBLP:journals/corr/abs-2005-00660}
S.~Bhakthavatsalam, C.~Anastasiades, P.~Clark, {GenericsKB}: {A} knowledge base
  of generic statements, CoRR abs/2005.00660.

\bibitem{DBLP:journals/corr/abs-2212-09246}
C.~Bhagavatula, J.~D. Hwang, D.~Downey, R.~Le~Bras, X.~Lu, L.~Qin,
  K.~Sakaguchi, S.~Swayamdipta, P.~West, Y.~Choi, {I}2{D}2: {I}nductive
  knowledge distillation with {N}euro{L}ogic and self-imitation, in:
  Proceedings of the 61st Annual Meeting of the Association for Computational
  Linguistics, ACL 2023, Toronto, Canada, 2023, pp. 9614--9630.

\bibitem{DBLP:journals/corr/abs-2205-11658}
E.~Allaway, J.~D. Hwang, C.~Bhagavatula, K.~R. McKeown, D.~Downey, Y.~Choi,
  Penguins don't fly: {R}easoning about generics through instantiations and
  exceptions, in: Proceedings of the 17th Conference of the European Chapter of
  the Association for Computational Linguistics, {EACL} 2023, Dubrovnik,
  Croatia, 2023, pp. 2610--2627.

\bibitem{DBLP:journals/corr/abs-1907-08015}
X.~Ding, Z.~Li, T.~Liu, K.~Liao, {ELG:} {A}n event logic graph, CoRR
  abs/1907.08015.

\bibitem{DBLP:conf/akbc/YuZSNS20}
C.~Yu, H.~Zhang, Y.~Song, W.~Ng, L.~Shang, Enriching large-scale eventuality
  knowledge graph with entailment relations, in: Conference on Automated
  Knowledge Base Construction, {AKBC} 2020, Virtual, 2020.

\bibitem{DBLP:conf/acl/YuWLBSLG0Y23}
C.~Yu, W.~Wang, X.~Liu, J.~Bai, Y.~Song, Z.~Li, Y.~Gao, T.~Cao, B.~Yin,
  {FolkScope}: {I}ntention knowledge graph construction for e-commerce
  commonsense discovery, in: Findings of the Association for Computational
  Linguistics: {ACL} 2023, Toronto, Canada, 2023, pp. 1173--1191.

\bibitem{baader2004description}
F.~Baader, I.~Horrocks, U.~Sattler, Description logics, in: Handbook on
  ontologies, Springer, 2004, pp. 3--28.

\bibitem{hanks1986default}
S.~Hanks, D.~McDermott, Default reasoning, nonmonotonic logics, and the frame
  problem, in: Proceedings of the Fifth AAAI National Conference on Artificial
  Intelligence, 1986, pp. 328--333.

\bibitem{brewka1987logic}
G.~Brewka, D.~Augustin, The logic of inheritance in frame systems, in: In Intl.
  Joint Conference on Artificial Intelligence, 1987.

\bibitem{horty1988mixing}
J.~F. Horty, R.~H. Thomason, Mixing strict and defeasible inheritance, in:
  Proceedings of the Seventh AAAI National Conference on Artificial
  Intelligence, 1988, pp. 427--432.

\bibitem{touretzky1984implicit}
D.~S. Touretzky, Implicit ordering of defaults in inheritance systems., in:
  AAAI, 1984, pp. 322--325.

\bibitem{vogel1996human}
C.~Vogel, Human reasoning with negative defaults, in: International Conference
  on Formal and Applied Practical Reasoning, Springer, 1996, pp. 606--621.

\bibitem{CAT}
W.~Wang, T.~Fang, B.~Xu, C.~Y.~L. Bo, Y.~Song, L.~Chen, {CAT}: A contextualized
  conceptualization and instantiation framework for commonsense reasoning, in:
  Proceedings of the 61st Annual Meeting of the Association for Computational
  Linguistics, Association for Computational Linguistics, Toronto, Canada,
  2023, pp. 13111--13140.

\bibitem{DBLP:conf/www/BrauerHHLNB10}
F.~Brauer, M.~Huber, G.~Hackenbroich, U.~Leser, F.~Naumann, W.~M. Barczynski,
  Graph-based concept identification and disambiguation for enterprise search,
  in: Proceedings of the 19th International Conference on World Wide Web, {WWW}
  2010, Raleigh, North Carolina, USA, {ACM}, 2010, pp. 171--180.

\bibitem{DBLP:conf/aaai/YatesGF15}
A.~Yates, N.~Goharian, O.~Frieder, Extracting adverse drug reactions from
  social media, in: Proceedings of the Twenty-Ninth {AAAI} Conference on
  Artificial Intelligence, AAAI 2015, Austin, Texas, {USA}, 2015, pp.
  2460--2467.

\bibitem{DBLP:conf/www/RajagopalCOK13}
D.~Rajagopal, E.~Cambria, D.~Olsher, K.~Kwok, A graph-based approach to
  commonsense concept extraction and semantic similarity detection, in: 22nd
  International World Wide Web Conference, {WWW} '13, Rio de Janeiro, Brazil,
  2013, pp. 565--570.

\bibitem{DBLP:conf/cicling/PoriaAGHH14}
S.~Poria, B.~Agarwal, A.~F. Gelbukh, A.~Hussain, N.~Howard, Dependency-based
  semantic parsing for concept-level text analysis, in: Computational
  Linguistics and Intelligent Text Processing - 15th International Conference,
  CICLing 2014, Kathmandu, Nepal, Vol. 8403 of Lecture Notes in Computer
  Science, Springer, 2014, pp. 113--127.

\bibitem{DBLP:conf/eacl/BeckerKF21}
M.~Becker, K.~Korfhage, A.~Frank, {COCO-EX:} {A} tool for linking concepts from
  texts to conceptnet, in: Proceedings of the 16th Conference of the European
  Chapter of the Association for Computational Linguistics: System
  Demonstrations, {EACL} 2021, Online, 2021, pp. 119--126.

\bibitem{DBLP:conf/cikm/ChenLXX18}
L.~Chen, J.~Liang, C.~Xie, Y.~Xiao, Short text entity linking with fine-grained
  topics, in: Proceedings of the 27th {ACM} International Conference on
  Information and Knowledge Management, {CIKM} 2018, Torino, Italy, {ACM},
  2018, pp. 457--466.

\bibitem{DBLP:conf/emnlp/GaoHKWMB22}
S.~Gao, J.~D. Hwang, S.~Kanno, H.~Wakaki, Y.~Mitsufuji, A.~Bosselut, Comfact:
  {A} benchmark for linking contextual commonsense knowledge, in: Findings of
  the Association for Computational Linguistics: {EMNLP} 2022, Abu Dhabi,
  United Arab Emirates, 2022, pp. 1656--1675.

\bibitem{DBLP:conf/acl/LevyZCC18}
E.~Choi, O.~Levy, Y.~Choi, L.~Zettlemoyer, Ultra-fine entity typing, in:
  Proceedings of the 56th Annual Meeting of the Association for Computational
  Linguistics, {ACL} 2018, Melbourne, Australia, 2018, pp. 87--96.

\bibitem{DBLP:conf/acl/OnoeBMD20}
Y.~Onoe, M.~Boratko, A.~McCallum, G.~Durrett, Modeling fine-grained entity
  types with box embeddings, in: Proceedings of the 59th Annual Meeting of the
  Association for Computational Linguistics and the 11th International Joint
  Conference on Natural Language Processing, {ACL/IJCNLP} 2021, Virtual Event,
  2021, pp. 2051--2064.

\bibitem{DBLP:conf/acl/DaiSW20}
H.~Dai, Y.~Song, H.~Wang, Ultra-fine entity typing with weak supervision from a
  masked language model, in: Proceedings of the 59th Annual Meeting of the
  Association for Computational Linguistics and the 11th International Joint
  Conference on Natural Language Processing, {ACL/IJCNLP} 2021, Virtual Event,
  2021, pp. 1790--1799.

\bibitem{DBLP:conf/emnlp/LiuLXH0W21}
Q.~Liu, H.~Lin, X.~Xiao, X.~Han, L.~Sun, H.~Wu, Fine-grained entity typing via
  label reasoning, in: Proceedings of the 2021 Conference on Empirical Methods
  in Natural Language Processing, {EMNLP} 2021, Virtual Event / Punta Cana,
  Dominican Republic, 2021, pp. 4611--4622.

\bibitem{meyers2007annotation}
A.~Meyers, Annotation guidelines for {NomBank}--noun argument structure for
  {PropBank} 2007 (2007).

\end{thebibliography}

\clearpage
\appendix

\section{Concept linking rules}
\label{Appendix:ec_rules}

As shown in Table~\ref{Tab:word_cases} above, in addition to the simple case when we can directly match the headword with some concept, we also consider various other phenomena. Additional explanations are provided below.

\paragraph{Headword concepts}
The simplest case is when the headword itself directly corresponds to the concept. For each nominal headword, we try to match possible concepts with both of its original and lemmatized forms (using spaCy), possibly prepended/appended by the modifiers before or after it within the constituent according to the DEP tag. Regarding predicative candidates, in addition to finding the nominal form by WordNet or NOMBANK (more specifically, NOMLEX-PLUS), we also search for the adjuncts and arguments by DEP tag after it within the constituent, so that we can identify phrasal verbs in WordNet or verb phrases by combining with the argument.

\paragraph{Non-headword concepts}
The headword of the constituent may not represent the actual concept. For nominal candidates, transparent nouns like classifiers (e.g. ``a \ul{cup} of tea''), which take the semantic property of their argument, are thoroughly discussed by NOMBANK \cite{meyers2007annotation}. In such cases, the head noun is more like an optional modifier to its argument, hence we directly conceptualize the argument without considering the head noun.
While there are also words that are not strictly transparent (e.g. ``a \ul{group} of people'') and not covered in NOMBANK, in which case we include both the head noun itself and the argument, as both of them make sense as the conceptualization of the constituent, and the combination of them, though more precise, may not appear in the taxonomy. After all, they can still be filtered out in later stages if found not suitable in the end. As for predicative candidates, we follow the same idea when handling light verbs and catenative verbs (e.g. ``\ul{take} a shower'' and ``\ul{want} to go'') by including both of them for further conceptualization. While in the more certain cases like auxiliary verbs (e.g. ``\ul{used} to swim'') or raising-to-subject verbs (e.g. ``\ul{appears} to be happy'') that are similar to transparent construction with limited semantic impact, we conceptualize the predicand directly. For certain verbs (e.g. ``\ul{go} shopping'') and adjectives (e.g. ``be \ul{able} to speak English'') with rather light semantic meaning comparable to auxiliary/modal verbs, we follow this protocol as well.

\paragraph{Conjunction}
A special case is a conjunction like ``doctors and nurses,'' which is unlikely to be present in the taxonomy. However, the inferences on the event are often both valid when applied to the two abstract events based on either side: after PersonX ``sees doctors and nurses,'' ``sees [doctor],'' or ``sees [nurse],'' in all the three cases PersonX will get treated.
Therefore, in this work, to enrich abstract knowledge for further use, we link the constituent to both ``[doctor]'' and ``[nurse]'' based on its two conjuncts, though the constituent is not precisely their instances. 

\section{Data collection}
\subsection{Data preparation}
\label{Appendix:clean}
Regarding the foremost step of data cleaning, we first attempt to fix a number of language errors in ATOMIC events, including incomplete sentences, errors in verb forms, misuse of inflections, inconsistency of tense and aspect, missing verbs and marks, etc. We curate a set of rules to fix them and to change some expressions (such as replacing PersonX/Y/Z with person names and underlines to proper-noun-like names, and fixing missing determiners and possession case marks) so that the sentences can be more accurately parsed. Total 5.4K events are changed in this cleaning process. Then we parse the sentences and restore the replaced words. Next, we remove events with obviously impossible dependency trees. We find that they indicate unresolved grammatical errors. Finally, we remove all duplicate events after the cleaning. 

Regarding the filtering of idioms, we match the events with the sources like entries crawled from the English Wiktionary labeled as idioms, which is supposed to be the original source of those idioms in ATOMIC. Later, when identifying the entities/eventualities, we exclude cases like PersonX/Y/Z and pronouns which have limited value for conceptualization, save transparent ones like ``all of the people." As for the de-duplication process, similar sentences and clauses such as those with only differences in a set of minor modifiers (e.g. determiners) are merged. Also, for better entity diversity, a candidate is skipped if it is linked to a non-verbal WordNet synset with a probability above 0.75 from GlossBERT and the synset has been included for more than five times. When considering the instance triples, all ``none'' tails are excluded.

\subsection{Crowd-sourced annotation}
\label{Appendix:annotation}

\paragraph{Annotation platform}
During the crowd-sourced annotation, we leverage the Mechanical Turk platform to recruit the workers, run qualifications, and collect the results. In addition to the HTML templates provided by the website, we also leverage the functionality of \textit{external questions} to host the question by ourselves, which allows flexible interaction with the back-end such as matching the typed concept with Probase instantly, so as to optimize the experience and productivity of the workers.

\paragraph{Annotation targets}
As reflected by the scale of Abstract ATOMIC, we can induce an extremely large number of possible abstract triples from ATOMIC, hence we are bound to annotate a very tiny subset of them. Therefore, for better representativeness, instead of sampling a set of largely unrelated triples to annotate, we choose to sample a subset of identified conceptualization candidates (entities or eventualities) in the original ATOMIC events. For each selected candidate, we include abstract triples upon all its corresponding conceptualizations. In this way, all the possible positive and negative samples based on each selected candidate (which contrast with each other) can be completely depicted. Furthermore, due to the imbalance of the types of candidates regarding their DEP, to be illustrated below, abstract events on candidates with less-presented DEP tags are up-sampled by exponential balance to give more diversity. The number of selected candidates of each DEP $N_i \propto p_i ^ {0.8}$, in which $p_i$ is the proportion of candidates with each DEP among all DEP tags.

\paragraph{Quality assurance}
We carry out multiple quality assurance measures. Regarding worker recruitment, we only consider experienced workers with above 95\% past acceptance rate on at least 1,000 tasks. We then provide carefully written instructions for workers that cover various cases in the annotation corresponding to those in Table~\ref{Tab:word_cases}, along with more than 30 examples for each round. In this way, the workers may understand the goal to find correct event conceptualizations and to determine the validity of abstract triples according to our defeasible and intuition-based criterion upon their instances.
For the first two rounds, tests with confirmed answers are randomly added to the first 100 responses of each worker to probe the workers' correctness. Additionally, for the second and the third rounds that require more accurate answers, workers need to pass qualification tests beforehand to prove their understanding of our requirements. As for the second round, two tests are used. The first practice test allows the worker to take it multiple times, and the worker could pass it only if all of the 13 yes-or-no questions are correctly answered, so that the workers are well-trained on the requirements. The second one, which requires a score of 13 out of 15, serves as a real gatekeeper that the worker may take only once a day. 
As for the third round, we conduct a qualification test as well, with a bar of 9 out of 10. All test questions are carefully curated to represent typical but challenging cases in the annotation.

\subsection{Annotation statistics}
\label{Appendix:anno_stat}

\begin{figure}[t]
\centering
\includegraphics[width=0.98\textwidth]{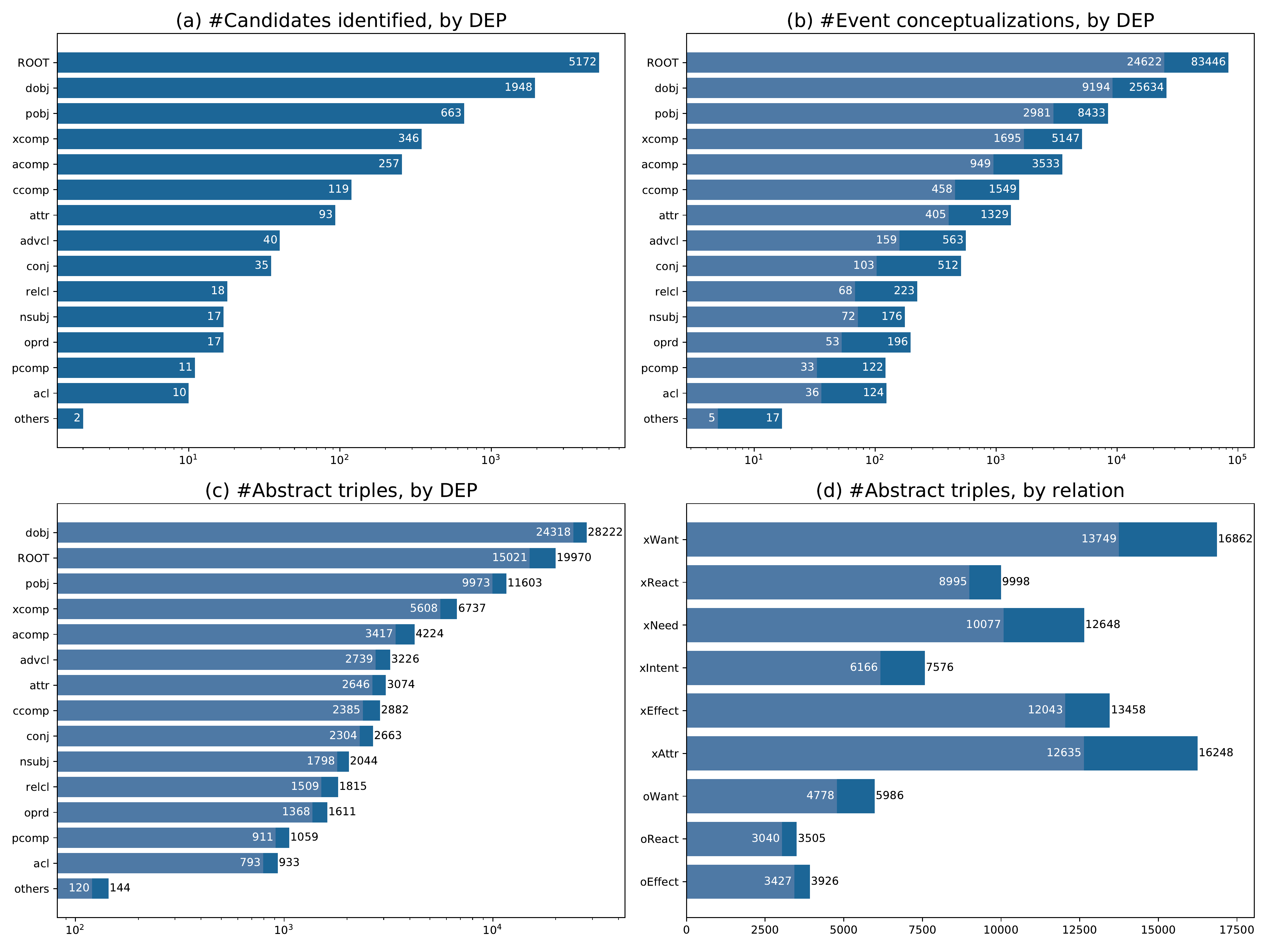}
\caption{Statistics of data used in and from annotation, when split by DEP tags and relations.
Positive samples are shown in the lighter-colored sub-bars. Due to the large range, the first three sub-figures are in log scale.
}
\label{Fig:anno_stat}
\end{figure}

Some additional statistics are given in Figure~\ref{Fig:anno_stat}, which highlights the distribution of candidates with different DEP tags. As shown in (a), 24,622 (60.3\%) entities or eventualities as conceptualization candidates have the \textit{ROOT} DEP tag corresponding to the whole sentence. This is intuitive because each ATOMIC event itself can be conceptualized as a whole, but it does not necessarily contain other candidates. Following \textit{ROOT}, other common DEP tags are \textit{dobj} and \textit{pobj} for nominal objects. There is a tail of other less common DEP tags, such as complement and clauses, many of them representing verbal concepts inside an event. This imbalance of DEP is the reason for us to implement the DEP-balanced sampling for picking the samples for triple-level annotation. To construct a diverse and complete dataset, all these types are included. As shown in (b) and (c), though not completely balanced in positive rate, candidates of each type of DEP produce a number of different event conceptualizations and abstract triples deemed positive. 
Numbers of abstract triples of each relation are imbalanced, which is due to the imbalance in the original ATOMIC, but added by the discrepancy of positive rates, ranging from 77.8\% to 90.0\%, as in (d).

\section{Empirical experiments}
\label{Appendix:exp_detail}
\subsection{Model configuration}

In all the experiments, the model configurations are decided empirically. In a series of pilot experiments, we attempt to use alternative text templates, such as different usage of special concept tokens and different prompts, but no meaningful improvements are found. The case is similar to alternative hyper-parameters.

The conceptualization verifier is derived from RoBERTa-base under the common fine-tuning setting, with batch size 64 and learning rate 2e-5 to match our computational resources, and so is the conceptualization generator, from GPT2-base with batch size 32 and learning rate 1e-5. 

The models are trained using the \textit{trn} subset of the dataset and validated on the \textit{dev} subset. Regarding the abstract event dataset, the partition of each abstract event is decided by the ATOMIC partition of its instance ATOMIC event. When there are many, we follow the event that appears the earliest in ATOMIC. The partition of each abstract triple follows the partition of the head abstract event.

The discriminator or verifier is measured by accuracy with the classification threshold tuned on \textit{dev} set, while the generator is measured by BLEU-2 scores from the best output of a 10-beam search. The model with the best performance on \textit{dev} subset across different training steps is used to report the result on the \textit{tst} subset.

\subsection{Model prompts}
\label{Appendix:prompt}

As mentioned above, we formulate the samples into textual inputs to the NLP model by textual templates/prompts chosen through pilot experiments, as demonstrated by the examples below.

As for the conceptualization verifier on event conceptualization, we concatenate the event (with the candidate indicated) and the concept, separated by the [SEP] token. For example, when conceptualizing ``PersonX drinks \ul{a cup of coffee}'' into ``beverage'', the whole prompt will be:

\begin{blocktemplate}
[CLS] PersonX drinks [a cup of coffee] [SEP] beverage [SEP]
\end{blocktemplate}

As for the neural concept generator, the model is expected to generate the tokens after the [GEN] token, similar to COMET. For the example above, the whole prompt will be:

\begin{blocktemplate}
PersonX drinks \textlangle c\textrangle~a cup of coffee \textlangle /c\textrangle~. \textlangle c\textrangle~a cup of coffee \textlangle /c\textrangle~is an instance of [GEN] beverage [EOS]
\end{blocktemplate}

As for zero-shot conceptualization verifiers like GPT2 without any fine-tuning, we replace the PersonX/Y/Z with persons' names and try to feed the input with a prompt more similar to natural language. For the example above, the whole prompt will be:

\begin{blocktemplate}
Anderson drinks ``a cup of coffee." ``a cup of coffee'' is an instance of beverage [EOS]
\end{blocktemplate}

As for the inference verifier, we use a template similar to the conceptualization verifier, but with the relations indicated using prompts given in Table~\ref{Tab:hint}. For example, for $\langle$\textit{h}: PersonX drinks [coffee], \textit{r}: xReact, \textit{t}: refreshed$\rangle$, the whole prompt will be:

\begin{blocktemplate}
PersonX drinks [coffee] [SEP] After that PersonX feels: refreshed [EOS]
\end{blocktemplate}

While regarding the Generator (ATOMIC) model trained in a COMET-like way for discriminating abstract triples, we also adopt the COMET template with a separate token per relation. For the example above, the whole prompt will be:

\begin{blocktemplate}
PersonX drinks coffee [EOS] [GEN] [xReact] refreshed [EOS]
\end{blocktemplate}

\begin{table}[t]
\centering
\begin{tabular}{ll}
\toprule
Relation & Prompt \\ \midrule
xNeed   & Before that PersonX needs:    \\
xIntent & PersonX's intention is:       \\
xAttr   & PersonX will be described as: \\
xEffect & Effects on PersonX will be:   \\
xWant   & After that PersonX wants:     \\
xReact  & After that PersonX feels:     \\
oEffect & Effects on others will be:    \\
oWant   & After that others want:       \\
oReact  & After that others feel:   \\   
\bottomrule
\end{tabular}

\caption{\label{Tab:hint} Textual prompt for the relations used in the template for inference verifier. }
\end{table}

\section{Abstract ATOMIC statistics}
\label{Appendix:abs_atomic}
\begin{figure}[h]
\centering
\includegraphics[width=0.95\textwidth]{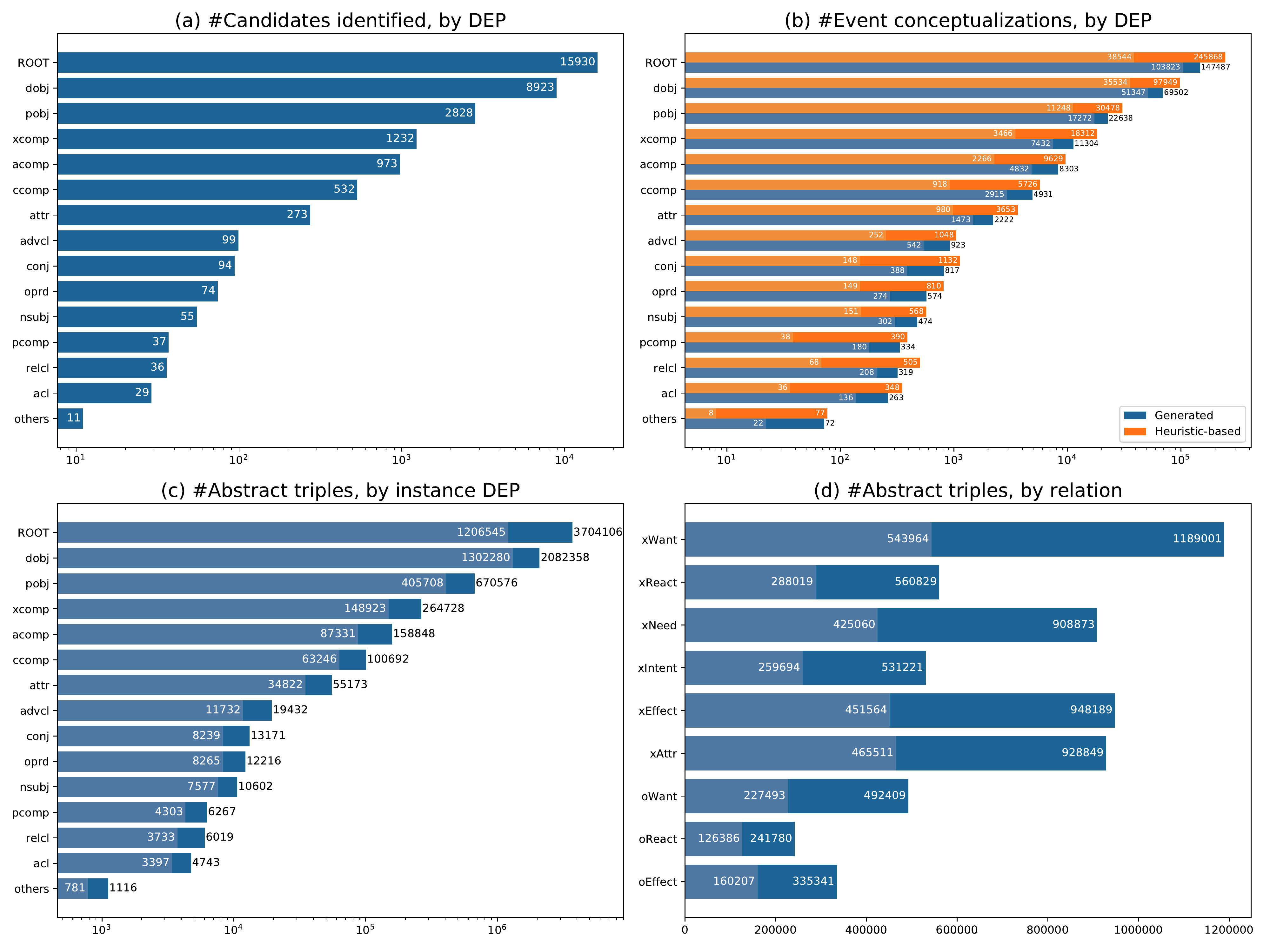}
\caption{Statistics of data used in and from Abstract ATOMIC, when split by DEP tags and relations. Positive samples are shown in the lighter-colored sub-bars. Due to the large range, the first three sub-figures are in log scale.
}
\label{Fig:aug_stat}
\end{figure}

More statistics of Abstract ATOMIC are demonstrated in Figure~\ref{Fig:aug_stat}. Discussions in Section~\ref{Sec:abs_atomic} are supported by (b): Heuristic rules (orange bar) produce a large number of potential event conceptualizations, but the acceptance rate is relatively low. While the generator (blue bar), as a PTLM, produces fewer samples but fits better with the context. This highlights the crucial role of the PTLM-based verifier to filter out the event conceptualizations unsuitable in the context. In addition, Abstract ATOMIC has an imbalance similar to the annotated data as shown in Figure~\ref{Fig:anno_stat} regarding DEP tags and relations, due to the limitation of the base dataset ATOMIC, but still it reaches a good coverage of various DEP tags (i.e., types of entities and eventualities) and relations. 

\end{document}